\documentclass[sn-nature]{sn-jnl}%

\usepackage{graphicx}%
\graphicspath{{./}{Figures/}}%
\usepackage{multirow}%
\usepackage{amsmath,amssymb,amsfonts}%
\usepackage{amsthm}%
\usepackage{mathrsfs}%
\usepackage{tcolorbox}
\usepackage[title]{appendix}%
\usepackage{xcolor}%
\usepackage{soul} 
\usepackage{textcomp}%
\usepackage{manyfoot}%
\usepackage{url}
\usepackage{booktabs}%
\usepackage{tabularx}
\usepackage{array}
\usepackage{listings}%
\usepackage{adjustbox}
\usepackage{siunitx} %
\usepackage{geometry}
\usepackage{multirow}
\usepackage{comment}

\newcommand{\SIref}[2]{#2}
\usepackage{floatpag}%

\newcommand{\FigFiveGhcnDay}{2022-05-28}

\newcommand{\FigFiveGhcnNTestDays}{847}

\newcommand{\FigFiveGridStep}{0.05}
\newcommand{\FigFiveGridReproMax}{0}

\newcommand{\FigFiveGridOutsidePct}{0.4}

\newcommand{\GhcnUniqueHeldOut}{265}
\newcommand{\GhcnDeltaDistRho}{-0.16}
\newcommand{\GhcnDeltaDistP}{0.011}

\newcommand{\GhcnDeltaClippedValues}{$-$2.91, $-$1.52, $-$1.43}
\newcommand{\GhcnDeltaUniqueMedian}{-0.008}
\newcommand{\GhcnDeltaUniqueNNeg}{135}
\newcommand{\GhcnBasePlaneRmse}{5.21}

\newcommand{\GhcnBaseNearestRmse}{6.60}
\newcommand{\GhcnBaseClimRmse}{5.17}

\newcommand{\FigFiveCSLstmRmse}{3.45}
\newcommand{\FigFiveCSLstmRmseSd}{0.04}
\newcommand{\FigFiveCTwoStageRmse}{3.53}

\newcommand{\GhcnPairSLstmZeroTopStation}{USW00023273}
\newcommand{\GhcnPairSLstmZeroTopSharePct}{51}

\newcommand{\FigFiveDhStation}{GUE}

\newcommand{\FigFiveDhNBlocks}{6}
\newcommand{\FigFiveDhNHeldOut}{40}

\newcommand{\PeakTestFirst}{2023-12-31}
\newcommand{\PeakTestLast}{2025-10-13}

\newcommand{\FigFiveDhStationElev}{2286}
\newcommand{\FigFiveDhSFcnWorseN}{37}

\newcommand{\PressTestFirst}{2024-01-01}

\newcommand{\PressNTestWindows}{11{,}086}

\newcommand{\PressTwoStageRmse}{1.91}

\newcommand{\PressHypsRmse}{2.71}

\newcommand{\PressClockOnlyRmse}{3.94}

\newcommand{\PressGridExtrapPct}{4.5}

\newcommand{\DhTestNBeforeCut}{70}
\newcommand{\DhTestPctBeforeCut}{0.7}
\newcommand{\DhTestN}{9974}
\newcommand{\DhGapHoursMissing}{19}

\newcommand{\DhValCutoff}{2023-12-31 00:00}
\newcommand{\GhcnTestDaysZero}{847}
\newcommand{\GhcnReachBackZero}{13}
\newcommand{\GhcnShareValZero}{13}

\newcommand{\GhcnTestDaysOne}{996}
\newcommand{\GhcnReachBackOne}{5}
\newcommand{\GhcnShareValOne}{0}

\newcommand{\GhcnTestDaysTwo}{885}
\newcommand{\GhcnReachBackTwo}{13}
\newcommand{\GhcnShareValTwo}{13}

\newcommand{\DhTwentyFourSFcnClockOnUnexpl}{5.2}

\newcommand{\DhTwentyFourSLstmClockOnUnexpl}{3.1}

\newcommand{\DhSeventyTwoSLstmClockOffUnexpl}{2.4}

\newcommand{\DhSeventyTwoSMlpClockOffUnexpl}{2.6}

\newcommand{\DhSeventyTwoSFcnClockOnUnexpl}{8.2}

\newcommand{\DhSeventyTwoSLstmClockOnUnexpl}{2.9}

\newcommand{\DhSeventyTwoSMlpClockOnUnexpl}{2.6}

\newcommand{\DhInstantSFcnClockOnUnexpl}{4.1}

\newcommand{\DhInstantSLstmClockOnUnexpl}{4.8}

\newcommand{\PressRT}{1.866}

\newcommand{\PressSeedSd}{0.048}

\newcommand{\PressHOneDelta}{-2.098}
\newcommand{\PressHOneLo}{-2.442}
\newcommand{\PressHOneHi}{-1.767}

\newcommand{\PressHTwoDelta}{-0.040}
\newcommand{\PressHTwoLo}{-0.151}
\newcommand{\PressHTwoHi}{0.073}

\newcommand{\PressHThreeOutcome}{pass}

\newcommand{\GhcnPcOneRawPct}{79}

\newcommand{\GhcnPcNLocations}{886}

\newcommand{\PeakPcOneRawPct}{92}

\newcommand{\PeakPcNLocations}{105}

\newcommand{\DosePcOnePct}{99.98}

\begin{document}

\title[Spatiotemporal Reconstruction of Global Radiation from Sparse Sensor Sequences
]{Spatiotemporal Reconstruction of Global Radiation from Sparse Sensor Sequences
}

\author[1]{\fnm{Kazuma} \sur{Kobayashi}}\email{kazumak2@illinois.edu}

\author[2]{\fnm{Tapas} \sur{Tripura}}\email{tapas.t@am.iitd.ac.in}

\author[1]{\fnm{Jay Phil} \sur{Yoo}}\email{jayyoo2@illinois.edu}

\author[1]{\fnm{Samrendra} \sur{Roy}}\email{roysam@illinois.edu}

\author[3,4]{\fnm{Diab} \sur{Abueidda}}\email{abueidd2@illinois.edu}

\author[4,5]{\fnm{Seid} \sur{Koric}}\email{koric@illinois.edu}

\author[1,2,6]{\fnm{Souvik} \sur{Chakraborty}}\email{souvik@am.iitd.ac.in}

\author*[1,4]{\fnm{Syed Bahauddin} \sur{Alam}}\email{alams@illinois.edu}

\affil*[1]{\orgdiv{Nuclear, Plasma and Radiological Engineering, The Grainger College of Engineering}, \orgname{University of Illinois Urbana-Champaign}, \state{IL}, \country{USA}}

\affil[2]{Department of Applied Mechanics, Indian Institute of Technology Delhi, New Delhi, India}

\affil[3]{\orgdiv{Civil and Urban Engineering Department}, \orgname{New York University Abu Dhabi}, \city{Abu Dhabi}, \country{UAE}}

\affil[4]{\orgdiv{National Center for Supercomputing Applications}, \orgaddress{\street{1205 W Clark St}, \city{Urbana}, \postcode{61801}, \state{IL}, \country{USA}}}

\affil[5]{\orgdiv{Mechanical Science and Engineering, The Grainger College of Engineering}, \orgname{University of Illinois Urbana-Champaign}, \state{IL}, \country{USA}}

\affil[6]{Yardi School of Artificial Intelligence, Indian Institute of Technology Delhi, New Delhi, India}

\abstract{ Much of the physical world is inaccessible to direct measurement and can be reached only through the sparse traces it leaves in other observable variables. Here we introduce the Temporal Radiation Operator Network (TRON), a spatiotemporal neural operator that learns proxy-to-field mappings, reconstructing otherwise unobserved physical fields from sparse, indirect sensor histories. For cosmic-radiation mapping, TRON reconstructs sea-level effective dose at 65,341 locations from histories of 12 neutron monitors, achieving 0.096\% mean relative $L_2$ error against EXPACS-generated reference fields in a chronologically held-out year; batched GPU inference evaluates a global field in approximately $\sim$1~ms, compared with $\sim$57.6~s for EXPACS on one CPU thread. On measured weather networks, TRON reconstructs temperature at held-out stations from thermometer or barometer histories, outperforming interpolation, calendar-based and hypsometric baselines. These results establish proxy-to-field reconstruction as a general route to recovering spatially distributed physical information beyond the direct reach of sparse sensor networks. }


\maketitle

\section{Introduction}
Global radiation exposure must be assessed from measurements that neither cover the globe nor directly measure the dose of interest. This mismatch between what can be observed and what must be inferred is a central obstacle to spatially resolved cosmic-radiation monitoring. Its consequences are particularly important for aviation and spaceflight, where elevated exposure makes reliable dose estimation essential for radiological protection~\cite{cucinotta2006cancer,durante2008heavy}. Yet even at ground level, translating a limited monitoring network into a global dose map requires resolving the combined effects of cosmic-ray variability, geomagnetic shielding and atmospheric particle cascades. Ground-based neutron monitors record secondary particles generated by cosmic-ray interactions with the atmosphere, providing long-term observations of the radiation environment at a sparse and geographically uneven set of locations~\cite{simpson2000cosmic,usoskin2010cosmic}. These observations do not directly reveal effective dose, which additionally depends on particle composition, energy spectra and the biological weighting of energy deposited in different tissues. The reconstruction problem therefore extends beyond filling gaps between sensors: it requires inferring the global distribution of one physical quantity from localized measurements of another. Established frameworks such as EXPACS (EXcel-based Program for Calculating Atmospheric Cosmic-ray Spectrum), built on the PARMA analytical model~\cite{sato2015analytical,sato2016analytical}, and CARI-7 for aviation doses~\cite{copeland2017cari}, bridge these quantities through physically grounded calculations of particle spectra and dose. However, repeated evaluation across dense global grids imposes a computational burden that grows with spatial coverage and update frequency. Expanding the monitoring network alone cannot remove the need to translate particle measurements into dose, while accelerating individual calculations alone does not establish a direct mapping from observation histories to global fields. The resulting scientific challenge is to turn sparse, evolving proxy measurements into timely, spatially resolved radiation estimates, drawing on established physical models without repeating the full dose-estimation pipeline for every field update.

Recovering a quantity that no sensor records directly, from the signatures it leaves across a distributed detector, is an established mode of physical inference. The IceCube Neutrino Observatory provides a prominent instance: optical sensors embedded in the Antarctic ice record Cherenkov light from secondary charged particles, and the direction and energy of the originating neutrino are reconstructed from that sparse light pattern rather than imaged directly~\cite{aartsen2017instrumentation}. The 2026 Nobel Prize in Physics was awarded to Francis Halzen for decisive contributions to the IceCube Neutrino Observatory and the discovery of high-energy neutrinos of astrophysical origin~\cite{nobel2026physics,aartsen2013icecube}. The detector, the physics and the scientific objective differ from radiation-field mapping, as does the evidentiary status of the result; the inverse-sensing structure, however, is shared. IceCube reconstructs discrete particle events from distributed optical measurements, whereas the problem addressed here is reconstruction of a continuous spatial field from sparse sensor histories. This structure motivates the central learning problem considered here: mapping sparse measurement histories to a spatial field of a different physical quantity. 

Neural operators provide a natural framework for expressing this relationship by learning mappings between input and output functions, enabling a shared model to represent a family of physical responses~\cite{li2020neural,li2023fourier,lu2021learning,tripura2023wavelet,tripura2026neural}. Among these frameworks, the Deep Operator Network (DeepONet) separates the representation of an input function from the coordinates at which the output function is evaluated: a branch network encodes sampled input measurements, while a trunk network encodes the output query coordinates~\cite{lu2021learning}. This separation provides a route from sparse observations to a coordinate-dependent field without first interpolating the measurements onto a dense input grid. DeepONet has supported surrogate modeling in heat transfer, structural deformation and manufacturing~\cite{koric2023data,he2024sequential,koric2024deep,he2024predictions,garg2022assessment,kushwaha2024advanced}, establishing the ability of operator learning to connect prescribed system inputs with distributed physical responses.
In observation-driven reconstruction, however, the available measurements may provide only indirect and incomplete information about those inputs. The task then becomes one of inferring a field from the observable traces of its underlying dynamics, requiring the input representation to capture relationships among sensors and information contained in their measurement histories. Cosmic-radiation monitoring exemplifies this challenge: geographically scattered neutron-count sequences must inform a global dose field, although neither the dose nor a complete description of the physical state is supplied at inference. This configuration also differs from the dense-field prediction tasks addressed by models such as FourCastNet, DGMR, MetNet-2 and NeuralGCM, which receive spatially extensive descriptions of the system state~\cite{pathak2022fourcastnet,ravuri2021skilful,espeholt2022deep,kochkov2024neural}. The relevant architectural question is therefore how to combine the spatial decoding capability of an operator network with an input representation suited to sparse, sequential proxy observations. A branch receiving only the latest snapshot cannot exploit preceding observations, while an encoder that processes multistation sequences jointly can represent temporal context alongside relationships among sensors. Coupling such an encoder to a spatial decoder provides a direct formulation for reconstructing the field at the time of the latest observation.

Here, we introduce the Temporal Radiation Operator Network (TRON), a spatiotemporal neural operator that reconstructs a global radiation dose field from the measurement histories of just 12 neutron monitors. TRON couples a recurrent branch network, implemented using long short-term memory (LSTM) or gated recurrent unit (GRU) architectures, with a coordinate-based spatial trunk. The branch jointly encodes the multistation observations into a shared latent representation, while the trunk maps spatial query coordinates into features that combine with this representation to produce local dose estimates. This formulation connects sparse observations to a global field in a single learned mapping, reconstructing the radiation environment at the end of each input window without requiring future observations. We demonstrate TRON using daily neutron counts from the Neutron Monitor Database~\cite{nmdb}, paired with global sea-level effective-dose fields generated using EXPACS over 2001--2023~\cite{sato2015analytical,sato2016analytical}. Training on 2001--2021, validation on 2022 and testing on the held-out year 2023 establish a chronological assessment of reconstruction accuracy. For a representative single-branch LSTM model using 30-day observation windows, TRON reconstructs the field at 65,341 locations with a mean relative $L_2$ error of 0.096\% against the EXPACS reference. On the evaluated NVIDIA A100 GPU, batched inference takes approximately 1~ms per global field, compared with 57.6~s for EXPACS on a single CPU thread. Dense simulated fields provide supervision during training; at inference, the network receives only the neutron-count histories and spatial query coordinates, with dose values recovered using stored training-period scaling parameters. The resulting capability is a rapid, observation-driven reconstruction of a physics-based reference field, connecting a sparse monitoring network to global dose estimates without repeating the underlying physical calculations for each update.

Beyond radiation mapping, this inverse-sensing perspective raises a broader scientific question: under what conditions can sparse observations of accessible quantities support reconstruction of spatially distributed quantities that are inaccessible, indirect or unavailable at the locations of interest? We examine this capability using two measured environmental networks: daily temperature reconstruction from a sparse set of thermometers, and hourly temperature reconstruction from station-pressure measurements. These experiments extend the assessment from agreement with simulated dose fields to prediction of independently measured quantities at held-out stations. Across these settings, TRON improves on non-learned interpolation, calendar-based and hypsometric baselines, while comparisons with a purpose-built two-stage linear model do not establish consistent superiority. An additional experiment reconstructs accumulated heating degree-hours, providing a target whose dependence on preceding conditions explicitly tests the contribution of observation history. Together, these studies establish the applicability of the sequence-to-field formulation to both direct measurements and indirect proxies, while distinguishing access to temporal information from the particular architecture used to encode it. The radiation and measured-network demonstrations address complementary aspects of the same challenge: rapid reconstruction of a physically defined quantity that cannot be measured directly, and spatial reconstruction of measured quantities beyond the locations supplying training targets. TRON thus provides a framework for converting sparse monitoring records into spatially resolved estimates, with physics-based reference fields or measured observations supplying supervision according to the application. Its broader significance lies in connecting the limited quantities that monitoring networks observe to the distributed quantities needed to characterize an evolving environment.

\section{Results}

\subsection{Overview of the reconstruction framework}
\label{sec:architecture}

\begin{figure}[p]
    \centering
    \makebox[\textwidth][c]{\includegraphics[width=180mm]{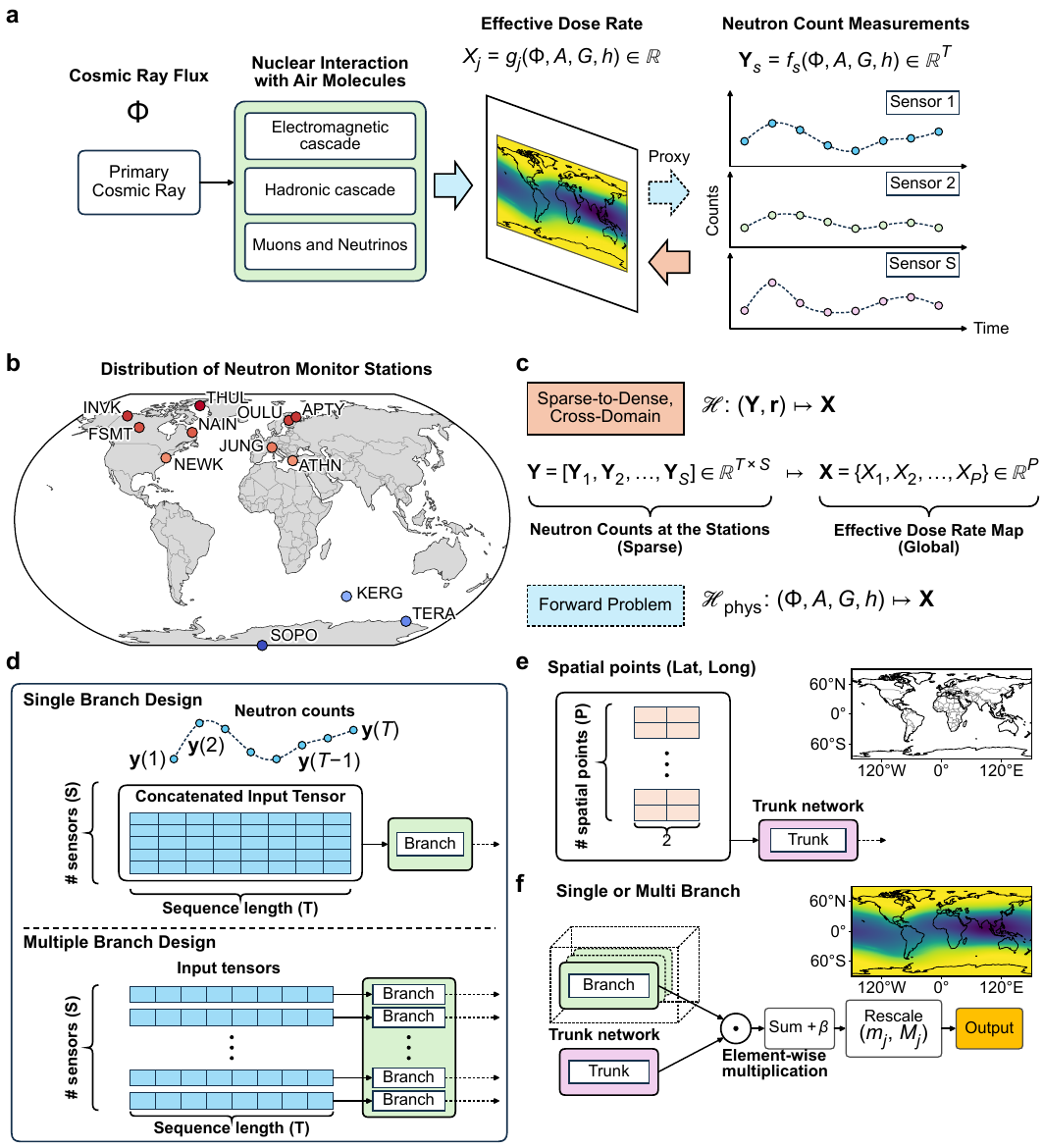}}%
\caption{Schematic overview of the effective-dose reconstruction framework.
\textbf{a} Primary cosmic-ray flux $\Phi$ drives atmospheric cascades that determine effective dose $X_j=g_j(\cdot)$ and neutron-monitor counts $\mathbf{Y}_s=f_s(\cdot)$ ($A$, atmospheric conditions; $G$, geomagnetic field; $h$, altitude; counts schematic). Light-blue arrows denote the forward process (dashed, proxy relationship); the salmon arrow denotes TRON reconstruction. The map shows the EXPACS reference on 10 September 2023, the median-error test day.
\textbf{b} Twelve input stations, colored by latitude; codes and coordinates are given in Supplementary Table~\SIref{tab:neutron_stations}{S1}.
\textbf{c} TRON maps $T$-day count histories $\mathbf{Y}$ from $S$ stations and query coordinates $\mathbf{r}$ to predicted dose $\hat{\mathbf{X}}$ at $P$ points; $\mathcal{H}_{\mathrm{phys}}$ denotes the physics-based forward mapping.
\textbf{d} Single-branch encoding of all stations versus separate station encoders followed by fusion; $\mathbf{y}(t)$ contains the $S$ counts on day $t$.
\textbf{e} Spatial trunk evaluated at latitude--longitude queries.
\textbf{f} Branch and trunk features are multiplied element-wise ($\odot$), summed and combined with bias $\beta$ to produce $\hat{z}_j$. Rescaling gives $\hat{X}_j=m_j+(M_j-m_j)\hat{z}_j$, using training-period reference extrema $m_j$ and $M_j$. The map shows the corresponding S-LSTM prediction (single-branch long short-term memory encoder; $T=30$ days; one trained model; relative $L_2$ error $0.096\%$). Both maps share a color scale of $0.027$--$0.036\,\mu\mathrm{Sv}\,\mathrm{h}^{-1}$.}
    \label{fig:problem}
\end{figure}

TRON reconstructs a global sea-level effective-dose field from the recent observation histories of a sparse neutron-monitor network (Fig.~\ref{fig:problem}a--c). Neutron counts and effective dose arise from the same underlying cosmic-ray environment, shaped by atmospheric particle cascades and geomagnetic shielding, but represent different physical quantities: the monitors record secondary particles, whereas effective dose additionally incorporates radiation- and tissue-weighting factors. We formulate their relationship as a learned reconstruction mapping,
\begin{equation}
    \hat{\mathbf{X}} = \mathcal{H}(\mathbf{Y},\mathbf{r}),
    \label{eq:tron_reconstruction}
\end{equation}
where $\mathbf{Y}\in\mathbb{R}^{T\times S}$ contains daily mean count rates from $S$ stations over $T$ consecutive days, $\mathbf{r}\in\mathbb{R}^{P\times 2}$ contains the latitude--longitude coordinates of $P$ query points, and $\hat{\mathbf{X}}\in\mathbb{R}^{P}$ contains the corresponding predicted effective dose rates on the final day of the observation window. The task is therefore reconstruction at the latest observation time, with no future measurements supplied. For training and evaluation, we paired daily observations from 12 stations in the Neutron Monitor Database (NMDB)~\cite{nmdb} with EXPACS reference fields spanning 2001--2023. The stations cover both hemispheres and a range of geomagnetic cut-off rigidities (Fig.~\ref{fig:problem}b; Supplementary Table~\SIref{tab:neutron_stations}{S1}). EXPACS implements the PARMA atmospheric cosmic-ray model and converts particle spectra into effective dose using established fluence-to-dose coefficients~\cite{sato2015analytical,sato2016analytical,icrp2010conversion}. Each reference field contains 65,341 evaluation points on a global $1^\circ\times1^\circ$ latitude--longitude grid, with dose rates expressed in $\mu\mathrm{Sv}\,\mathrm{h}^{-1}$. We used 2001--2021 for training, 2022 for validation and 2023 for testing, and considered observation windows of 7, 30, 60 and 90 days. This chronological design evaluates reconstruction in a held-out year against a physics-based reference; the dose fields are simulated targets rather than direct dosimetric measurements.

TRON implements this mapping through a temporal branch encoder and a spatial trunk encoder, building on the Deep Operator Network (DeepONet) formulation~\cite{lu2021learning} (Fig.~\ref{fig:problem}d--f). The branch compresses the observation window into a latent vector $\mathbf{b}\in\mathbb{R}^{d}$, where $d$ is the latent dimension, while the feed-forward trunk maps each query coordinate $\mathbf{r}_j$ to a vector $\mathbf{t}(\mathbf{r}_j)\in\mathbb{R}^{d}$. Their inner product determines the normalized dose prediction, which is converted to physical units using stored output-scaling parameters:
\begin{equation}
    \hat{z}_j =
    \mathbf{b}^{\mathsf{T}}\mathbf{t}(\mathbf{r}_j)+\beta,
    \qquad
    \hat{X}_j =
    m_j+(M_j-m_j)\hat{z}_j,
    \label{eq:tron_readout}
\end{equation}
where $\beta$ is a learnable scalar bias, and $m_j$ and $M_j$ are the minimum and maximum reference dose rates at point $j$ over the training period. We evaluate two branch configurations: a single encoder that processes all station sequences jointly, and separate station-specific encoders whose latent outputs are combined by element-wise multiplication. Each configuration uses either a long short-term memory (LSTM) or a gated recurrent unit (GRU) encoder, yielding four TRON variants. The shared branch representation conditions the field on the observation history, while the trunk supplies its coordinate-dependent spatial readout. Evaluation at coordinates outside the training grid additionally requires output-scaling parameters at those locations, as examined in Section~\ref{sec:generalization}. Architectural specifications, preprocessing and training procedures are provided in Methods and the Supplementary Information.

\subsection{Global and local reconstruction accuracy}
\label{sec:accuracy}

From the histories of twelve neutron monitors, the single-branch TRON models reconstruct the global sea-level effective-dose field with errors of approximately $0.1\%$ against EXPACS over the held-out year 2023. Across observation windows of 7, 30, 60 and 90 days, the single-branch LSTM and GRU variants (S-LSTM and S-GRU) achieve mean relative $L_2$ errors of $0.099\%$ and $0.098\%$, respectively, averaged over the four window lengths (Fig.~\ref{fig:metrics_summary}a). Both are more accurate than their multi-branch counterparts (M-LSTM and M-GRU), whose errors are roughly an order of magnitude higher, while longer observation windows yield no consistent improvement. The static-branch DeepONet, which receives only the final day's counts, gives errors of $0.128$--$0.193\%$. This comparison does not isolate the benefit of temporal encoding: the static model used a shorter early-stopping patience and a fixed learning rate, and its run-to-run spread exceeds its gap to the single-branch variants at $T=30$ days (Methods; Supplementary Table~\SIref{tab:performance}{S3}). All single-branch TRON and static-branch DeepONet runs remain below $0.2\%$, compared with $0.34$--$0.39\%$ for per-point linear regressions. A per-point quadratic function of the mean monitor count rate achieves a lower error of $0.063\%$, exploiting the reference field's dependence on a single time-varying index (Methods; Supplementary Section~\SIref{sec:intrinsic}{S10}). Thus, TRON provides accurate reconstruction within a sequence-to-field formulation, although it is not the most accurate estimator for this particular reference. Unless stated otherwise, subsequent analyses use the S-LSTM model with a 30-day window, whose mean relative $L_2$ error is $0.096\%$; each dose-model configuration was trained once.

The low global errors are accompanied by close agreement in spatial structure and relatively consistent accuracy across the test year. S-LSTM achieves structural similarity index measure (SSIM) values near $0.999$ across the observation windows, although SSIM does not reproduce the accuracy ranking of all models (Fig.~\ref{fig:metrics_summary}b; Methods). Its daily relative $L_2$ errors form narrow, unimodal distributions, while S-GRU has a similar spread except at $T=30$ days, where its distribution is broader; the static-branch DeepONet exhibits the greatest day-to-day spread (Fig.~\ref{fig:metrics_summary}c). The reconstructed fields on the 5th-, 50th- and 95th-percentile error days preserve the reference's large-scale latitudinal structure (Fig.~\ref{fig:metrics_summary}d). Their signed relative errors form persistent bands of alternating sign, reaching magnitudes of $0.22$, $0.29$ and $0.37\%$, respectively. Across all 365 test days, the correlation of each day's error field with the annual-mean error field has a median of $0.99$ and a 5th--95th percentile range of $0.91$--$1.00$. The dominant reconstruction discrepancy is therefore a persistent spatial bias, rather than an error pattern that changes substantially from day to day.

\begin{figure}[p]\thisfloatpagestyle{empty}
    \centering
    \makebox[\textwidth][c]{%
        \includegraphics[width=180mm]{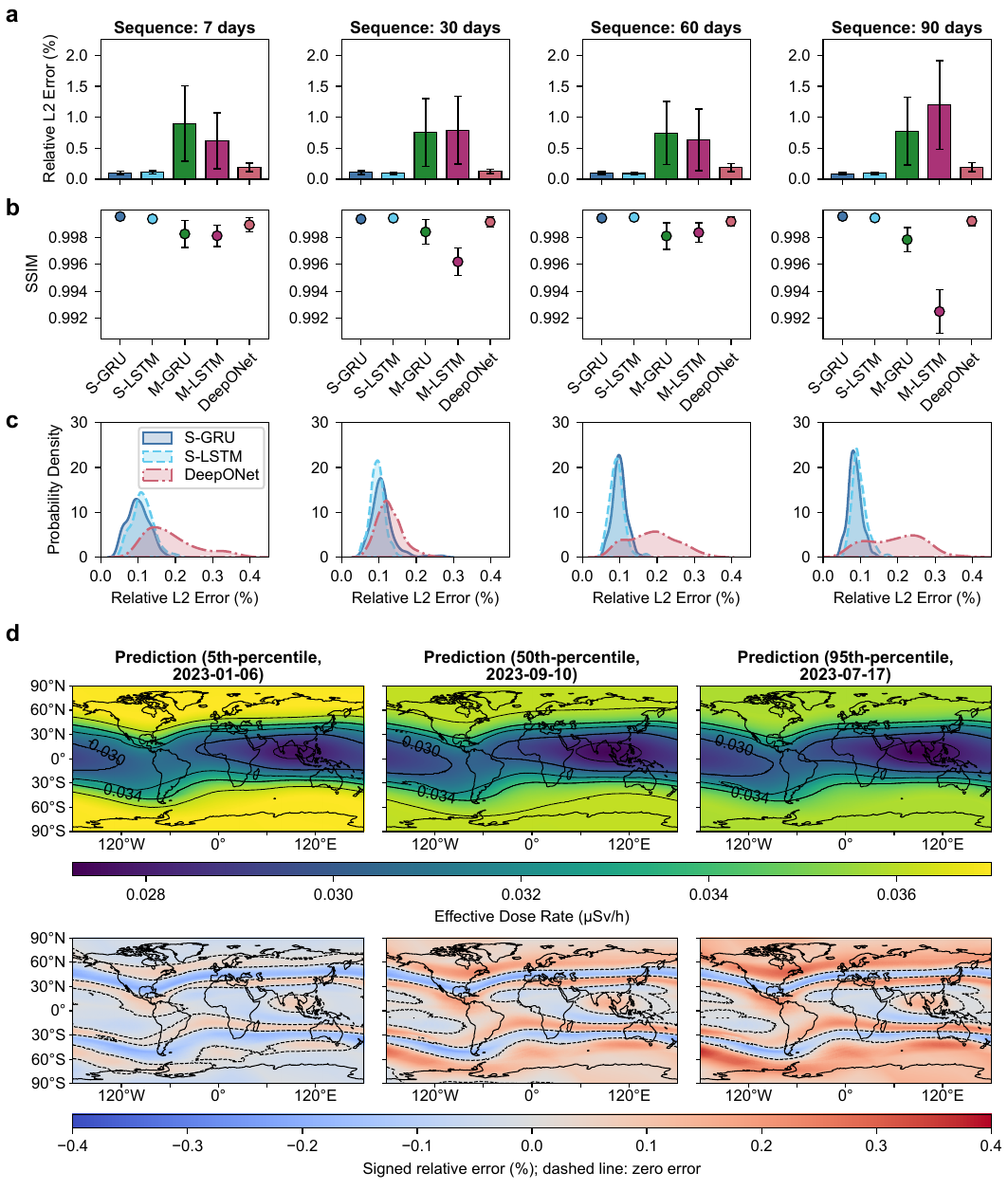}}
\end{figure}
\begin{figure}[!htbp]
\caption{Global reconstruction accuracy across observation windows and model variants.
\textbf{a} Relative $L_2$ error and \textbf{b} structural similarity index measure (SSIM; the y axis does not start at zero) for input windows of 7, 30, 60 and 90 days (columns). Bars in \textbf{a} and points in \textbf{b} show means over the $n=365$ daily fields of the 2023 test year; error bars denote $\pm1$ s.d. across days. Each bar or point represents one trained model. S-GRU and S-LSTM, single-branch TRON with gated recurrent unit or long short-term memory encoders; M-GRU and M-LSTM, corresponding multi-branch variants; DeepONet, deep operator network with a static feed-forward branch.
\textbf{c} Kernel density estimates of daily relative $L_2$ errors for the single-branch variants and static-branch DeepONet; y axes are shared within the row.
\textbf{d} S-LSTM with a 30-day window on the 5th-, 50th- and 95th-percentile error days: predicted fields (top; black isolines every $0.002\,\mu\mathrm{Sv}\,\mathrm{h}^{-1}$) and signed relative errors, $(\mathrm{prediction}-\mathrm{reference})/\mathrm{reference}$, against EXPACS (bottom).}
\label{fig:metrics_summary}
\end{figure}

\begin{figure}[t]
    \centering
    \makebox[\textwidth][c]{%
        \includegraphics[width=180mm]{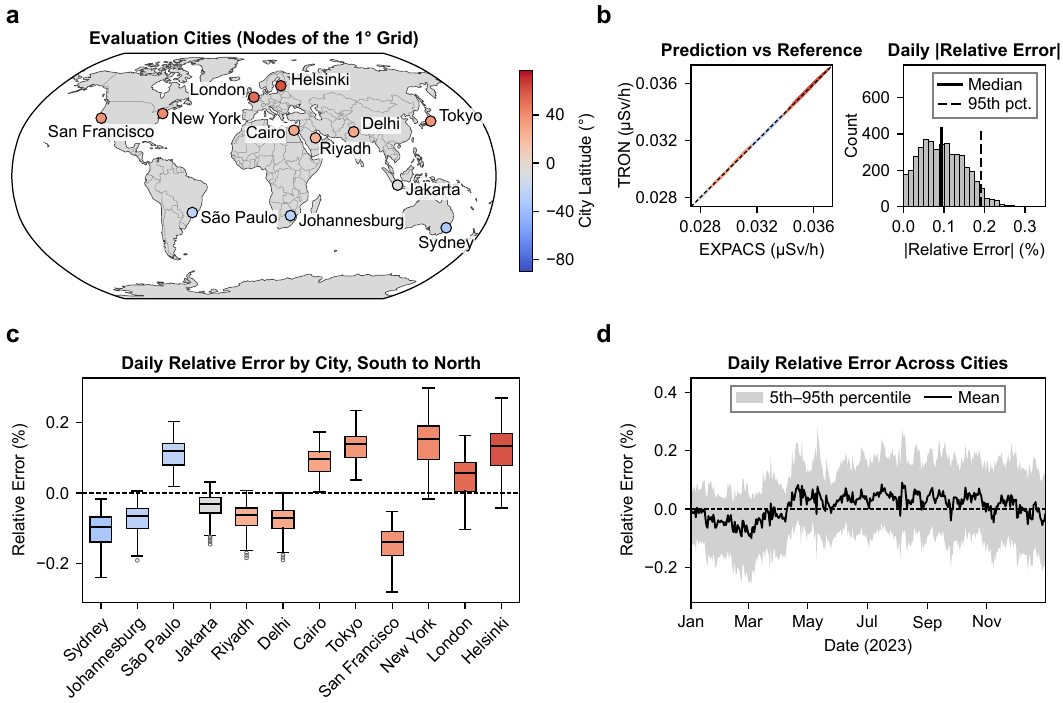}}
\caption{Location-specific dose reconstruction across twelve globally distributed cities.
Results are for S-LSTM with a 30-day input window, evaluated over the 2023 test year using one trained model. Each city is represented by the nearest node of the $1^\circ$ grid, whose 2001--2021 values were training targets.
\textbf{a} City locations, colored by latitude.
\textbf{b} Left, predicted versus EXPACS reference dose rates for all city--day pairs ($n=4{,}380$), colored as in \textbf{a}; dashed line, identity. Right, distribution of absolute relative errors over the same pairs; solid line, median; dashed line, 95th percentile.
\textbf{c} Daily signed relative errors, $(\mathrm{prediction}-\mathrm{reference})/\mathrm{reference}$, for each city ($n=365$ days), ordered from south to north and colored as in \textbf{a}. Center lines indicate medians, boxes the interquartile ranges, whiskers 1.5 times the interquartile range, and points observations beyond the whiskers; dashed line, zero.
\textbf{d} Daily mean signed relative error across the twelve cities (line), with the 5th--95th percentile range across cities (band); dashed line, zero.}
\label{fig:point_pred}
\end{figure}

Evaluation at twelve globally distributed cities shows how this spatial bias translates into location-specific dose estimates (Fig.~\ref{fig:point_pred}a). Each city is represented by its nearest node on the $1^\circ$ training grid; these tests therefore assess reconstruction in the held-out year at fixed locations, rather than generalization to locations excluded from training. Across the 4,380 city--day pairs, the predicted and reference dose rates lie close to the identity line, with a median absolute relative error of $0.094\%$ and a 95th percentile of $0.191\%$ (Fig.~\ref{fig:point_pred}b). Nevertheless, the errors exhibit systematic offsets: median signed errors range from $-0.14\%$ in San Francisco to $+0.15\%$ in New York, and the 5th--95th percentile interval excludes zero for ten of the twelve cities (Fig.~\ref{fig:point_pred}c). These offsets sample the persistent spatial pattern visible in the global error maps. The mean signed error across cities also changes during the year, from approximately $-0.04\%$ in January--March to $+0.04\%$ in May--October (Fig.~\ref{fig:point_pred}d). This temporal drift is smaller than most city-specific offsets, showing that high global accuracy coexists with small but systematic local biases.

\begin{figure}[!htbp]
    \centering
    \makebox[\textwidth][c]{%
        \includegraphics[width=180mm]{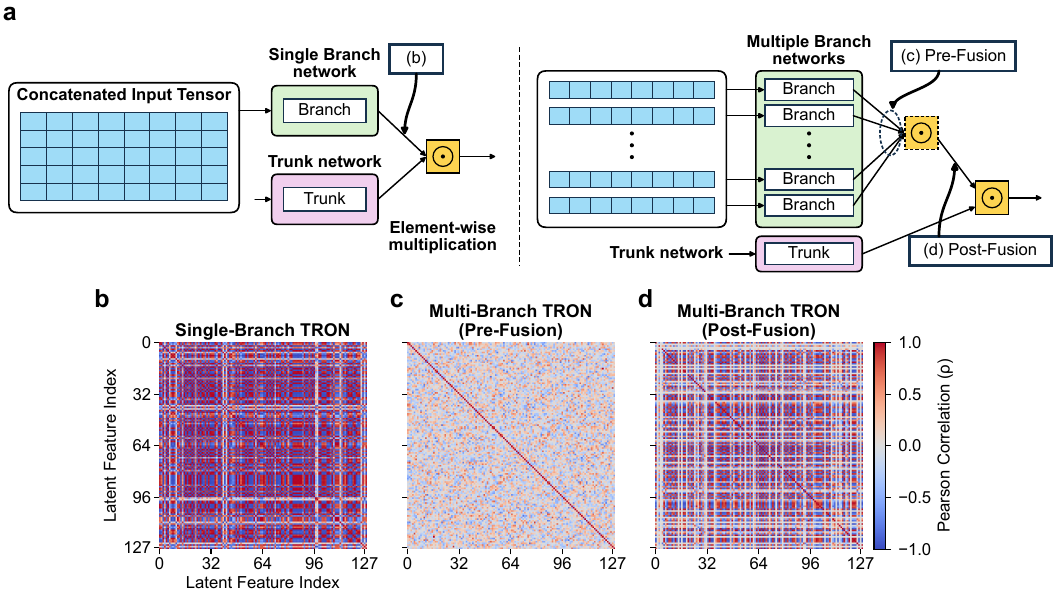}}
\caption{Correlation structure of the learned latent representations.
\textbf{a} Single-branch and multi-branch architectures. Blue cells, input neutron counts; green, branch networks; pink, trunk network; $\odot$, element-wise multiplication. Labels (b)--(d) identify the branch-side tensors analyzed in the corresponding panels, before multiplication with the trunk output; the dashed ring marks the station-specific outputs pooled in \textbf{c}.
\textbf{b}--\textbf{d} Pearson correlations $\rho$ between 128 latent features, with the mean $|\rho|$ over off-diagonal pairs:
\textbf{b} single-branch S-LSTM, $0.814$, calculated over $n=365$ test days;
\textbf{c} multi-branch M-LSTM before fusion, $0.246$, calculated by pooling the twelve station outputs over $n=4{,}380$ station--days;
\textbf{d} M-LSTM after fusion, $0.522$, calculated over $n=365$ test days, with the 16 features constant over the year assigned $\rho=0$.
Both models use 30-day input windows; one trained model is analyzed per architecture.}
\label{fig:correlations}
\end{figure}

To examine the representations accompanying the architectural performance differences, we compared the branch-side latent features of the 30-day S-LSTM and M-LSTM models (Fig.~\ref{fig:correlations}; Supplementary Section~\SIref{sec:latent}{S8}). The single-branch representation exhibits strong correlations among its 128 latent features, with a mean absolute off-diagonal Pearson correlation of $0.814$ over the test year. This co-variation is consistent with the shared variability of the neutron-monitor inputs. In the multi-branch model, the corresponding mean is $0.246$ when the station-specific outputs are pooled before fusion and $0.522$ after their element-wise multiplication, counting the 16 features that remain constant over the year as uncorrelated. The two architectures also differ in how their representations are combined: M-LSTM multiplies the station-specific latent vectors before combining the result with the trunk output, whereas S-LSTM combines its jointly encoded representation directly with the trunk. These observations associate the more accurate single-branch model with a more strongly correlated latent representation. They do not establish a causal explanation for its performance, because the architectures also differ in fusion and parameter count, and the effect of multiplicative fusion on training was not isolated.

\subsection{Computational efficiency}
\label{sec:efficiency}

\begin{figure}[!htbp]
    \centering
    \makebox[\textwidth][c]{%
        \includegraphics[width=180mm]{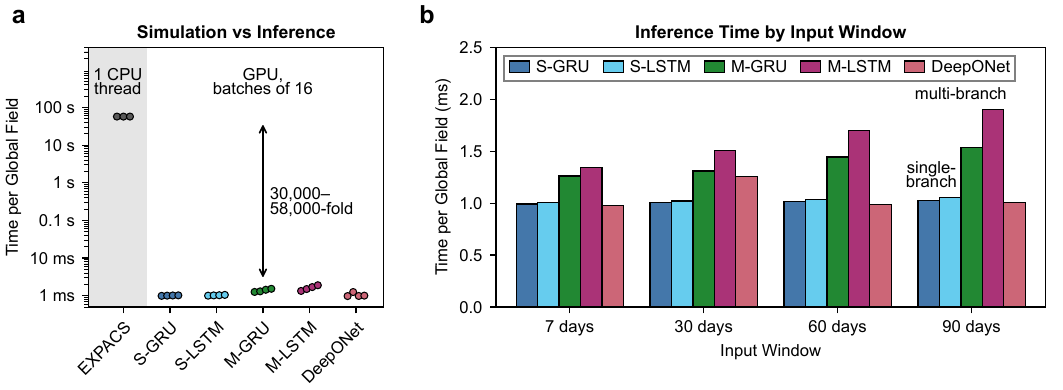}}
\caption{Computational cost of global dose-field reconstruction.
\textbf{a} Time per global field for EXPACS on one CPU thread (AMD Ryzen 9 7950X3D; gray points, three runs) and the trained networks on an NVIDIA A100 GPU in batches of 16, shown on a logarithmic scale. Colored points represent observation windows of 7, 30, 60 and 90 days, from left to right. The arrow indicates the speed-up range across the four TRON variants, including differences in hardware and parallel execution.
\textbf{b} Neural-network inference time as a function of observation-window length. S- and M-, single- and multi-branch TRON; GRU, gated recurrent unit; LSTM, long short-term memory; DeepONet, deep operator network with a static branch. GPU times are the total inference time divided by the number of fields evaluated over the 2023 windows, measured in one timed pass; no error bars are shown. These are average times per field under batching, not independently measured single-field latencies. Numerical values are provided in Supplementary Table~\SIref{tab:speed}{S5}.}
\label{fig:speed}
\end{figure}

TRON reconstructs a global dose field at 65,341 locations in approximately one millisecond per field under batched GPU evaluation (Fig.~\ref{fig:speed}). Across the four TRON variants and observation windows of 7--90 days, inference times range from $0.99$ to $1.90$\,ms, compared with $57.6$\,s for EXPACS to evaluate the same global grid on a single CPU thread. The single-branch architectures combine the lowest reconstruction errors with the shortest inference times among the TRON variants: S-LSTM and S-GRU require $0.99$--$1.05$\,ms per field, with little change as the observation window increases from 7 to 90 days. The multi-branch variants require $1.26$--$1.90$\,ms, consistent with their separate station encoders and additional fusion operations; M-LSTM with a 90-day window is the slowest configuration. Under the evaluated hardware configurations, these timings correspond to runtime reductions of approximately $(3.0\text{--}5.8)\times10^{4}$ across all TRON variants and $(5.5\text{--}5.8)\times10^{4}$ for the single-branch models. Together with the accuracy results, these measurements demonstrate the capacity for rapid, repeated global reconstruction once the model has been trained.

The benchmark compares inference on an NVIDIA A100 GPU, using batches of 16 fields, with EXPACS execution on one thread of an AMD Ryzen 9 7950X3D CPU. The reported acceleration therefore combines the learned model's computational advantage with differences in hardware and parallel execution. GPU times are calculated by dividing the total elapsed inference time by the number of fields evaluated over the 2023 observation windows; they characterize average time per field under batching, rather than the latency of an isolated field request, which was not measured. EXPACS timings are averaged over three runs, and its independent grid-point calculations could also be parallelized. The comparison quantifies evaluation cost after training and excludes the offline cost of reference-field generation and model training. It therefore supports efficient repeated evaluation under the tested configuration, rather than an equivalent reduction in the total cost of developing and deploying the reconstruction model.

\subsection{Evaluation beyond the training grid}
\label{sec:generalization}

\begin{figure}[!htbp]
    \centering
    \makebox[\textwidth][c]{%
        \includegraphics[width=180mm]{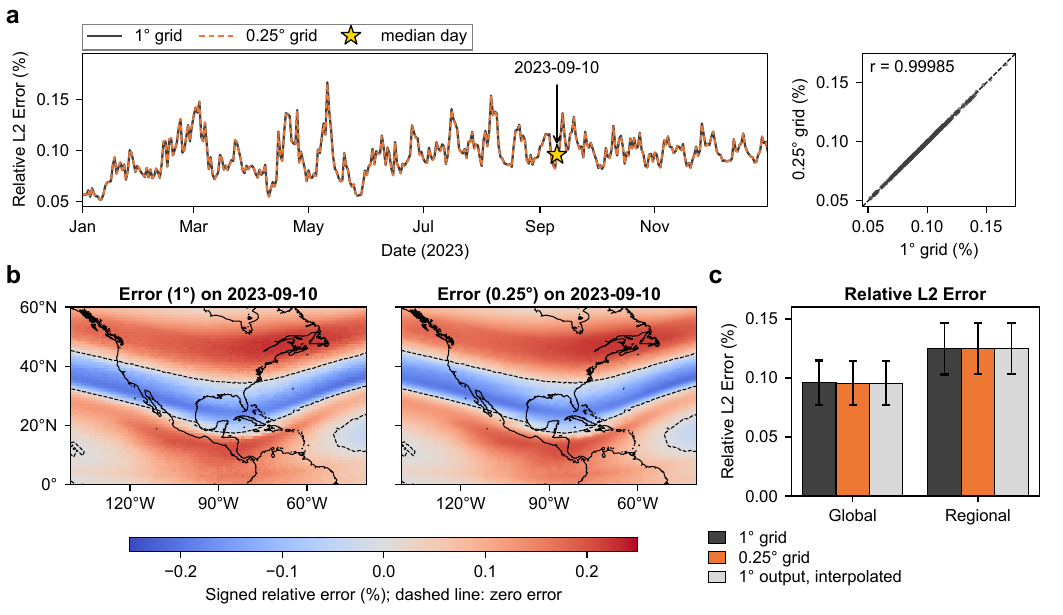}}
\caption{Evaluation on a finer spatial grid without retraining the network.
The S-LSTM model uses a 30-day observation window and was trained on $1^\circ$ EXPACS reference fields; one trained model is evaluated throughout. Errors are calculated against the EXPACS reference at the corresponding resolution.
\textbf{a} Daily global relative $L_2$ errors on the $1^\circ$ and $0.25^\circ$ grids over the 2023 test year. The star marks the median-error day, 10 September 2023. Right, comparison of errors at the two resolutions, with one point per day ($n=365$); dashed line, identity; $r$, Pearson correlation.
\textbf{b} Signed relative errors, $(\mathrm{prediction}-\mathrm{reference})/\mathrm{reference}$, on the representative day over the North America sector ($140^\circ$--$40^\circ$\,W, $0^\circ$--$60^\circ$\,N).
\textbf{c} Relative $L_2$ errors over the global domain and the regional sector shown in \textbf{b}. Light gray bars ($1^\circ$ output, interpolated), the normalized network output bilinearly interpolated from $1^\circ$ to $0.25^\circ$. For both finer-grid evaluations, the per-point output-scaling parameters $(m_j,M_j)$ are fitted using the $0.25^\circ$ EXPACS reference from 2001--2021. Bars show means over $n=365$ test days; error bars denote $\pm1$ s.d. Regional SSIM values are $0.9985\pm0.0006$ at $1^\circ$ and $0.9997\pm0.0001$ at $0.25^\circ$ (mean $\pm$ s.d. over test days).}
\label{fig:zeroshot}
\end{figure}

The coordinate-based trunk allows TRON to be evaluated on a spatial grid different from the one used during training. To examine this capability, we evaluated the 30-day S-LSTM model on a global $0.25^\circ$ grid after training exclusively on $1^\circ$ reference fields, increasing the sampling resolution fourfold in each coordinate direction without updating the network weights (Fig.~\ref{fig:zeroshot}). The location-specific output-scaling parameters $(m_j,M_j)$ were fitted at the finer-grid coordinates using EXPACS reference values from the training years, 2001--2021. Under this protocol, the mean relative $L_2$ error remains $0.096\%$ over the 2023 test year. Daily errors at the two resolutions correlate at $r>0.999$ and differ by at most $0.001$ percentage points (Fig.~\ref{fig:zeroshot}a). The signed error maps also retain the same spatial structure, as illustrated over North America on the median-error day, 10 September 2023 (Fig.~\ref{fig:zeroshot}b). Across the test year, the regional relative $L_2$ error in this sector is $0.125\%\pm0.022\%$ at both resolutions, with an error of approximately $0.13\%$ on the representative day (Fig.~\ref{fig:zeroshot}c). Mean global structural similarity increases from $0.9994$ at $1^\circ$ to $0.9999$ at $0.25^\circ$, but this change does not indicate improved reconstruction: the SSIM windows contain a fixed number of pixels and therefore cover a smaller geographic area on the finer grid.

The finer-grid experiment establishes flexibility in spatial evaluation, while the smoothness of the reference limits what it can establish about recovering additional spatial detail. Interpolating the $0.25^\circ$ reference from its own $1^\circ$ nodes reproduces it with a relative $L_2$ error of only $0.003\%$, indicating little structure below the training-grid scale. Correspondingly, querying the trunk directly at $0.25^\circ$ and bilinearly interpolating its normalized $1^\circ$ output both yield a global error of $0.096\%$ when combined with the finer-grid output scaling (Fig.~\ref{fig:zeroshot}c). When the $1^\circ$ scaling parameters are interpolated instead of refitted, both evaluations give an error of $0.121\%$. The accuracy at new coordinates therefore depends on how the physical output scale is supplied, even though the network itself requires no retraining. For context, the separately generated $1^\circ$ EXPACS dataset, interpolated to $0.25^\circ$, differs from the $0.25^\circ$ EXPACS dataset by $0.064\%$, despite identical nominal settings; the source of this discrepancy remains unresolved. Thus, TRON preserves reconstruction accuracy under finer spatial sampling when suitable scaling parameters are available, but this experiment does not demonstrate recovery of spatial structure unresolved by the training grid.

\subsection{Sensitivity to sensor availability and network adaptation}
\label{sec:sensor_sensitivity}

The accuracy of observation-driven reconstruction depends on how the trained model responds when its input network changes. We examined this dependence by masking neutron-monitor channels at inference, replacing each affected sequence with its training-period mean in normalized space (Fig.~\ref{fig:missing_sensor}a). This removes the channel's time-varying information while retaining values within its training range, although the resulting combination of station inputs may differ from those encountered during training. Losing any one of the twelve stations increases the mean relative $L_2$ error from $0.096\%$ to a median of $0.32\%$ across the twelve outage cases (Fig.~\ref{fig:missing_sensor}c). The increase varies substantially by station: removing NEWK or OULU produces errors of $0.58\%$ and $0.56\%$, approximately six times the intact-network error, whereas removing ATHN or INVK gives $0.11\%$ and $0.12\%$. Only the ATHN outage remains within the intact-network mean error $\pm1$ standard deviation. Nevertheless, every single-station outage retains lower error than the $2.55\%$ obtained by predicting the 2001--2021 mean field, used here as a climatological baseline. Pairwise outages amplify this dependence (Fig.~\ref{fig:missing_sensor}d): the median error across the 66 pairs is $0.61\%$, the largest is $1.08\%$ for simultaneous loss of NEWK and OULU, and only three pairs yield errors at or below $0.30\%$. Every one of the ten most damaging pairs includes NEWK or OULU, identifying these channels as particularly influential for the trained model.

With only one station active and the other eleven masked, reconstruction errors rise to $2.67$--$3.04\%$, approximately thirty times the intact-network value and slightly above the climatological baseline (Fig.~\ref{fig:missing_sensor}b). This loss of skill reflects the behavior of a model trained to receive the complete network; it does not establish that a single monitor contains insufficient information for dose reconstruction. Indeed, separately fitted per-point quadratic functions of individual station count rates achieve errors of $0.22$--$1.08\%$ on the same test year (Supplementary Section~\SIref{sec:intrinsic}{S10}). Nor does a station's predictive accuracy in isolation determine the cost of removing it from TRON: OULU and NEWK are respectively the best and worst single-station predictors under the quadratic fit, yet both are among the most consequential channels to lose. Although the monitor records predominantly share a common modulation signal, the trained network depends unevenly on their joint input. The outage experiments quantify this dependence but do not identify its cause.

We next examined adaptation to an expanded input network by treating the Mexico City monitor, MXCO, as a thirteenth station newly available in 2023 (Fig.~\ref{fig:missing_sensor}e). Its input weights were initialized to zero, so the expanded S-LSTM initially reproduced the twelve-station model. We then adapted a restricted set of parameters using the first 7, 14 or 30 days of 2023 and evaluated the remaining days, keeping all other parameters frozen (Supplementary Section~\SIref{sec:newstation}{S7}). The adapted model gives global relative $L_2$ errors of $0.104$--$0.110\%$, compared with $0.097$--$0.098\%$ for the original model on the corresponding evaluation days. Errors over the North America sector similarly increase from $0.126$--$0.128\%$ to $0.134$--$0.141\%$ (Supplementary Tables~\SIref{tab:adapt_local}{S7} and~\SIref{tab:adapt_world}{S8}). Replacing the added station's observations with their training-period mean leaves the adapted model's error unchanged, indicating that the brief adaptation did not produce a measurable benefit from its time-varying input. This experiment has two qualifications: MXCO does not contribute to the modulation index used to generate the reference, so the target cannot reward information unique to that station, and its input scaling uses the historical 2001--2021 record, which would be unavailable for a genuinely new installation. Under the tested protocol, adding and briefly adapting to a thirteenth channel slightly reduces accuracy. Together, the experiments show that the trained model retains skill over climatology under single- and pairwise-station outages, loses that skill when only one station remains, and does not automatically benefit from an expanded sensor network.

\begin{figure}[p]\thisfloatpagestyle{empty}
    \centering
    \makebox[\textwidth][c]{%
        \includegraphics[width=180mm]{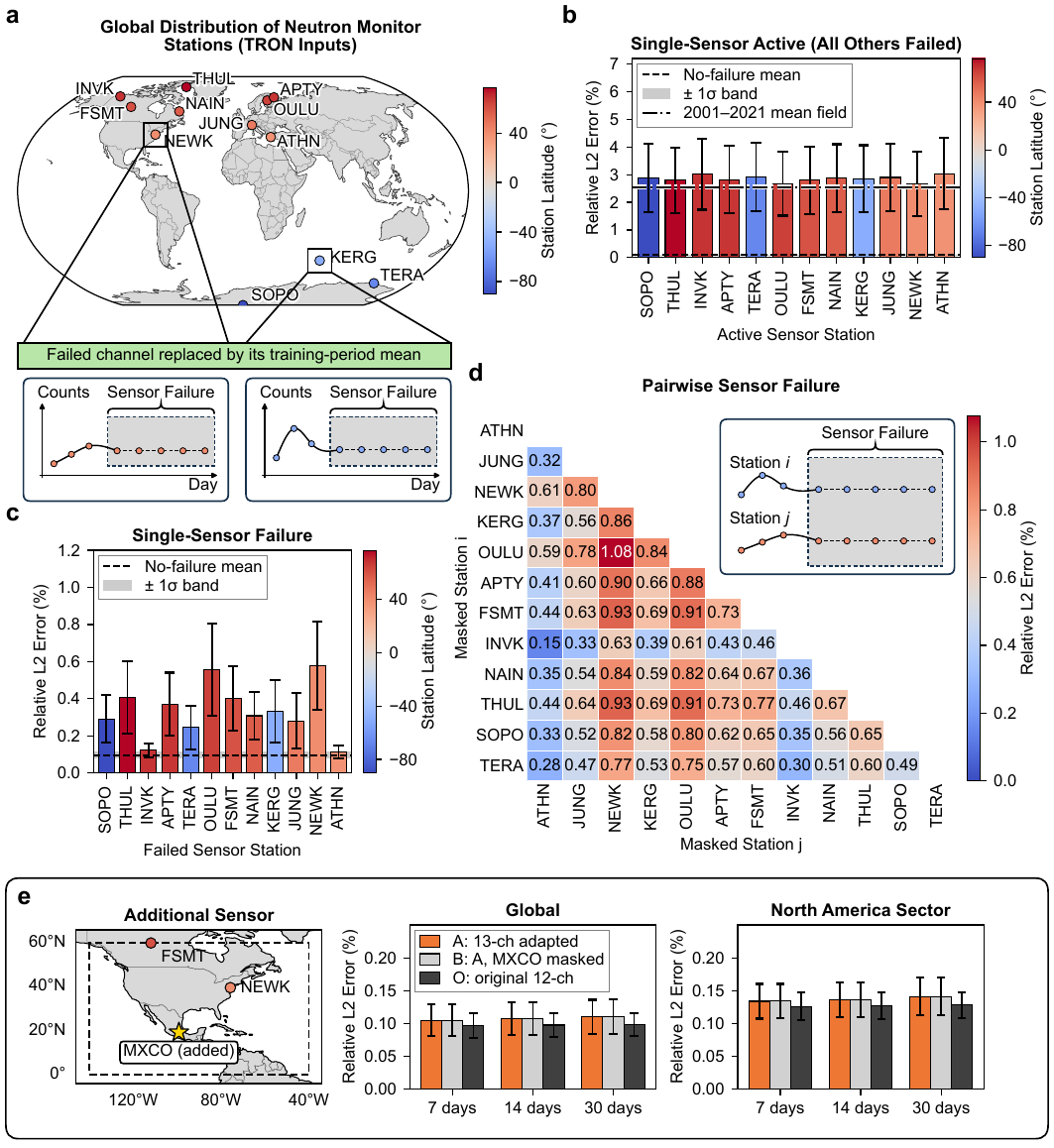}}
\end{figure}
\begin{figure}[!htbp]
\caption{Sensitivity to station outages and adaptation to an additional monitor.
Results use the S-LSTM model with a 30-day input window, evaluated in 2023. A failed station's input is replaced by its training-period mean in normalized space.
\textbf{a} The twelve input stations, colored by latitude; station codes and coordinates are provided in Supplementary Table~\SIref{tab:neutron_stations}{S1}. Below, schematic failures of NEWK and KERG: gray shading indicates the failure period and dashed lines the substituted means.
\textbf{b} Relative $L_2$ error when only one station remains active. The dash-dotted line denotes the $2.55\%$ error of the 2001--2021 mean field (climatology; Supplementary Section~\SIref{sec:intrinsic}{S10}).
\textbf{c} Error when one station is missing.
In \textbf{b} and \textbf{c}, bars show means over $n=365$ test days, error bars denote standard deviations, and the dashed line and gray band show the intact-network mean $\pm1$ standard deviation. Colors follow \textbf{a}.
\textbf{d} Mean error when two stations are missing, evaluated over all 66 pairs; inset, schematic pairwise outage.
\textbf{e} Addition of MXCO (Mexico City, star) as a thirteenth input channel. Left, the North America evaluation sector (dashed box, $140^\circ$--$40^\circ$\,W, $0^\circ$--$60^\circ$\,N). Middle and right, global and regional errors after adaptation on the first 7, 14 or 30 days of 2023. A, adapted 13-channel model; B, adapted model with MXCO replaced by its training-period mean; O, original twelve-station model evaluated on the same days. Error bars denote standard deviations over the remaining $n=358$, 351 or 335 days. One adapted model is evaluated for each adaptation window.}
\label{fig:missing_sensor}
\end{figure}

\subsection{Reconstruction from measured environmental networks}
\label{sec:measured_results}

The radiation experiments assess reconstruction against a simulated reference at locations included among the training targets. To examine reconstruction against independent measurements at locations excluded from training, we applied the same branch--trunk formulation to two environmental station networks (Fig.~\ref{fig:measured}). These experiments address complementary questions: whether sparse temperature observations support reconstruction at held-out stations, whether pressure measurements support reconstruction of temperature as a different physical quantity, and whether observation histories improve reconstruction of a temporally accumulated target. Held-out station values were excluded from the training loss, output scaling and early stopping. Unlike the dose models, the measured-data models use one global output scaling, include elevation among the trunk inputs and are trained with three seeds per configuration. Unless stated otherwise, the inputs also include calendar channels representing the time of year and, for hourly observations, the time of day. We compare single-branch LSTM and GRU encoders with a static-branch DeepONet and a flattened-window DeepONet whose feed-forward branch receives the entire observation window, allowing the contribution of historical information to be distinguished from that of recurrence (Methods; Supplementary Section~\SIref{sec:measured}{S11}).

For daily maximum temperature from the Global Historical Climatology Network (GHCN-Daily)~\cite{menne2012ghcnd,menne2012ghcnddata}, the models reconstruct observations at 100 held-out stations from 14-day histories at 12 input stations, with \GhcnPcNLocations{} other stations supplying training targets in each of three random station splits (Fig.~\ref{fig:measured}a--e). S-LSTM, designated as the primary variant before training on the second and third splits and on the pressure task, achieves a held-out root-mean-square error (RMSE) of \FigFiveCSLstmRmse$\pm$\FigFiveCSLstmRmseSd$\,^\circ$C over nine runs (three splits $\times$ three seeds; mean $\pm$ s.d.). Non-learned interpolation from the latest observations gives larger errors of \GhcnBasePlaneRmse--\GhcnBaseNearestRmse$\,^\circ$C, while a calendar climatology constructed from the held-out stations' own historical records gives \GhcnBaseClimRmse$\,^\circ$C (Fig.~\ref{fig:measured}c). Throughout these comparisons, the RMSE difference is defined as the network's RMSE minus that of its comparator, so negative values favor the network. A stronger two-stage comparator, combining ridge regression at the training stations with a cross-validated kernel decoder, reaches \FigFiveCTwoStageRmse$\,^\circ$C, averaged over one fit per split. S-LSTM's difference from this comparator is unresolved in all three splits when calendar channels are included (Fig.~\ref{fig:measured}d). Across the 24 comparisons covering four branches, calendar-channel inclusion and three splits, 19 are unresolved, three favor the network, two are ties and none meet the criterion for a worse result, using the confidence-interval and $0.10\,^\circ$C practical-difference rules defined in Methods. These counts are not corrected for multiplicity, and all three favorable comparisons occur in the first split (Supplementary Tables~\SIref{tab:meas_ghcnd}{S11} and~\SIref{tab:meas_ghcnd_paired}{S12}). S-LSTM has lower error than the two-stage model at \GhcnDeltaUniqueNNeg{} of the \GhcnUniqueHeldOut{} distinct held-out stations, with a median station-level RMSE difference of $\GhcnDeltaUniqueMedian\,^\circ$C. An exploratory analysis finds a weak negative correlation between this difference and distance to the nearest input station (Spearman $\rho=\GhcnDeltaDistRho$, $p=\GhcnDeltaDistP$; Fig.~\ref{fig:measured}e). The results establish reconstruction at held-out locations more accurately than the non-learned baselines, but no consistent advantage over the two-stage model. Continuous maps illustrate the learned field; accuracy is assessed only at measured stations, and continuous querying is also available to the kernel decoder.

To test reconstruction from an indirect proxy, we used 24\,h histories from 30 barometers to estimate hourly temperature at 40 held-out stations in the Swiss PeakWeather network~\cite{zambon2025peakweather}, with \PeakPcNLocations{} other stations supplying training targets (Fig.~\ref{fig:measured}f,g). No thermometer observations enter the inputs, and the three overlapping barometer draws exclude the held-out locations. Pressure differences between elevations contain information about intervening air temperature through the hypsometric relation, providing a physical basis for the proxy mapping. Comparators, hypotheses and decision rules were pre-registered before training on these data. Across nine runs, S-LSTM achieves an RMSE of \PressRT$\,^\circ$C, compared with \PressHypsRmse$\,^\circ$C for a hypsometric baseline with calendar channels and \PressClockOnlyRmse$\,^\circ$C for a calendar-only linear model. The two-stage model reaches \PressTwoStageRmse$\,^\circ$C, and its comparison with S-LSTM remains unresolved: the RMSE difference is $\PressHTwoDelta\,^\circ$C, with a 95\% confidence interval of [$\PressHTwoLo$, $\PressHTwoHi$]$\,^\circ$C. The magnitude of this estimated difference is smaller than the S-LSTM seed-to-seed standard deviation of \PressSeedSd$\,^\circ$C. S-LSTM nevertheless improves substantially over the same architecture trained on temporally shuffled pressure with calendar channels kept aligned, with an RMSE difference of $\PressHOneDelta\,^\circ$C and a 95\% confidence interval of [$\PressHOneLo$, $\PressHOneHi$]$\,^\circ$C. Pressure therefore contributes information beyond calendar seasonality, although TRON is not shown to extract more of it than the two-stage comparator. The flattened-window DeepONet with calendar channels is worse than that comparator under the pre-registered rule. The relative comparison is more favorable to TRON with pressure than with thermometer inputs, but this largely reflects its poorer performance against the two-stage model on thermometer inputs; complete contrasts and outcomes are reported in Supplementary Tables~\SIref{tab:meas_pressure}{S13} and~\SIref{tab:meas_pressure_tests}{S14}.

A temporally accumulated target provides a direct test of whether access to observation history benefits reconstruction. Using 72\,h windows from 30 thermometers in PeakWeather, we reconstructed heating degree-hours at 40 held-out stations for accumulation periods of 24, 48 and 72\,h (one station split, three training seeds; Fig.~\ref{fig:measured}h,i). Heating degree-hours sum the hourly temperature deficit below $18\,^\circ$C over the specified period. For the 24\,h target, the static-branch DeepONet leaves \DhTwentyFourSFcnClockOnUnexpl\% of the variance unexplained, compared with \DhTwentyFourSLstmClockOnUnexpl\% for S-LSTM; for the 72\,h target, the corresponding values are \DhSeventyTwoSFcnClockOnUnexpl\% and \DhSeventyTwoSLstmClockOnUnexpl\%. These percentages are compared within each target because target variances differ. In contrast, for instantaneous temperature, the static-branch model performs at least as well as S-LSTM (\DhInstantSFcnClockOnUnexpl\% versus \DhInstantSLstmClockOnUnexpl\% unexplained variance). For 24\,h degree-hours, the static model has higher RMSE at \FigFiveDhSFcnWorseN{} of the \FigFiveDhNHeldOut{} held-out stations for the first seed, with the remaining seeds reported in Supplementary Table~\SIref{tab:meas_dh}{S15}. The flattened-window model performs comparably to S-LSTM within the seed spread when calendar channels are included: at 72\,h, their unexplained variances are \DhSeventyTwoSMlpClockOnUnexpl\% and \DhSeventyTwoSLstmClockOnUnexpl\%, respectively. Without calendar channels, the flattened-window model is numerically less accurate (\DhSeventyTwoSMlpClockOffUnexpl\% versus \DhSeventyTwoSLstmClockOffUnexpl\%). These comparisons demonstrate the value of the observation window for accumulated targets, without establishing that recurrence is necessary to obtain that benefit. No linear comparator was evaluated for heating degree-hours.

Together, the measured-network experiments extend the assessment to observations at held-out locations and to a pressure-to-temperature proxy mapping. They establish improvements over the non-learned interpolation, calendar and hypsometric baselines, while comparisons with the two-stage model remain largely unresolved. The held-out stations lie among the training stations, so the experiments assess spatial reconstruction within the sampled networks rather than extrapolation into unmonitored regions; they also do not validate dosimetry, outage robustness or inference speed. Applicability is further bounded by three measured settings with little or no gain: no tested branch outperformed a sensor-free climatology for held-out weekly sea-surface temperature cells~\cite{reynolds2002oisst}; all methods reconstructed SNOTEL snow water equivalent only weakly~\cite{nrcs_snotel}; and instantaneous PeakWeather temperature showed little improvement over an hourly fitted linear temperature--elevation relation. A proposed mooring-wind-to-sea-level-anomaly task using the Tropical Atmosphere Ocean/Triangle Trans-Ocean Buoy Network (TAO/TRITON) was excluded before training because it failed preset input-availability and linear-skill thresholds (Supplementary Table~\SIref{tab:meas_negative}{S16}). The evidence therefore supports the formulation for specific measured reconstruction tasks, with its comparative value depending on the information carried by the observations and the structure of the target.

\begin{figure}[p]\thisfloatpagestyle{empty}
    \centering
    \makebox[\textwidth][c]{\includegraphics[width=180mm]{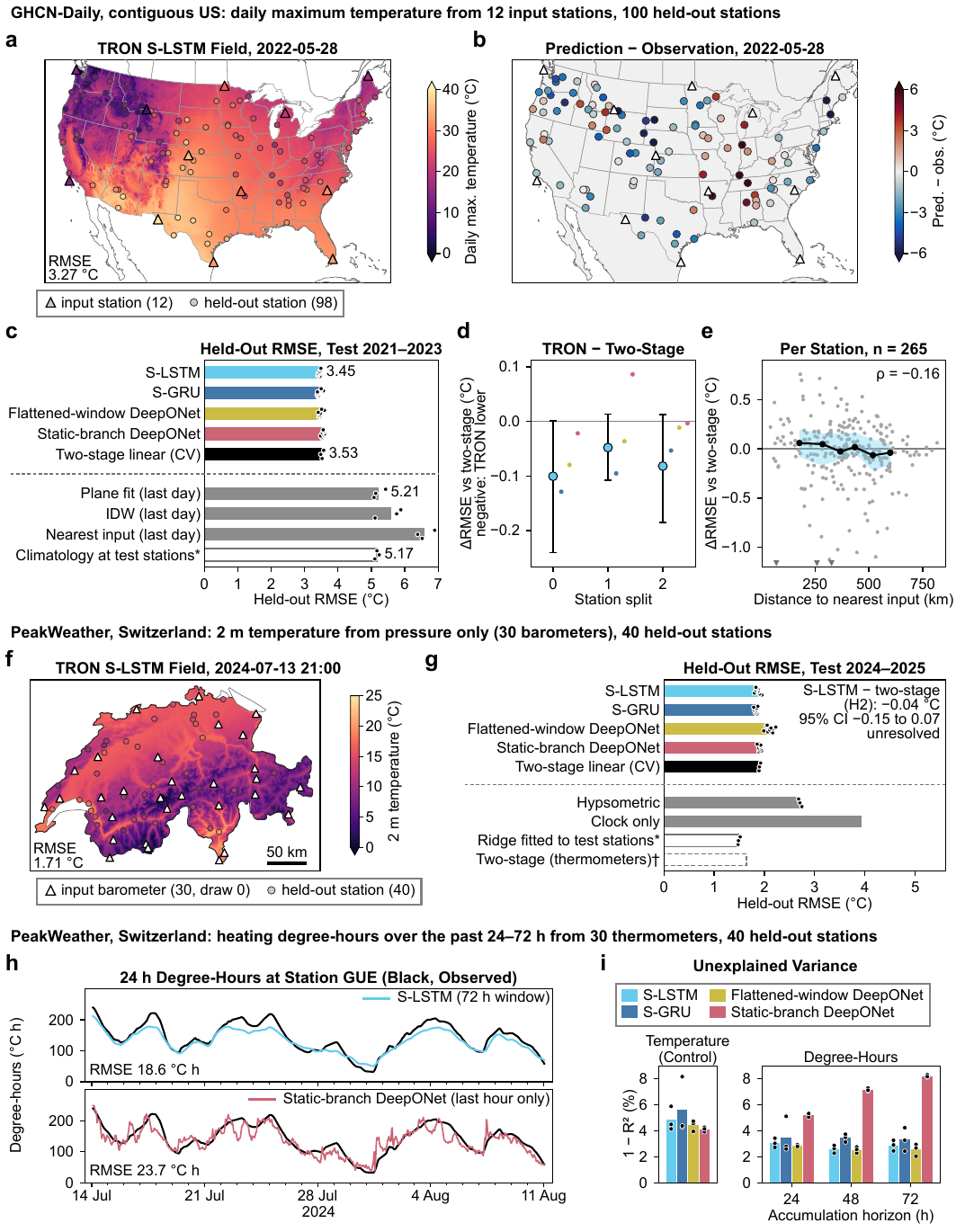}}%
\end{figure}
\begin{figure}[!htbp]
\caption{Reconstruction at held-out stations in measured environmental networks.
S-LSTM and S-GRU, single-branch long short-term memory (LSTM) and gated recurrent unit (GRU) models; DeepONet, deep operator network; two-stage, ridge regression with a cross-validated kernel decoder; IDW, inverse-distance weighting; RMSE, root-mean-square error. Calendar inputs are included unless stated. Bars show means and dots individual runs; splits and draws are numbered 0--2. Asterisks and solid-edged hollow bars indicate use of held-out stations' own records or targets, unavailable in practice.
\textbf{a--e} GHCN-Daily: 14-day windows, three station splits and three seeds per split.
\textbf{a,b} S-LSTM reconstruction and prediction error for split 0 on the median-RMSE test day; maps use a \FigFiveGridStep$^\circ$ grid but are scored only at stations.
\textbf{c} Held-out RMSE; non-learned baselines exclude calendar inputs.
\textbf{d} Network-minus-two-stage RMSE differences with paired-bootstrap 95\% confidence intervals; large markers, S-LSTM; small markers, other branches (colors as in \textbf{c}).
\textbf{e} Exploratory station-level differences versus distance to the nearest input: median and interquartile range by distance sextile; $\rho$, Spearman correlation; $n$, distinct stations; $\blacktriangledown$, clipped values (\GhcnDeltaClippedValues$\,^\circ$C).
\textbf{f,g} PeakWeather pressure-to-temperature reconstruction: three barometer draws, three seeds per draw and one station split. \textbf{f}, S-LSTM for draw 0 at the lower-median-RMSE hour; \textbf{g}, held-out RMSE. H2, pre-registered test of S-LSTM against the two-stage model (Methods); hypsometric, barometer-pair baseline; Clock only, calendar-only model; \textdagger{} and dashed-edged hollow bar, thermometer-driven two-stage model. The latter two models use single deterministic fits.
\textbf{h,i} Heating degree-hours below $18\,^\circ$C: three seeds and one station split.
\textbf{h} Station \FigFiveDhStation{} (G\"utsch; \FigFiveDhStationElev\,m).
\textbf{i} Unexplained variance, $1-R^2$ (\%; $R^2$, coefficient of determination), pooled over stations and times. Bars show S-LSTM, S-GRU, flattened-window and static-branch DeepONet, respectively. Inputs span 72\,h except for the static branch (last hour); compare models within each target.}
\label{fig:measured}
\end{figure}

\section{Discussion}

TRON connects the observation histories of a sparse monitoring network to a global field of a physically defined quantity that the sensors do not measure directly. In the radiation demonstration, a single neural operator reconstructs sea-level effective dose rates at 65,341 locations from twelve neutron monitors, achieving a mean relative $L_2$ error of $0.096\%$ against EXPACS on the held-out year 2023 for the representative 30-day S-LSTM model. Millisecond-scale average inference per field under batched GPU evaluation makes repeated global reconstruction computationally inexpensive after training, with the reported acceleration incorporating differences in hardware relative to single-threaded EXPACS evaluation. The significance of this capability lies in connecting accessible particle measurements to spatially distributed dose estimates through a learned mapping informed by physical reference calculations. Effective dose is a radiological protection quantity defined through tissue-weighted equivalent doses in a reference phantom; dosimeters instead measure operational quantities such as ambient dose equivalent~\cite{icrp2007recommendations}. Accordingly, the radiation results establish agreement with EXPACS/PARMA, a validated computational model~\cite{sato2015analytical} incorporating International Commission on Radiological Protection fluence-to-dose coefficients~\cite{icrp2010conversion}, rather than direct validation against a measured effective-dose field. This distinction defines the role of TRON as a fast, observation-driven reconstruction of the physical reference.

The measured-network experiments address two questions that the radiation demonstration alone cannot resolve: reconstruction of independently measured targets and evaluation at locations excluded from training. The same branch--trunk formulation reconstructs temperature at held-out stations from both thermometer observations and pressure proxies, extending the assessment beyond targets generated by the dose model (Fig.~\ref{fig:measured}). Its improvements over non-learned interpolation, calendar and hypsometric baselines establish useful reconstruction capability in these settings, although the primary variant's comparisons with the purpose-built two-stage linear model remain unresolved. These experiments also clarify when measurement history contributes to reconstruction. Extending the radiation input window from 7 to 90 days yields little change in accuracy, consistent with the reference's dependence on a contemporaneous modulation index. For accumulated heating degree-hours, however, access to the observation window improves reconstruction relative to a branch receiving only the latest snapshot. Comparable performance from the flattened-window model shows that this benefit is attributable to historical information without establishing a unique advantage of recurrence. The value of temporal encoding therefore depends on the target's relationship to the observation history. The temperature fields also contain more varied temporal structure than the dose reference: their first principal components account for $\GhcnPcOneRawPct\%$ and $\PeakPcOneRawPct\%$ of training-period variance in GHCN and PeakWeather, respectively, compared with $\DosePcOnePct\%$ for the dose field (Supplementary Tables~\SIref{tab:intrinsic_pca}{S9} and~\SIref{tab:meas_datasets}{S10}).

The architectural and perturbation experiments distinguish flexibility in spatial output from dependence on the input network. Among the evaluated neural architectures, the single-branch encoders achieve the lowest dose-reconstruction errors at every observation-window length. Their more strongly correlated latent features are consistent with the shared variation of the monitor records, whose first principal component accounts for $95\%$ of the variance (Fig.~\ref{fig:correlations}; Supplementary Section~\SIref{sec:intrinsic}{S10}). A similar advantage of single-branch encoders has been reported for tightly coupled multiphysics problems~\cite{park2026network}. This association does not identify the mechanism responsible for their higher accuracy: the multi-branch models also use different fusion operations and have approximately eleven times as many parameters, while the static-branch comparator was trained with different optimization settings. On the output side, coordinate-based decoding permits evaluation on a finer grid without updating network weights, but the dose estimates still require location-specific scaling. Because the reference contains little structure below the training-grid scale, the finer-grid experiment establishes flexibility in evaluation rather than recovery of additional spatial detail (Fig.~\ref{fig:zeroshot}). On the input side, the trained model retains skill over climatology under single- and pairwise-station outages, but loses that skill when only one station remains active; brief adaptation to an additional station also provides no accuracy gain (Fig.~\ref{fig:missing_sensor}). Thus, a continuous spatial readout does not confer independence from the sensor configuration used during training. Supporting changing networks will require explicit treatment of missing channels and adaptation beyond the protocol evaluated here.

Several limitations determine the scope of these findings. Most importantly, the EXPACS configuration produces a one-parameter family of dose fields in time, with its modulation parameter derived from up to thirteen neutron monitors, twelve of which are TRON inputs (Methods). The radiation benchmark therefore tests whether TRON can reproduce, from twelve of these stations, the mapping implemented by the reference through that index. A per-point quadratic function of the mean monitor count rate exploits this structure and reaches $0.063\%$ error, below the single-branch TRON results (Supplementary Section~\SIref{sec:intrinsic}{S10}). The headline accuracy consequently does not establish superiority over simpler estimators for this reference or recovery of a high-dimensional evolving radiation state. Radiation testing is also restricted to one held-out year, daily observations and sea-level fields. One-day-ahead models do not improve on carrying the current nowcast forward, and no separate evaluation covers Forbush decreases or ground-level enhancements (Supplementary Section~\SIref{sec:forecasting}{S6}); the solar-particle dose contribution associated with the latter is omitted by the PARMA reference used here. Each dose configuration was trained once, limiting conclusions about small architectural differences. The measured-network experiments provide complementary evidence, but their held-out stations lie among the training stations, they do not validate dosimetry, outage robustness or inference speed, and three additional measured settings show little or no gain. Two simulated examples involving heat-exchanger velocity and cylinder-wake vorticity demonstrate application to fields with sharper gradients, but neither includes a comparison model, and one uses test sequences for early stopping, limiting its independence as a generalization assessment (Supplementary Section~\SIref{sec:physics}{S9}).

These findings identify concrete steps toward broader radiation-monitoring applications. Altitude-dependent reference fields are needed to assess aviation exposure, while sub-daily observations and event-specific evaluations are needed to examine rapidly changing conditions. Assessment during ground-level enhancements additionally requires reference calculations that include solar energetic particles or suitable independent measurements; increasing temporal resolution alone cannot validate a radiation component absent from the target model. Comparisons with measured operational quantities such as ambient dose equivalent would provide an independent dosimetric assessment, with the distinction from effective dose maintained explicitly. Replicated training should quantify variability across initializations, alongside dedicated assessment of predictive uncertainty and methods designed for incomplete or changing sensor networks. Extending the radiation reference beyond a single temporal parameter would further test the formulation under richer physical variability.

More broadly, the inverse-sensing structure set out in the Introduction has a long reach in physics. The 2026 Nobel Prize in Physics, awarded to Francis Halzen for decisive contributions to the IceCube Neutrino Observatory and the discovery of high-energy neutrinos of astrophysical origin~\cite{nobel2026physics}, recognizes a programme in which the quantity of interest is never sensed directly: individual neutrino events are reconstructed from the Cherenkov light their secondaries deposit in a sparse array of optical modules~\cite{aartsen2013icecube,aartsen2017instrumentation}. The parallel with the present work is structural, not one of scale or of evidentiary standing. IceCube established a previously unobserved astrophysical phenomenon; the radiation reconstructions reported here are assessed against a computational reference, and the measured-network reconstructions against station observations at held-out sites. What the two share is that the scientifically relevant observable need not be the quantity the sensors record. For learned proxy-to-field reconstruction, the consequential next step is therefore not lower surrogate error but independent evidence that sparse indirect observations can render an inaccessible physical quantity reliably observable in space and time. Within the scope demonstrated here, TRON converts sparse proxy histories into rapidly queryable spatial fields, which is a prerequisite for that goal rather than a demonstration of it.

\section{Methods}
\subsection{Simulation of reference dose rates using EXPACS}
Reference effective dose rates were generated using the Fortran version of EXPACS 4.14 (solar modulation data through 21 March 2024)~\cite{sato2016analytical,expacs,sato2015analytical}, compiled and executed by the authors to simulate daily radiation fields at sea-level atmospheric depth. EXPACS estimates cosmic ray fluxes and effective dose rates by applying the PARMA model, which parameterizes primary and secondary particle interactions with the atmosphere under geomagnetic and solar modulation.

Simulations were performed daily from January 1, 2001, through December 31, 2023 (8,400 days), at sea level on the 1$^{\circ}$ grid described in Results (65,341 points). The atmospheric depth was set to 1033\,g\,cm$^{-2}$ to represent sea-level pressure. Over elevated terrain, the field is therefore that at sea-level depth, not at the local ground. The water fraction in the ground, which affects only the neutron spectrum, was set to 0.15 at every grid point. This is the value EXPACS recommends when the local value is unknown.

EXPACS accounts for Earth's magnetic field effects by automatically calculating the vertical geomagnetic cut-off rigidity at each location using its built-in MAGNETOCOSMICS database \cite{desorgher2005atmocosmics}. Each simulation was tagged with the actual date (year, month, day) to capture temporal variations in solar activity, enabling EXPACS to incorporate the appropriate solar modulation parameter $W$ (derived from neutron monitor data) for that day. The $W$-index used by EXPACS is a global scalar obtained from neutron monitor count rates through the affine relation $W_s = u_{1,s} C_s + u_{2,s}$, averaged across stations~\cite{sato2015analytical}. Since version 4.05 (2018), EXPACS evaluates the daily index from the count rates, on that date, of up to 13 selected neutron monitors~\cite{expacs}, twelve of which, all except McMurdo, are TRON's inputs. The modulation index is therefore determined almost entirely by our input stations, and Supplementary Section~\SIref{sec:intrinsic}{S10} shows that it can be reconstructed from them. Under the configuration adopted here -- sea level, fixed atmospheric depth, and the time-invariant cut-off rigidity map -- the reference field is a deterministic function of the static cut-off rigidity and of this single scalar, $X(\mathbf{r},t) = F(R_{\mathrm{c}}(\mathbf{r}), W(t))$, so the temporal variability of the dataset carries one degree of freedom (Supplementary Section~\SIref{sec:intrinsic}{S10}). Unlike the per-point quadratic function of the mean count rate (Results), which is built on this structure, TRON uses no index and no functional form chosen for it, and reproduces the mapping from the station series to about $0.1\%$.

The dose rate calculations included contributions from a comprehensive set of particles: neutrons, protons, muons (positive and negative), electrons, positrons, photons, and ions from helium up to nickel. These encompass both primary cosmic rays and secondary particles produced by atmospheric cascades. The output of each EXPACS simulation is an effective dose rate (\textmu Sv\,h$^{-1}$) at sea-level atmospheric depth.

\subsection{Implementation of TRON}
The branch network in TRON is designed using either a single-branch or multi-branch architecture, as illustrated in Supplementary Fig.~\SIref{fig:blocks}{S3}.

In the single-branch configuration (Supplementary Fig.~\SIref{fig:blocks}{S3}a), count-rate time series from all $S$ monitoring stations are concatenated at each time step into a single $T \times S$ input matrix (with $T$ the sequence length and $S=12$ stations). This multivariate sequence is processed by a stacked recurrent network (either LSTM or GRU) with 4 layers of 128 hidden units each. We denote these single-branch models as S-LSTM or S-GRU, respectively (see Table \ref{tab:model-architecture}). The final hidden state at the last time step is normalized via layer normalization and passed through a fully connected layer to produce a 128-dimensional latent representation of the temporal features.

\begin{table}[htbp]
\centering
\caption{Architectural specifications of TRON variants. S-GRU and S-LSTM denote TRON models with single-branch GRU and LSTM encoders, respectively. M-GRU and M-LSTM denote TRON models with multi-branch GRU and LSTM encoders, where each sensor is processed independently. $d$, latent dimension shared by the branch and trunk outputs.}
\label{tab:model-architecture}
\begin{tabular}{@{}cccccccc@{}}
\toprule
Model  & Type  & \# Branches & Input Size  & Hidden Dim. & \# layers & $d$ & \# parameters \\ \midrule
S-GRU  & GRU   & 1                  & 12         & 128         & 4      & 128 & 401,921     \\
S-LSTM & LSTM  & 1                  & 12         & 128         & 4      & 128 & 519,169     \\
M-GRU  & GRU   & 12                 & 1 (per sensor) & 128    & 4      & 128  & 4,404,865    \\
M-LSTM & LSTM  & 12                 & 1 (per sensor) & 128    & 4      & 128  & 5,794,945    \\ 
\bottomrule
\end{tabular}%
\end{table}

In the multi-branch configuration (Supplementary Fig.~\SIref{fig:blocks}{S3}b), each station's time series is encoded independently by its own LSTM or GRU sub-network. This design (denoted M-LSTM or M-GRU for 12 branches) yields $S$ separate 128-dimensional latent vectors, one per sensor, which are then fused by element-wise multiplication across all stations, following the multiple-input operator network MIONet~\cite{jin2022mionet}.

The trunk network is common to all TRON variants and encodes spatial query points. Given a location (latitude, longitude), the trunk uses a fully connected network of three linear layers (two hidden layers of 128 units, with ReLU activations) to transform the coordinates into a 128-dimensional spatial embedding. This embedding provides a continuous representation of location and is dimension-matched to the branch output.

For baseline comparison, we implemented a static-branch DeepONet (labeled DeepONet in Figs.~\ref{fig:metrics_summary} and~\ref{fig:speed}) using a feed-forward branch network. In this static model, the branch takes a snapshot of sensor readings (the current values from all 12 stations, i.e., a 12-dimensional input vector) instead of a sequence. It then passes this through a fully connected network of four linear layers, $12 \to 128 \to 128 \to 128 \to 128$, with ReLU activations after the first three. The trunk network in this baseline is identical to the one in TRON (the same coordinate encoder), so the two differ in the branch. The baseline was, however, trained with different early-stopping and learning-rate settings (see below), so the comparison is not fully controlled. This static-branch DeepONet approximates the original operator-network formulation.

The branch and trunk outputs are combined as in Section~\ref{sec:architecture} (Supplementary Fig.~\SIref{fig:blocks}{S3}c). Table \ref{tab:model-architecture} summarizes the architectural specifications of the TRON variants, including the number of branch networks, input sizes, hidden dimensions, and total trainable parameters for each model.

All TRON models were trained using a mean squared error (MSE) loss on the per-point normalized dose. For a batch of $B$ samples, each evaluated at the $P$ query points, the loss is

\begin{equation} \mathcal{L} = \frac{1}{BP}\sum_{k=1}^{B}\sum_{j=1}^{P}\big(\hat{z}^{(k)}_{j} - z^{(k)}_{j}\big)^2, \quad z^{(k)}_j = \frac{X^{(k)}_j - m_j}{M_j - m_j}, \end{equation}

with $B = 16$ the batch size and $P = 65{,}341$; all reported metrics are computed after the inverse transform, on dose rates. For the 0.25$^{\circ}$ evaluation (Section~\ref{sec:generalization}), the network weights are unchanged and $m_j$, $M_j$ are the minimum and maximum of the 2001--2021 0.25$^{\circ}$ reference at each grid point. We trained the networks using the Adam optimizer at an initial learning rate of $1\times10^{-3}$ and weight decay $1\times10^{-6}$, halving the learning rate when the validation loss had not improved for more than five epochs, with shuffled mini-batches of size 16. To prevent overfitting, early stopping was employed based on validation loss, with a patience of 10 epochs and a cap of 1000 epochs (i.e., training was halted if the validation error did not improve for 10 consecutive epochs). The static-branch DeepONet used a patience of 5 epochs and a fixed learning rate. Each configuration was trained once from a fixed seed (1234567 for S-LSTM, 12345 for S-GRU and the static baseline, and 1234 for the multi-branch variants); the tabulated dose-field results are therefore single runs without seed replication.

Model performance was evaluated on a held-out test set using three complementary error metrics: root mean square error (RMSE), mean absolute error (MAE) and relative $L_2$ error. These metrics provide a comprehensive assessment of prediction accuracy, capturing both absolute error magnitudes and the relative error normalized by true dose values. Each metric is computed for the field of each test day over its $P$ points and then averaged over the $N_{\mathrm{d}} = 365$ test days of 2023, where $\hat{X}^{(i)}_j$ and $X^{(i)}_j$ are the predicted and reference dose rates at point $j$ on test day $i$. RMSE is the square root of the mean squared prediction error:

\begin{equation} \text{RMSE} = \frac{1}{N_{\mathrm{d}}}\sum_{i=1}^{N_{\mathrm{d}}} \sqrt{\frac{1}{P}\sum_{j=1}^{P} \Big(\hat{X}^{(i)}_j - X^{(i)}_j\Big)^2}\,, \end{equation}
which penalizes larger errors more heavily. MAE is the mean of absolute errors:
\begin{equation} \text{MAE} = \frac{1}{N_{\mathrm{d}}}\sum_{i=1}^{N_{\mathrm{d}}} \frac{1}{P}\sum_{j=1}^{P} \Big|\hat{X}^{(i)}_j - X^{(i)}_j\Big|\,, \end{equation}
providing an easily interpretable average error in the same units as the target. 

The relative $L_2$ error quantifies the prediction error as a fraction of the true field's energy norm:
\begin{equation} \text{relative }L_2 = \frac{1}{N_{\mathrm{d}}}\sum_{i=1}^{N_{\mathrm{d}}} \frac{\|\hat{\mathbf{X}}^{(i)} - \mathbf{X}^{(i)}\|_2}{\|\mathbf{X}^{(i)}\|_2}\,, \end{equation}
where $\|\cdot\|_2$ denotes the Euclidean norm of the dose field (in practice, each sample here is a global map unrolled into a vector).

In addition to error-based metrics, we evaluated structural similarity using the Structural Similarity Index (SSIM). For each daily prediction, the dose field vectors were first reshaped into 2D latitude-longitude grids ($181 \times 361$ for $1^\circ$ resolution). SSIM was then computed over these spatial images to assess perceptual quality by comparing local luminance, contrast, and structure patterns:

\begin{equation}
\text{SSIM}(x, y) = \frac{(2\mu_x\mu_y + c_1)(2\sigma_{xy} + c_2)}{(\mu_x^2 + \mu_y^2 + c_1)(\sigma_x^2 + \sigma_y^2 + c_2)}
\end{equation}

where $\mu_x$ and $\mu_y$ are local means, $\sigma_x^2$ and $\sigma_y^2$ are variances, $\sigma_{xy}$ is covariance, and $c_1 = (0.01L)^2$, $c_2 = (0.03L)^2$ are stabilization constants, with $L$ the data range, the maximum minus the minimum of the reference field of that day; the local statistics are computed in a $7 \times 7$ uniform window (scikit-image defaults). SSIM ranges from $-1$ to $1$ (perfect similarity). SSIM does not rank the models by accuracy here: the per-point linear regressions, less accurate than every single-branch and static-branch model, have higher SSIM (Supplementary Table~\SIref{tab:performance}{S3}).

Two supplementary benchmarks apply the formulation to simulated fields: (1) reconstruction of the interior velocity field of a compact heat exchanger from its boundary conditions, and (2) field reconstruction of a fluid system governed by the Navier-Stokes equations. The problem setups and results are given in Supplementary Section~\SIref{sec:physics}{S9}.

\subsection{Measured-data experiments}
GHCN-Daily daily maximum temperature~\cite{menne2012ghcnd,menne2012ghcnddata} (contiguous United States, 2003--2023) and hourly PeakWeather MeteoSwiss data~\cite{zambon2025peakweather} are used; Supplementary Table~\SIref{tab:meas_datasets}{S10} lists the datasets, station counts and splits, and Supplementary Section~\SIref{sec:measured}{S11} gives the station selection, gap handling, window assignment, training recipe and the details of Fig.~\ref{fig:measured}. In each of 3 GHCN station splits, 12 input stations are chosen by farthest-point sampling, 100 held-out stations are drawn at random and the rest are training targets. PeakWeather uses 40 held-out stations and 30 inputs: barometers away from the held-out sites for pressure (3 draws), and thermometers for degree-hours. Windows are split in time. For GHCN, training runs to 2018, validation covers 2019--2020 and the test period 2021--2023. For PeakWeather, training runs to 2022, validation covers 2023 and the test period starts in 2024, or on 31 December 2023 for degree-hours (Supplementary Table~\SIref{tab:meas_datasets}{S10}).

The four branches (S-LSTM, S-GRU, the static-branch DeepONet reading the last step, and a flattened-window DeepONet whose feed-forward branch reads the whole window) and a trunk on longitude, latitude and elevation have 128 units per layer. No per-point output scaling exists at a held-out station, so the target uses one global min--max range from the training stations over the training period. Calendar channels (sine and cosine of the day of year and, for PeakWeather, of the hour) are included unless stated. Each configuration is trained with three seeds, a larger batch and a longer early-stopping patience than the dose models (Supplementary Section~\SIref{sec:measured}{S11} and Table~\SIref{tab:meas_runs}{S17}).

The two-stage linear model fits a ridge regression from the input window (with calendar harmonics when the network has calendar channels) to the training stations and decodes to held-out coordinates with a kernel decoder in longitude, latitude and scaled elevation, whose family and settings are chosen by cross-validation grouped by training station on the training period, so held-out values are read only for scoring. The non-learned, calendar-only and hypsometric baselines are described in Supplementary Section~\SIref{sec:measured}{S11}; rows that use held-out data are references only.

Held-out RMSE differences between a trained network (averaged over seeds, and over draws for pressure) and the two-stage model are tested by a paired bootstrap over held-out stations and calendar weeks, resampled jointly with the same indices for both models (2000 replicates, 95\% percentile intervals). Each contrast is classified as better, worse, a tie or unresolved with a band of $0.10\,^\circ$C, as defined in Results. For pressure, the comparators, this rule and the hypotheses were pre-registered before any network was trained on that dataset. In Supplementary Table~\SIref{tab:meas_pressure_tests}{S14}, which lists all pre-registered tests, the tests reported in Results are labeled H1 (TRON against the time-shuffled pressure control), H2 (TRON against the two-stage model), H2b (TRON's difference from the two-stage model, pressure against thermometer inputs) and H4 (thermometer inputs at least as accurate as pressure). H3 is a pre-set floor on the RMSE of the shuffled runs that guards against leakage. Its outcome was a \PressHThreeOutcome. For GHCN the rule was applied per split by analogy, without pre-registration.

\backmatter

\bmhead{Data availability}
The ground-level neutron monitor count rates used as model inputs are publicly available from the Neutron Monitor Database (NMDB, \url{https://www.nmdb.eu}). The reference effective dose fields were generated with the EXPACS/PARMA toolkit (\url{https://phits.jaea.go.jp/expacs/}) under the settings given in Methods, and can be regenerated from that toolkit. The processed neutron monitor time series and the reference effective dose fields used for training and testing ($1^\circ$ and $0.25^\circ$ grids, 2001--2023) are archived at Zenodo (\href{https://doi.org/10.5281/zenodo.18406046}{doi:10.5281/zenodo.18406046}); further derived datasets are available at \url{https://github.com/kkazuma19/CosmicRays-Operator}. The measured station data are public: GHCN-Daily daily maximum temperature~\cite{menne2012ghcnddata} from the NOAA Open Data Dissemination archive (\url{https://noaa-ghcn-pds.s3.amazonaws.com}; CC0-1.0), and PeakWeather~\cite{zambon2025peakweather} MeteoSwiss station data from \url{https://huggingface.co/datasets/MeteoSwiss/PeakWeather} (CC-BY-4.0, attribution to MeteoSwiss). Station elevations for the maps of Fig.~\ref{fig:measured} are from ETOPO1~\cite{amante2009etopo1}, obtained from the NOAA CoastWatch ERDDAP server (dataset etopo180). The datasets reported as negative results are NOAA OI SST v2~\cite{reynolds2002oisst}, SNOTEL~\cite{nrcs_snotel}, TAO/TRITON daily winds (\url{https://data.pmel.noaa.gov/pmel/erddap/tabledap/pmelTaoDyW}) and the Copernicus Marine Service global L4 sea-level anomaly (\href{https://doi.org/10.48670/moi-00148}{doi:10.48670/moi-00148}). The prepared measured-data arrays and station lists, together with the per-run metrics, model checkpoints and held-out predictions, are archived in a later version of the same Zenodo record (\href{https://doi.org/10.5281/zenodo.23395344}{doi:10.5281/zenodo.23395344}).

\bmhead{Code availability}
All code used to train and evaluate TRON, to run the baseline architectures, and to produce the figures and quantitative results reported in this work, including the analysis underlying Supplementary Section~\SIref{sec:intrinsic}{S10} and the measured-data training, baseline and analysis scripts (\texttt{code/analysis/realdata/}) and the script for Fig.~\ref{fig:measured} (\texttt{code/figures/v5/fig8\_measured\_v3style.py}), is available at \url{https://github.com/kkazuma19/CosmicRays-Operator}. The code and data of the GNO and GINO baselines (Supplementary Section~\SIref{sec:baselines}{S4}) and of the additional physics problems (Supplementary Section~\SIref{sec:physics}{S9}) are shared through the data folder linked from the repository README.

\bibliography{sn-bibliography}

\bmhead{Acknowledgments}
S.B. Alam thanks Prof. Eyad Abed for valuable discussions that helped shape aspects of the broader research perspective underlying this work, particularly concerning learning and sensing in dynamic energy systems. These discussions were associated with research directions subsequently supported through the NSF CAREER Award No. ECCS-2543177. S.B. Alam also acknowledges support from the U.S. DOE Distinguished Early Career Award DOE DE-NE009635. S.B.A. and K. Kobayashi also gratefully acknowledge Prof. Rizwan Uddin. 

This research is a part of the Delta research computing project, which is supported by the National Science Foundation, (award OCI 2005572) and the State of Illinois, as well as the Illinois Computes program supported by the University of Illinois Urbana-Champaign and the University of Illinois System.

We acknowledge the NMDB database (www.nmdb.eu), founded under the European Union's FP7 programme (contract no. 213007) for providing data. Additionally, we express our gratitude to the institutions and observatories that maintain and operate the individual neutron monitor stations, whose invaluable contributions made this work possible. Athens neutron monitor data were kindly provided by the Physics Department of the National and Kapodistrian University of Athens. Jungfraujoch neutron monitor data were kindly provided by the Physikalisches Institut, University of Bern, Switzerland. Newark/Swarthmore, Fort Smith, Inuvik, Nain, and Thule neutron monitor data were provided by the University of Delaware Department of Physics and Astronomy and the Bartol Research Institute. Kerguelen and Terre Adelie neutron monitor data were kindly provided by Observatoire de Paris and the French Polar Institute (IPEV), France. Oulu neutron monitor data were obtained from the Sodankylä Geophysical Observatory, University of Oulu, Finland, via \url{http://cosmicrays.oulu.fi}. Apatity neutron monitor data were kindly provided by the Polar Geophysical Institute of the Russian Academy of Sciences (PGI, Apatity). South Pole neutron monitor data were provided by the University of Wisconsin, River Falls. Mexico City neutron monitor data were kindly provided by the Cosmic Ray Group, Geophysical Institute, National Autonomous University of Mexico (UNAM), Mexico.

We acknowledge the GTMBA Project Office of NOAA/PMEL for the TAO/TRITON data. This study has been conducted using E.U. Copernicus Marine Service Information; \url{https://doi.org/10.48670/moi-00148}.

During the preparation of this work, the authors used a large language model solely for language editing and refinement. The authors reviewed and edited the content as needed and take full responsibility for the content of the publication.

\bmhead{Funding}
S.B. Alam acknowledges support from the U.S. Nuclear Regulatory Commission under grant numbers 31310025M0012 and 31310024M0041, NSF CAREER Award No. ECCS-2543177, and the U.S. DOE Distinguished Early Career Award DOE DE-NE009635. Jay Yoo acknowledges support from the U.S. DOE Office of Nuclear Energy's Nuclear Energy University Program (NEUP), Award DOE DE-NE0009076. S. Chakraborty acknowledges the Fulbright-Nehru Academic and Professional Excellence Fellowship for supporting his visit to the University of Illinois Urbana-Champaign.

\bmhead{Author contributions}
S.B.A. and K.K. contributed to the conceptualization of the study. Methodology was developed by K.K., S.B.A., D.A. and S.K.\@ Software was implemented by K.K. and T.T.\@ Formal analysis was carried out by K.K., T.T. and S.R.\@ Resources were provided by D.A., S.K. and S.B.A.\@ The original draft was written by K.K., T.T., S.C. and S.B.A.\@ Writing, review and editing were performed by K.K., J.P.Y., S.C. and S.B.A.\@ Visualization was created by K.K. and S.R.\@ S.C. and S.B.A. supervised the study. Funding was acquired by S.B.A.

\bmhead{Competing interests}
The authors declare no competing interests.

\clearpage

\setcounter{section}{0}
\setcounter{subsection}{0}
\setcounter{figure}{0}
\setcounter{table}{0}
\setcounter{equation}{0}
\renewcommand{\thesection}{S\arabic{section}}
\renewcommand{\thesubsection}{S\arabic{section}.\arabic{subsection}}
\renewcommand{\thefigure}{S\arabic{figure}}
\renewcommand{\thetable}{S\arabic{table}}
\renewcommand{\theequation}{S\arabic{equation}}

\begin{center}
{\LARGE\bfseries Supplementary Information\par}
\vspace{0.6em}
{\large Spatiotemporal Reconstruction of Global Radiation from Sparse Sensor Sequences\par}
\end{center}

\vspace{1.2em}

\section{List of Neutron Counters}
\label{sup:1}
The neutron counting data analyzed in this study were obtained from the Neutron Monitor Database (NMDB) \cite{mavromichalaki2011applications,nmdb}, spanning the period from January~1,~2001, to December~31,~2023. Table~\ref{tab:neutron_stations} summarizes the geographical and instrumental characteristics of the monitoring stations included in this dataset.

The NMDB archive distributes neutron monitor count rates from a global network of ground-based counters. We used the daily mean count rates as distributed by NMDB, which are corrected for barometric pressure and detector efficiency; this compensates for atmospheric and instrument-specific effects at each station and makes each station's series comparable over time. No further correction or temporal averaging was applied.

Data gaps were filled with Piecewise Cubic Hermite Interpolating Polynomial (PCHIP) interpolation, which avoids overshoot between the end points of a gap. Two stations have long filled stretches: SOPO from 22 November 2005 to 3 February 2010 (1{,}535 days) and TERA from 31 December 2005 to 1 January 2011 (1{,}828 days, interrupted by two measured days on 12--13 December 2006) and from 31 March to 1 June 2020 (63 days). Shorter stretches were filled at ATHN (27 days in 2015) and KERG (97 days in 2016). The long stretches lie in the training period. Stretches of more than two weeks were also filled at ATHN in 2005 (23 days) and 2016 (21 days), at TERA in January 2016 (20 days) and at NAIN in 2017 (18 days), all in the training period, and gaps of up to two weeks at most stations. ATHN also has filled stretches in the validation year, from 24 to 31 January 2022 (8 days), and in the test year, from 7 to 13 February, from 10 to 20 April and from 14 to 18 December 2023 (7, 11 and 5 days); here, as above, a stretch runs from the last measured day before a gap to the first measured day after it. Because PCHIP interpolates between measured values on both sides of a gap, the inputs on the 17 filled test days (all at ATHN) depend on measurements made up to ten days later, which a real-time system would not have. In 2022, gaps of one to three days were also filled at ATHN, KERG, OULU and INVK. Within them the input varies smoothly between the end points and does not contain the variations recorded by the other stations over the same period. NMDB gives INVK daily means of zero on 17 and 18 September 2011 and reduced values on 16 and 19 September, because hourly data are missing from 16 September 21:00 to 19 September 21:00 UTC; these four values are missing data rather than measurements, and they were used unchanged (training period).

\begin{table}[htbp]
\caption{List of neutron monitoring stations used in this study. Lat., latitude (positive north); Lon., longitude (positive east), in degrees. $R_{\mathrm{c}}$: vertical geomagnetic cut-off rigidity from the NMDB station list~\cite{nmdb}. JUNG is the IGY-type monitor at Jungfraujoch.}
\centering
\begin{tabular}{lllrrrl}
\toprule
\textbf{Station Code} & \textbf{Location} & \textbf{Country} & \textbf{Lat.} & \textbf{Lon.} & \textbf{$R_{\mathrm{c}}$ (GV)} & \textbf{References} \\
\midrule
ATHN     &  Athens         &  Greece      & $37.97$ & $23.78$ & $8.53$ & \cite{mavromichalaki2001athens, mavromichalaki2005new, SOUVATZOGLOU2009728} \\
JUNG     &  Jungfraujoch   &  Switzerland & $46.55$ & $7.98$ & $4.49$ & \cite{FLUCKIGER20091155} \\
NEWK     &  Newark         &  USA         & $39.68$ & $-75.75$ & $2.40$ & \cite{newk} \\
KERG     &  Kerguelen      &  France      & $-49.35$ & $70.25$ & $1.14$ & \cite{kerg} \\
OULU     &  Oulu           &  Finland     & $65.05$ & $25.47$ & $0.81$ & \cite{kananen1991quarter} \\
APTY     &  Apatity        &  Russia      & $67.57$ & $33.40$ & $0.65$ & \cite{balabin2015upgrade} \\
FSMT     &  Fort Smith     &  Canada      & $60.02$ & $-111.93$ & $0.30$ & \cite{fsmt} \\
INVK     &  Inuvik         &  Canada      & $68.36$ & $-133.72$ & $0.30$ & \cite{invk} \\
NAIN     &  Nain           &  Canada      & $56.55$ & $-61.68$ & $0.30$ & \cite{evenson2005neutron} \\
THUL     &  Thule          &  Greenland   & $76.50$ & $-68.70$ & $0.30$ & \cite{evenson2005neutron} \\
SOPO     &  South Pole     &  Antarctica  & $-90.00$ & $0.00$ & $0.10$ & \cite{bieber2007long} \\
TERA     &  Terre Adélie   &  Antarctica  & $-66.65$ & $140.00$ & $0.01$ & \cite{tera} \\
\bottomrule
\end{tabular}
\label{tab:neutron_stations}
\end{table}

\section{Sequential Dataset Preparation}
\label{sec:dataset}

The original neutron counting dataset, covering the period from January 1, 2001, to December 31, 2023, consisted of daily neutron counts from 12 monitoring stations, resulting in a total of 8400 days of observations. The daily mean count rates were used as distributed by NMDB, corrected for barometric pressure and detector efficiency, with gaps filled as described in Section~\ref{sup:1}.

The sliding window technique was employed to convert the original time series data into sequential datasets suitable for training and evaluation. This approach generates overlapping input sequences from the original data, enabling the model to learn temporal dependencies effectively. Each input sequence corresponds to a specific time window of length $T$, while the target value represents the effective dose rate at the final time step of the sequence.

The sliding window process was performed as follows: given an input dataset of daily neutron counts $\mathbf{y}(1), \mathbf{y}(2), \dots, \mathbf{y}(N)$, where $\mathbf{y}(t) \in \mathbb{R}^{S}$ holds the counts of the $S$ stations on day $t$, and a target dataset of effective dose rate fields $\mathbf{X}(1), \mathbf{X}(2), \dots, \mathbf{X}(N)$, the sliding window function iterates over the data, forming sequences of length $T$ from the neutron count series while associating each sequence with the corresponding effective dose rate at the final time step. Specifically, the input sequence at time step $i$ is defined as:

\begin{equation}
    \mathbf{Y}^{(i)} = [\mathbf{y}(i), \mathbf{y}(i+1), \dots, \mathbf{y}(i+T-1)]^{\top}
\end{equation}
and the corresponding target value is:

\begin{equation}
    \mathbf{X}(i+T-1)
\end{equation}
After sequentialization, the resulting input tensor has the shape:

\begin{equation}
    \text{Input shape: } [N', T, S]
\end{equation}
where $N'$ is the number of generated sequences after applying the sliding window method, $T$ is the length of the sequence (7, 30, 60, or 90 days), and $S$ is the number of monitoring stations (12). The corresponding target tensor has the shape:

\begin{equation}
    \text{Target shape: } [N', P, 1]
\end{equation}
where $P$ denotes the number of spatial points (65,341) corresponding to the global grid. The sequential datasets were prepared with four different window sizes (7, 30, 60, and 90 days), resulting in varying numbers of training, validation, and test samples. Detailed information about the dataset sizes for each window size is provided in Table \ref{tab:datasets}.

\begin{table}[htbp]
\caption{Number of sequences for each dataset and sequence length after applying the sliding window approach. Each sequence corresponds to a contiguous temporal segment of daily neutron counts across 12 monitoring stations, with target fields of size 65{,}341 grid points.}
\centering
\begin{tabular}{@{}lcccc@{}}
\toprule
\multirow{2}{*}{Data Set} & \multicolumn{4}{c}{Sequence Length ($T$)} \\ \cmidrule(l){2-5} 
                          & 7       & 30      & 60     & 90     \\ \midrule
Training                  & 7664    & 7641    & 7611   & 7581   \\
Validation                & 365     & 365     & 365    & 365    \\
Test                      & 365     & 365     & 365    & 365    \\ \bottomrule
\end{tabular}
\label{tab:datasets}
\end{table}

\section{Seasonality analysis of the grid-point mean effective dose}

To assess whether an annual cycle in the data could influence the performance of different input sequence lengths, we examined the seasonality of the simulated global effective dose over the full 2001–2023 period. For each day $t$ we first computed the grid-point mean effective dose
\begin{equation}
D(t) = \frac{1}{P} \sum_{j=1}^{P} X_j(t),
\end{equation}
where $X_j(t)$ denotes the effective dose at grid point $j$ and $P = 65{,}341$ is the number of spatial points. This yields a daily time series $D(t)$ for 23 years. The slow modulation associated with the solar cycle was removed by applying a 365-day running mean to $D(t)$,
\begin{equation}
\tilde{D}(t) = \mathrm{mean}\big(D(t-182),\ldots,D(t+182)\big),
\end{equation}
with a minimum of 180 available days at the edges. We then defined the detrended anomaly
\begin{equation}
D'(t) = D(t) - \tilde{D}(t),
\end{equation}
which isolates shorter-term variability about the slowly varying background.

\begin{figure}[htbp]
\centering
\makebox[\textwidth][c]{\includegraphics[width=180mm]{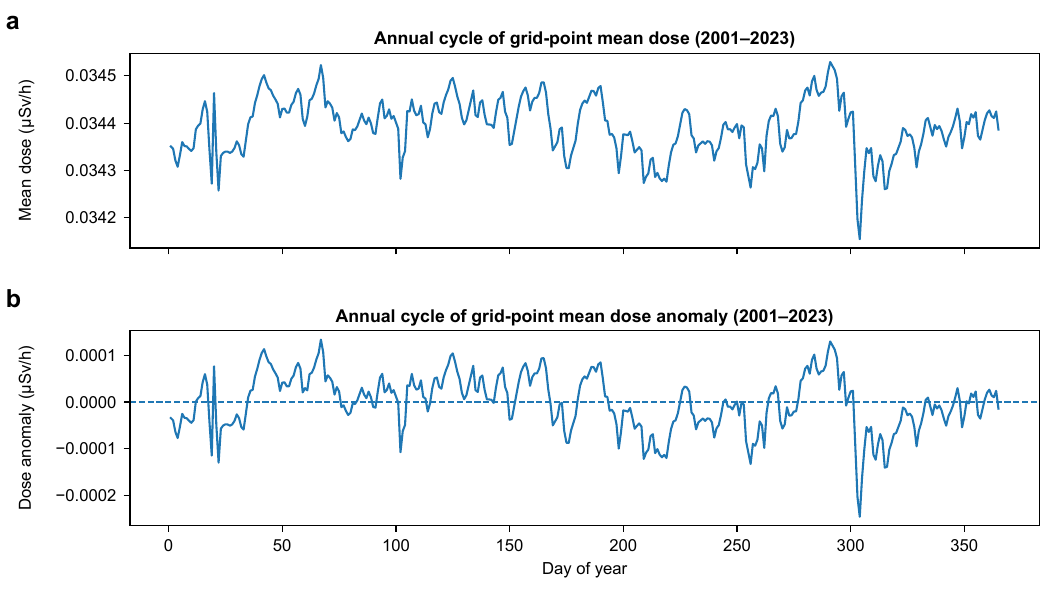}}%
\caption{Climatological annual cycle of the simulated grid-point mean effective dose over 2001–2023. 
\textbf{a} Daily grid-point mean dose obtained by averaging the effective dose field over all $65{,}341$ grid points and then averaging by calendar day (day-of-year). 
\textbf{b} Corresponding anomaly after removal of a 365-day running mean, with the dashed line indicating zero. }
\label{fig:annual_cycle}
\end{figure}

To construct a climatological annual cycle, we removed 29 February from leap years and grouped the remaining dates by their calendar day of year, $\mathrm{doy}$; in leap years the dates after 28 February therefore carry an index one higher than in other years and 31 December is excluded, which has a negligible effect on the averages. For both the raw series $D(t)$ and the anomaly $D'(t)$ we then averaged across all years for each $\mathrm{doy}$,
\begin{equation}
\bar{D}(\mathrm{doy}) = \frac{1}{N_{\mathrm{y}}} \sum_{y=1}^{N_{\mathrm{y}}} D_y(\mathrm{doy}),
\qquad
\bar{D}'(\mathrm{doy}) = \frac{1}{N_{\mathrm{y}}} \sum_{y=1}^{N_{\mathrm{y}}} D'_y(\mathrm{doy}),
\end{equation}
where $N_{\mathrm{y}}$ is the number of years and the subscript $y$ denotes the restriction of the time series to year $y$. The resulting climatological mean dose $\bar{D}(\mathrm{doy})$ and anomaly $\bar{D}'(\mathrm{doy})$ are shown in Fig.~\ref{fig:annual_cycle}a and b, respectively. The grid-point mean dose remains nearly constant at approximately $3.44 \times 10^{-2}$\,\textmu Sv\,h$^{-1}$ throughout the year, and the anomaly exhibits only small fluctuations of order $10^{-4}$\,\textmu Sv\,h$^{-1}$ without a smooth, repeatable winter–summer pattern.

\begin{figure}[htbp]
\centering
\includegraphics[width=120mm]{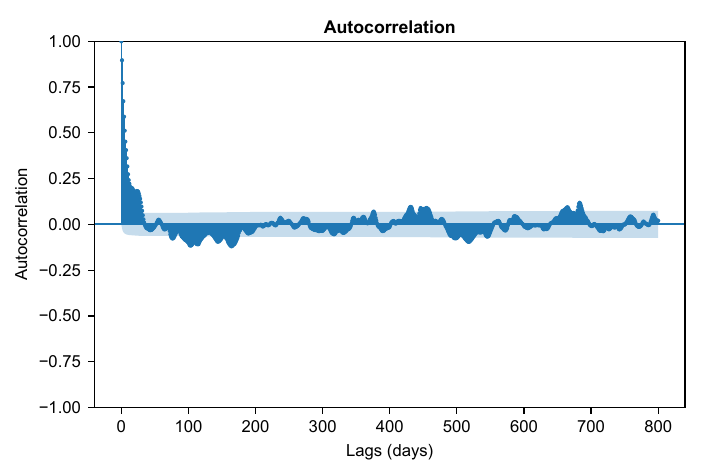}%
\caption{Autocorrelation function of the grid-point mean effective dose anomaly $D'(t)$ for the period 2001–2023, shown for lags up to 800 days. The shaded band denotes the 95\% confidence interval. The autocorrelation first enters the confidence bounds at a lag of 32 days and exhibits no secondary peak near a one-year lag, indicating short-to-medium-term temporal correlations and the absence of a strong annual seasonal cycle.}
\label{fig:acf}
\end{figure}

We further quantified temporal structure by computing the autocorrelation function of the anomaly $D'(t)$ over lags up to 800 days (Fig.~\ref{fig:acf}). The autocorrelation decays rapidly and first falls within the 95\% confidence bounds at a lag of 32 days; at longer lags it stays within about $\pm 0.12$, crossing the bounds in a few lag ranges, and there is no distinct secondary peak at a lag of one year. This indicates that the global effective dose field is governed by short-to-medium-term temporal correlations and slow multi-year heliospheric modulation rather than by a strong annual seasonal cycle. Consequently, the modest performance differences observed between the input sequence lengths considered in the main text (7, 30, 60, and 90 days) are not attributable to a classical yearly seasonality in the data.

\newpage
\section{Baseline Models}
\label{sec:baselines}
\subsection{Linear Regressions (Static and Temporal)}

To establish classical regression baselines, we implement two linear regression variants that differ in their treatment of temporal information.

\subsubsection{Static Linear Regression}

The static linear regression baseline learns independent linear models for each of the 65,341 spatial locations using only the current day's neutron monitor observations. For a given test day with neutron count vector $\mathbf{y}(t) = (Y_1(t), \dots, Y_{12}(t)) \in \mathbb{R}^{12}$, the predicted effective dose rate at spatial location $j$ is computed as:

\begin{equation}
    \hat{X}_j(t) = \gamma_0^{(j)} + \sum_{s=1}^{12} \gamma_s^{(j)} Y_s(t)
\end{equation}
where $\gamma_0^{(j)}$ is the intercept and $\{\gamma_s^{(j)}\}_{s=1}^{12}$ are the learned regression coefficients specific to location $j$. Each location thus learns a separate mapping from the 12-dimensional neutron count space to a scalar dose prediction, resulting in a total of $13 \times 65{,}341 = 849{,}433$ parameters. Ridge regularization with $\alpha = 1.0$ is applied.

\subsubsection{Temporal Linear Regression}

The temporal linear regression baseline extends the static approach by incorporating the full $T$-day temporal window of neutron monitor observations. For an input sequence $\mathbf{Y}(t-T+1:t) \in \mathbb{R}^{T \times 12}$, the sequence is flattened into a single feature vector $\tilde{\mathbf{y}}(t) \in \mathbb{R}^{12T}$ by concatenating all timesteps:

\begin{equation}
    \tilde{\mathbf{y}}(t) = [\mathbf{y}(t-T+1)^\top, \mathbf{y}(t-T+2)^\top, \ldots, \mathbf{y}(t)^\top]^\top
\end{equation}

The predicted dose at location $j$ is then computed as a weighted linear combination of all $12T$ features:

\begin{equation}
    \hat{X}_j(t) = \gamma_0^{(j)} + \sum_{k=1}^{T} \sum_{s=1}^{12} \gamma_{k,s}^{(j)} Y_s(t-T+k)
\end{equation}
where $\gamma_{k,s}^{(j)}$ represents the contribution of station $s$ at temporal offset $k$ to the dose prediction at location $j$. With $T=30$ days and 12 stations, this yields $(360 + 1) \times 65{,}341 = 23{,}588{,}101$ parameters. Ridge regularization with $\alpha = 1.0$ is applied.

Both models are trained independently for each spatial location using ordinary least squares with ridge penalty, minimizing the mean squared error between predicted and reference dose fields over the training period (2001--2021).

\subsection{DeepONet}

DeepONet \cite{lu2021learning} is used as a static operator-learning baseline and provides a mathematically grounded reference against which the sequence branch of TRON is compared. In its classical form, DeepONet approximates a nonlinear operator
\begin{equation}\label{eq:nonlin_map}
    \mathcal{G} : u(x) \mapsto v(y),
\end{equation}
where the input function \(u\) is represented through its evaluations at a finite set of points, and the output function \(v\) is queried at arbitrary coordinates \(y\). The branch network encodes the sampled values of \(u\) into a latent vector
\begin{equation}
    \mathbf{b} = \mathcal{B}\!\left(u(x_1),u(x_2),\dots,u(x_S)\right) \in \mathbb{R}^d,
\end{equation}
while the trunk network embeds the spatial coordinate \(y\) into
\begin{equation}
    \mathbf{t} = \mathcal{T}(y) \in \mathbb{R}^d.
\end{equation}
The predicted output is obtained through the standard DeepONet fusion
\begin{equation}
    \hat{v}(y) = \sum_{i=1}^{d} b_i\, t_i + \beta,
\end{equation}
where \( \beta \) is a learnable scalar.

In adapting DeepONet to the cosmic-ray inverse reconstruction problem, the model receives the same multivariate sequential input tensor used by the single-branch TRON architecture. Each training example consists of a temporal window of length \(T\) from all \(S=12\) neutron monitors:
\begin{equation}
    \mathbf{Y} = \{ Y_s(\tau) \}_{\substack{s=1,\dots,12 \\ \tau = t-T+1,\dots,t}} \in \mathbb{R}^{T \times 12}.
\end{equation}
Instead of processing this tensor sequentially, the static-branch DeepONet applies a fully connected branch network \(\mathcal{B}\) independently at each time step of the window and retains only the encoding of the most recent step. Writing \(\mathbf{y}(\tau) \in \mathbb{R}^{12}\) for the vector of station readings on day \(\tau\), the branch output used for fusion is
\begin{equation}
    \mathbf{b} = \mathcal{B}\!\left( \mathbf{y}(t) \right) \in \mathbb{R}^{d}, \qquad d = 128,
\end{equation}
so that the effective branch input is the 12-dimensional snapshot of the stations on the last day of the window. Here \(\mathcal{B}\) is the network of four linear layers, $12 \to 128 \to 128 \to 128 \to 128$, described in the main-text Methods. The preceding \(T-1\) steps are passed through the same layers but are discarded before fusion with the trunk, so this formulation eliminates any mechanism for modeling ordering, recurrence, or causal time dependence, and gives the baseline no access to the history within the window.

The trunk encoder mirrors the one used in TRON. Each spatial query coordinate
\begin{equation}
    \mathbf{r} = (\mathrm{lat},\mathrm{lon})
\end{equation}
is mapped to a spatial latent embedding
\begin{equation}
    \mathbf{t} = \mathcal{T}(\mathbf{r}) \in \mathbb{R}^{128}.
\end{equation}
The predicted normalized dose at \(\mathbf{r}\) is then given by
\begin{equation}
    \hat{z}(\mathbf{r}) = \sum_{i=1}^{128} b_i\, t_i(\mathbf{r}) + \beta,
\end{equation}
rescaled with the per-point output scaling $(m_j, M_j)$ as in the main-text Methods ($\hat{X}_j = m_j + (M_j - m_j)\hat{z}_j$).
The baseline is trained on the same data and splits as TRON, with early stopping after 5 epochs without improvement of the validation loss and a fixed learning rate of $10^{-3}$.

This static baseline corresponds mathematically to approximating the operator
\begin{equation}\label{eq:operator_snapshot}
    \mathcal{G}_{\text{snapshot}} : \mathbf{y}(t) \mapsto \mathbf{X}(t),
\end{equation}
which depends only on the most recent time step and not on the temporal evolution within the window. It therefore cannot express temporal structure: two windows that differ anywhere except on their final day are indistinguishable under this operator.

\subsection{GNO}
Graph neural operator (GNO) \cite{li2020neural} aims to approximate the nonlinear map in Eq. \eqref{eq:nonlin_map} by constructing a parametric map of the form,
\begin{equation}
    \mathcal{G} : u(x) \times \theta \mapsto v(y),
\end{equation}
with some finite-dimensional parameter space $\theta$. 
The parametric map is represented using an iterative architecture for $j=0, \ldots, J-1$ as
\begin{align}
    b_0(y)& = P\left(x, u(x) \right) \\
    b_{j+1}(y)& = \sigma\left(W_{\phi} b_j(y)+\int_{\Omega} \kappa_\phi \left(y, \xi, u(y), u(\xi) \right) b_j(\xi) \mathrm{d} \xi\right) \label{eq:gkn} \\ 
    v(y)& = Q \left(b_J(y)\right) 
\end{align}
where the ($d+1$)-dimensional input field $(x, u(x))$ is first lifted to an $n$-dimensional vector field using a parametric transformation $P$.
The transformed field $b_0(y)$ is then used as an initialization to the graph kernel network, where $J$ iterative updates are performed. 
In the kernel updates in Eq. \eqref{eq:gkn}, $\sigma$ represents a nonlinear activation operator, while $W_{\phi}$ and $\kappa_\phi$ denote the learnable neural network kernels. 
The final update is then projected to the solution field using a parametric transformation $Q$.

GNO approximates the kernel integration in Eq. \eqref{eq:gkn} by a Monte Carlo sum via a message passing graph network.
For each node $y \in \Omega$, a neighborhood set $N(y)$ is constructed by collecting the nodes within a ball $B(y, h)$ of radius $h$. For each neighbor $\xi \in N(y)$, the graph assigns the edge weight $e(y, \xi)=(y, \xi, u(y), u(\xi))$. 
Given node features $b_j(y) \in \mathbb{R}^n$, edge features $e(y, \xi) \in \mathbb{R}^{n_{\mathrm{e}}}$, and a graph, the iterative updates are rephrased in terms of message passing as
\begin{equation}
    b_{j+1}(y)=\sigma\left(W b_j(y)+\frac{1}{|N(y)|} \sum_{\xi \in N(y)} \kappa_\phi(e(y, \xi)) b_j(\xi)\right)
\end{equation}
where $\kappa_\phi(e(y, \xi))$ is a neural network with edge features as input and a $\mathbb{R}^{n \times n}$ matrix as output. 

In the current cosmic-ray inverse reconstruction problem, the model input is identical to the DeepOnet, where each training example is arranged in $Y \in \mathbb{R}^{T \times 12}$ with temporal window length $T$. 
Since the neutron monitors are sparse compared to the domain of cosmic rays, the data from neutron monitor stations are interpolated to match the shape of the cosmic ray coordinates. Similarly, to the DeepONet, the training examples are processed as static fields, yielding the block $\hat{Y} \in \mathbb{R}^{\hat{T} \times \hat{S}}$. 
For GNO implementation, the \(\hat{T} \times \hat{S}\) block is flattened into a single vector
\begin{equation}
    \tilde{Y}(t) = \mathrm{vec}\!\left( \hat{Y} \right) \in \mathbb{R}^{\hat{T}  \hat{S}},
\end{equation}
so that this baseline approximates the flattened-window operator
\begin{equation}\label{eq:operator}
    \mathcal{G}_{\text{static}} : \tilde{Y}(t) \mapsto X(t),
\end{equation}
which depends on the whole window but not on the order of its time steps.
The prediction of the effective dose rate field for an arbitrary input $\tilde{Y}(t)$ is then constructed at once by the following compositional network
\begin{equation} 
    \begin{aligned}
        \hat{X}(r) &= Q \circ \mathcal{K}_1 \circ \ldots \circ \mathcal{K}_J \circ P\left(t, \tilde{Y}(t) \right) \\
        &\text{with \;\;} \mathcal{K}_1 (\cdot) = \sigma\left(W_{\phi} (\cdot) + \int_{B(y,h)} \kappa_\phi\left(t, \xi, \tilde{Y}(t), \tilde{Y}(\xi)\right) \;(\cdot) \mathrm{d} \xi\right) 
    \end{aligned}
\end{equation}
where $P$ and $Q$ are implemented as a shallow fully connected network with 55 perceptions. For message passing, the number of edge features is taken as 6. The neighborhood information is aggregated using a ball of radius $h=0.1$. With this information, the kernel neural network is modeled as a three-layered dense network with 6, 512, and $512^2$ perceptions in each layer. 
This static baseline approximates the operator in Eq. \eqref{eq:operator}.
Like DeepOnet, this approach is unable to capture temporal structure, as the sequence or ordering of time steps is irrelevant to how the function operates.

\subsection{GINO}
Geometry-Informed Neural Operator (GINO) \cite{li2023geometry} offers an improvement over the GNO architecture for operator learning on arbitrary geometries and mesh discretizations through kernel integration in the spectral domain. 
It approximates the nonlinear map in Eq. \eqref{eq:gkn} using a composition of functions as follows
\begin{equation}\label{eq:gino}
    \mathcal{G} = Q \circ \mathcal{K}_L \circ \ldots \circ \mathcal{K}_1 \circ P
\end{equation}
The transformations $P$ and $Q$ serve the same purpose as GNO; however, they are modeled as message-passing graph networks instead of linear layers. 

Given the input point cloud $\{x_1, \ldots, x_N\}$, $P$ transforms the input features to a function on a uniform latent grid $\{z_1, \ldots, z_S\}$. Given $M$ neighborhood nodes over a ball $B_{h^{\text{in }}}(z)$ of radius $h^{\text{in }}$, the encoded features are computed as
\begin{equation}
    b_0(z) \approx \sum_{i=1}^M k^{in}_{\theta}\left(z, p_i\right) \mu\left(p_i\right)
\end{equation}
where $\mu\left(p_i\right)$ denotes the Riemannian sum weights corresponding to the space of $B_{h^{\text{in }}}(z)$ to be computed by GINO. 
In case of $Q$, a function defined on the uniform latent grid $\{z_1, \ldots, z_S\}$ is transformed to a function on the output point cloud $\{y_1, \ldots, y_N\}$ as
\begin{equation}
    u(y) \approx \sum_{i=1}^M k^{out}_{\theta} \left(y, q_i\right) b_l\left(q_i\right) \mu\left(q_i\right)
\end{equation}
using a ball $B_{r^{\text {out }}}(y)$ of radius $r^{\text{out }}$. The Riemannian weight in this case is $\mu(q_i)=1/S$. 

The integral operators $\mathcal{K}_j: b_{j} \mapsto b_{j+1}$ are defined as
\begin{equation}
    b_{j+1}(z) = \sigma \left(W_{\phi} b_{j}(z) + \int_{\Omega} \kappa_{\phi}(z, \xi) b_{j}(\xi) \mathrm{d} \xi \right)
\end{equation}
where $b_0(z) = P(x,u(x),\varphi(x))$ is the initialization on the latent grid with $\varphi(x)$ being the signed distance function (SDF) and $\kappa_{\phi}$ is a learnable kernel function. 
In the uniform latent grid, the kernel integration is performed as global convolution using element-wise multiplication in Fourier space
\begin{equation}
    b_{j+1}(z) = \sigma \left(W_{\phi} b_{j}(z) + \mathcal{F}^{-1}\left(\kappa_{\phi} \cdot \mathcal{F}(b_j)(z)\right) \right)
\end{equation}
where $\mathcal{F}$ and $\mathcal{F}^{-1}$ are the Fourier and inverse Fourier transforms. The learnable kernel $\kappa_{\phi}$ is parameterized directly by some fixed number of Fourier modes in the Fourier domain.
The final output is obtained as
\begin{equation}
    v(y) = Q(z,b_J(z),\varphi(z))
\end{equation}
where $\varphi(x)$ and $\varphi(z)$ are different SDFs.
The encoder-decoder and the integral kernels are trained end-to-end.

In the cosmic-ray inverse reconstruction problem, the input field is defined on a uniform point cloud consisting of data collected over a temporal window of length \(T\) across \(S=12\) neutron monitoring locations. Thus, each example has a shape of $Y \in \mathbb{R}^{T \times 12}$.
Similar to previous methods, the entire window is treated as a static function field, rather than being processed sequentially. For GINO implementation, the \(T \times 12\) block is flattened into a single vector $\tilde{Y}(t) \in \mathbb{R}^{12T}$.

The uniform latent grid is constructed on
\begin{equation}
    \Omega_z \in [0,1] \times [0,1]
\end{equation}
with $30 \times 30$ grid. 
Given the input coordinate defined by $r = (\mathrm{lon},\mathrm{lat})$, the input neutron information is mapped to the latent space using a ball of radius $h^{\text{in}} = 0.05$. 
At the decoder, the output point cloud is obtained from the latent grid using a ball of radius $h^{\text{out}} = 0.05$. 
Both the encoder and decoder kernels, $k^{in}_{\theta}$ and $k^{out}_{\theta}$, are modeled as two-layered feed-forward networks with 128 and 64 perceptions. 
The kernel integral operator utilizes the first 10 Fourier features to learn the nonlinear mapping in the latent grid. 

Similar to GNO, this approach constructs the effective dose rate field at once, given an arbitrary input $Y$ using the map in Eq. \eqref{eq:gino}.
In analogy to previous methods, this static baseline approximates the operator
in Eq. \eqref{eq:operator} and does not express temporal structure because the order of the time steps plays no functional role. 

\section{Prediction Performance Summary}
\label{sec:performance}
The predictive performance of TRON and baseline architectures was evaluated over the 2023 test period (January–December 2023), following training on data from 2001 to 2021. The TRON variants and the static-branch DeepONet use 2022 for validation and early stopping; the linear regressions use no validation set. Each model was provided with historical neutron monitor sequences of varying temporal lengths (7, 30, 60, and 90 days) and tasked with reconstructing the corresponding daily global effective dose fields at sea level. The results, summarized in Table~\ref{tab:performance}, report the mean accuracy across four evaluation metrics: relative $L_2$ error, root mean square error (RMSE), mean absolute error (MAE), and structural similarity index (SSIM).

Across all sequence lengths, the two single-branch variants (S-GRU and S-LSTM) achieved the lowest relative $L_2$ error, RMSE and MAE, with relative $L_2$ errors below $0.12\%$ and SSIM values above $0.9993$; the linear regressions have higher SSIM, above $0.99994$. Errors are essentially flat across window lengths, and neither single-branch variant improves monotonically with sequence length: S-LSTM ranges from $0.112\%$ at 7 days to $0.092\%$ at 60 days and $0.097\%$ at 90 days, while S-GRU attains the lowest value of the sweep, $0.085\%$, at 90 days. The two encoders are therefore comparable, and their ordering is not stable across window lengths. These are single training runs without seed replication, and we do not read a trend into differences of this size.

The multi-branch configurations (M-GRU, M-LSTM) displayed noticeably degraded performance at every sequence length, although their errors do not follow a consistent trend with window length. The multi-branch variants differ from the single-branch ones in encoding each station separately, in their twelve-way multiplicative fusion and in their parameter count, and these runs do not isolate which of these causes the reduction. The static-branch DeepONet, whose branch encodes only the final time step of the input window, yielded higher relative $L_2$ errors ($0.13$--$0.19\%$) and lower SSIM values than the single-branch variants under identical data and splits.

Within this comparison the single-branch configurations are the most accurate. They differ from the multi-branch variants in encoding all stations jointly; the length of the temporal context does not separate them. 

\begin{table}[!htbp]
\centering
\caption{Performance comparison across sequence lengths. All TRON variants—S-GRU, S-LSTM, M-GRU, and M-LSTM—share the same TRON backbone and differ only in temporal encoding (single vs.\ multi-branch, GRU vs.\ LSTM). Linear regression baselines are included for comparison. All values are from a single training run per configuration.}
\label{tab:performance}
\begin{tabular*}{\textwidth}{@{\extracolsep\fill}cccccc@{}}
\toprule
\multicolumn{1}{l}{Seq} & Model & Relative $L_2$ Error (\%) & RMSE & MAE & SSIM \\ 
\midrule
\multirow{9}{*}{7}  
  & Static-branch DeepONet & $1.87\times10^{-1}$ & $6.26\times10^{-5}$ & $5.05\times10^{-5}$ & $0.998934$ \\
  & S-GRU        & $1.00\times10^{-1}$ & $3.37\times10^{-5}$ & $2.84\times10^{-5}$ & $0.999546$ \\
  & S-LSTM       & $1.12\times10^{-1}$ & $3.78\times10^{-5}$ & $3.11\times10^{-5}$ & $0.999370$ \\
  & M-GRU        & $8.98\times10^{-1}$ & $3.01\times10^{-4}$ & $2.69\times10^{-4}$ & $0.998256$ \\
  & M-LSTM       & $6.24\times10^{-1}$ & $2.09\times10^{-4}$ & $1.88\times10^{-4}$ & $0.998120$ \\
  & GNO          & $1.86$ & $6.24\times10^{-4}$ & $4.86\times10^{-4}$ & $0.738379$ \\
  & GINO         & $2.61$ & $8.74\times10^{-4}$ & $8.22\times10^{-4}$ & $0.996221$ \\
  \cmidrule(lr){2-6}
  & Linear Reg (Static)   & $3.71\times10^{-1}$ & $1.25\times10^{-4}$ & $1.21\times10^{-4}$ & $0.999958$ \\
  & Linear Reg (Temporal) & $3.43\times10^{-1}$ & $1.16\times10^{-4}$ & $1.12\times10^{-4}$ & $0.999963$ \\
\midrule
\multirow{9}{*}{30} 
  & Static-branch DeepONet & $1.28\times10^{-1}$ & $4.32\times10^{-5}$ & $3.56\times10^{-5}$ & $0.999145$ \\
  & S-GRU        & $1.09\times10^{-1}$ & $3.66\times10^{-5}$ & $2.97\times10^{-5}$ & $0.999360$ \\
  & S-LSTM       & $9.59\times10^{-2}$ & $3.23\times10^{-5}$ & $2.59\times10^{-5}$ & $0.999424$ \\
  & M-GRU        & $7.55\times10^{-1}$ & $2.53\times10^{-4}$ & $2.26\times10^{-4}$ & $0.998406$ \\
  & M-LSTM       & $7.93\times10^{-1}$ & $2.66\times10^{-4}$ & $2.36\times10^{-4}$ & $0.996193$ \\
  & GNO          & $1.80$ & $6.04\times10^{-4}$ & $4.70\times10^{-4}$ & $0.502741$ \\
  & GINO         & $2.60$ & $8.72\times10^{-4}$ & $8.25\times10^{-4}$ & $0.996207$ \\
  \cmidrule(lr){2-6}
  & Linear Reg (Static)   & $3.74\times10^{-1}$ & $1.26\times10^{-4}$ & $1.22\times10^{-4}$ & $0.999957$ \\
  & Linear Reg (Temporal) & $3.53\times10^{-1}$ & $1.19\times10^{-4}$ & $1.15\times10^{-4}$ & $0.999960$ \\
\midrule
\multirow{9}{*}{60} 
  & Static-branch DeepONet & $1.88\times10^{-1}$ & $6.32\times10^{-5}$ & $5.18\times10^{-5}$ & $0.999180$ \\
  & S-GRU        & $9.95\times10^{-2}$ & $3.35\times10^{-5}$ & $2.73\times10^{-5}$ & $0.999429$ \\
  & S-LSTM       & $9.22\times10^{-2}$ & $3.10\times10^{-5}$ & $2.49\times10^{-5}$ & $0.999486$ \\
  & M-GRU        & $7.47\times10^{-1}$ & $2.51\times10^{-4}$ & $2.21\times10^{-4}$ & $0.998100$ \\
  & M-LSTM       & $6.34\times10^{-1}$ & $2.13\times10^{-4}$ & $1.96\times10^{-4}$ & $0.998357$ \\
  & GNO          & $1.96$ & $6.57\times10^{-4}$ & $5.20\times10^{-4}$ & $0.493162$ \\
  & GINO         & $2.62$ & $8.77\times10^{-4}$ & $8.30\times10^{-4}$ & $0.996190$ \\
  \cmidrule(lr){2-6}
  & Linear Reg (Static)   & $3.76\times10^{-1}$ & $1.27\times10^{-4}$ & $1.23\times10^{-4}$ & $0.999957$ \\
  & Linear Reg (Temporal) & $3.76\times10^{-1}$ & $1.26\times10^{-4}$ & $1.22\times10^{-4}$ & $0.999951$ \\
\midrule
\multirow{9}{*}{90} 
  & Static-branch DeepONet & $1.93\times10^{-1}$ & $6.49\times10^{-5}$ & $5.40\times10^{-5}$ & $0.999215$ \\
  & S-GRU        & $8.52\times10^{-2}$ & $2.87\times10^{-5}$ & $2.32\times10^{-5}$ & $0.999555$ \\
  & S-LSTM       & $9.70\times10^{-2}$ & $3.27\times10^{-5}$ & $2.61\times10^{-5}$ & $0.999445$ \\
  & M-GRU        & $7.77\times10^{-1}$ & $2.61\times10^{-4}$ & $2.28\times10^{-4}$ & $0.997830$ \\
  & M-LSTM       & $1.20\times10^{0}$  & $4.01\times10^{-4}$ & $3.58\times10^{-4}$ & $0.992508$ \\
  & GNO          & $1.87$ & $6.29\times10^{-4}$ & $4.93\times10^{-4}$ & $0.608173$ \\
  & GINO         & $2.63$ & $8.80\times10^{-4}$ & $8.33\times10^{-4}$ & $0.996180$ \\
  \cmidrule(lr){2-6}
  & Linear Reg (Static)   & $3.79\times10^{-1}$ & $1.28\times10^{-4}$ & $1.24\times10^{-4}$ & $0.999956$ \\
  & Linear Reg (Temporal) & $3.89\times10^{-1}$ & $1.31\times10^{-4}$ & $1.26\times10^{-4}$ & $0.999949$ \\
\bottomrule
\end{tabular*}
\end{table}

\begin{figure}[!htbp]
    \centering
    \makebox[\textwidth][c]{\includegraphics[width=180mm]{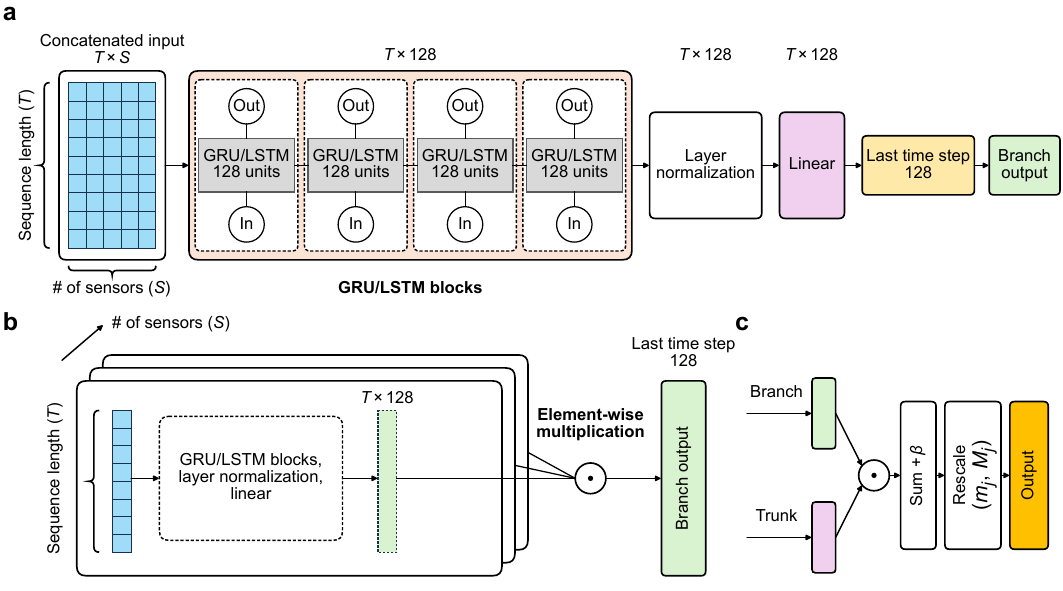}}%
\caption{\textbf{Branch network architectures in TRON and the fusion with the trunk output.}
\textbf{a} Single-branch TRON: neutron count sequences from all monitoring stations are concatenated into a $T \times S$ tensor ($T$, input window length in days; $S = 12$ stations) and processed through four stacked GRU (gated recurrent unit) or LSTM (long short-term memory) layers of 128 units, followed by layer normalization and a linear layer applied at every time step ($T \times 128$); the output at the last time step (128 values) is the branch output.
\textbf{b} Multi-branch TRON: each neutron monitor's sequence ($T \times 1$) is processed independently by its own GRU/LSTM encoder of the same form. The resulting sensor-specific latent representations are fused via element-wise multiplication across the stations to give the branch output (last time step, 128 values).
\textbf{c} Fusion with the trunk network: the branch output is multiplied element-wise ($\odot$) with the trunk network's spatial embedding and summed, and a learnable scalar bias $\beta$ is added, giving a normalized dose; this is rescaled with the per-point output scaling $(m_j, M_j)$ to the effective dose rate at query point $j$.}
    \label{fig:blocks}
\end{figure}

\begin{figure}[htbp]
\centering
\makebox[\textwidth][c]{\includegraphics[width=126mm]{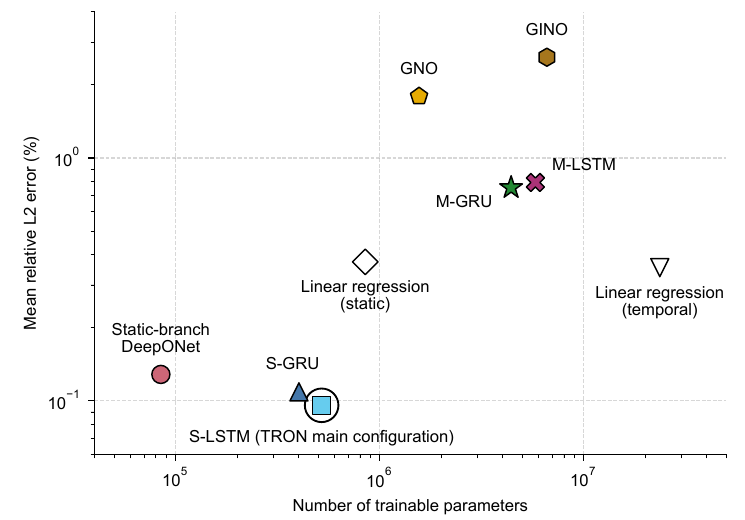}}%
\caption{Comparative performance of neural operator architectures. Mean relative $L_2$ error versus number of trainable parameters at $T = 30$ days; each point is a single training run. Ring, S-LSTM, the main TRON configuration. TRON variants (S-, single-branch; M-, multi-branch; LSTM, long short-term memory; GRU, gated recurrent unit) and the static-branch DeepONet (deep operator network) colored as in main-text Fig.~2; yellow, GNO (graph neural operator); brown, GINO (geometry-informed neural operator); open symbols, linear regressions (static, current day only; temporal, whole window; Section~\ref{sec:baselines}). Values in Tables~\ref{tab:performance} and~\ref{tab:parameters}.}
\label{fig:params_vs_l2error}
\end{figure}

Fig.~\ref{fig:blocks} shows the branch architectures of the TRON variants and their fusion with the trunk output; Table~\ref{tab:parameters} lists the number of trainable parameters of every model, and Fig.~\ref{fig:params_vs_l2error} plots the error against the number of parameters.

\begin{table}[htbp]
\centering
\caption{Summary of number of model parameters. The temporal linear regression has $(12T+1)\times 65{,}341$ parameters; the value shown is for $T = 30$ days.}
\label{tab:parameters}
\begin{tabular}{@{}ccc@{}}
\toprule
Type                                & Variation    & \# of trainable parameters \\ \midrule \midrule
\multirow{2}{*}{Linear Regressions} & Static       & 849,433                    \\
                                    & Temporal     & 23,588,101                 \\ \midrule
\multirow{3}{*}{Neural Operators}   & Static-branch DeepONet & 84,609                     \\
                                    & GNO          & 1,558,765                  \\
                                    & GINO         & 6,596,995                  \\ \midrule
\multirow{4}{*}{TRON}               & S-GRU        & 401,921                    \\
                                    & S-LSTM       & 519,169                    \\
                                    & M-GRU        & 4,404,865                  \\
                                    & M-LSTM       & 5,794,945                  \\ \bottomrule
\end{tabular}
\end{table}

Supplementary Table~\ref{tab:speed}, plotted in main-text Fig.~5, gives the inference time of each trained model for one daily global field, with the timing protocol in its caption, and the main-text Results give the resulting speed-ups relative to the EXPACS simulation. The inference times are averaged over the windows lying entirely within 2023 (359, 336, 306 and 276 for $T = 7$, 30, 60 and 90 days; accuracy is evaluated on all 365 windows ending in 2023), after five warm-up passes, and include the transfer of each batch to the GPU; each per-field time is the batch time divided by the mean batch size, from one timed pass. The timing notebook and its log are \texttt{code/analysis/resource/inference\_time.ipynb} and \texttt{inference\_time\_log.txt}. The timed multi-branch models were loaded non-strictly and carried an untrained final $128 \to 1$ output layer, whose cost is negligible next to the twelve recurrent encoders. The timing notebook divided their batch times by the number of stations instead of the batch size; the values in Supplementary Table~\ref{tab:speed} and main-text Fig.~5 are corrected for this normalization error by \texttt{code/analysis/resource/correct\_multibranch\_timing.py} (\texttt{inference\_time\_corrected.log}).

\begin{table}[htbp]
\caption{Simulation and inference time per daily global field ($65{,}341$ points), plotted in main-text Fig.~5. Inference on an NVIDIA A100 GPU in batches of 16: total GPU time divided by the number of fields, over the input windows lying entirely within 2023, from one timed pass. The EXPACS simulation of the same field ran on one thread of an AMD Ryzen 9 7950X3D CPU (mean of three runs). No spread is given: the standard deviation of the batch times, divided by the mean batch size, is dominated by the last, smaller batch (7, 2 and 4 fields at $T = 7$, 60 and 90 days; none at 30 days) rather than by timing noise. Values from \texttt{inference\_time\_corrected.log} and \texttt{expacs\_timing/speedup.log} (\texttt{code/analysis/resource/}).}
\label{tab:speed}
\centering
\begin{tabular}{lccccc}
\toprule
\multirow{2}{*}{Method} & \multirow{2}{*}{Simulation (s)} & \multicolumn{4}{c}{Inference time (ms), input window $T$ (days)} \\
\cmidrule(l){3-6}
 & & 7 & 30 & 60 & 90 \\
\midrule
EXPACS (one CPU thread) & 57.6 & - & - & - & - \\
DeepONet & - & 0.981 & 1.256 & 0.987 & 1.007 \\
Single-branch GRU & - & 0.993 & 1.006 & 1.016 & 1.025 \\
Single-branch LSTM & - & 1.005 & 1.021 & 1.032 & 1.054 \\
Multi-branch GRU & - & 1.262 & 1.309 & 1.444 & 1.532 \\
Multi-branch LSTM & - & 1.342 & 1.509 & 1.700 & 1.901 \\
\bottomrule
\end{tabular}
\end{table}

\section{Short-Horizon Forecasting (One-Day Ahead)}
\label{sec:forecasting}

TRON is trained and evaluated as a nowcasting operator: the target is the field on the last day of the input window. This section tests whether the same architecture, trained for the purpose, predicts the field one day beyond its most recent observation. With the daily station readings $\mathbf{y}(1), \dots, \mathbf{y}(N)$ and reference fields $\mathbf{X}(1), \dots, \mathbf{X}(N)$ of Section~\ref{sec:dataset}, the input at step $i$ is the window
\begin{equation}
    \mathbf{Y}^{(i)} = [\mathbf{y}(i), \dots, \mathbf{y}(i+T-1)]^{\top} \in \mathbb{R}^{T \times S},
\end{equation}
and the target is the field one day after its last element, $\mathbf{X}(i+T)$, instead of $\mathbf{X}(i+T-1)$ as in the nowcasting setup.

The window length was $T = 30$~days, and samples were assigned to training (2001--2021), validation (2022) and test (2023) by their target day. Two architectures were trained for this task: the single-branch LSTM configuration of TRON (S-LSTM), with and without dropout of probability 0.2 in the branch and trunk sub-networks, and the static-branch DeepONet baseline of Section~\ref{sec:baselines}, whose branch uses the station readings on the last day of the window only. For the dropout model the prediction is the mean of 50 Monte Carlo dropout samples. The inputs are min--max scaled per station and the targets per grid point on the training data, and the trunk takes latitude and longitude in degrees, unlike the main-text model, which normalizes them. The five further runs of the dropout configuration (seeds 12345--12349) were trained with Adam (learning rate $10^{-3}$, weight decay $10^{-3}$), a step learning-rate schedule (factor 0.7 every 20 epochs), a mean-squared-error loss, at most 300 epochs and early stopping after 20 epochs without improvement of the validation loss; the optimizer settings of the three trained models in Table~\ref{tab:forecast_comparison} are not recorded. The trained models are compared with two predictors that need no training for the task. The first is the reference field of the previous day (persistence); it is not available at inference, because the reference field is not observed, and is included only to show how much the field changes from one day to the next. The second is the nowcast of the previous day by the S-LSTM ($T = 30$) of the main text, carried forward unchanged; it uses exactly the same inputs as the forecasting models.

\begin{table}[htbp]
\caption{One-day-ahead prediction of the 2023 reference fields (365 target days). Each trained model is a single training run; five further runs of the dropout configuration with different seeds gave relative $L_2$ errors of 0.90--2.04\% with dropout disabled at inference. SSIM as defined in the main-text Methods.}
\centering
\begin{tabular}{@{}>{\raggedright\arraybackslash}p{0.53\linewidth}ccc@{}}
\toprule
Predictor & \begin{tabular}[c]{@{}c@{}}Trained for\\ the task\end{tabular} & \begin{tabular}[c]{@{}c@{}}Relative $L_2$\\ error (\%)\end{tabular} & SSIM \\ \midrule
Reference field of the previous day (persistence)$^{\mathrm{a}}$ & no  & 0.291 & 0.9999 \\
TRON nowcast of the previous day, carried forward                & no  & 0.314 & 0.9994 \\
Static-branch DeepONet                                                     & yes & 0.546 & 0.9919 \\
TRON (S-LSTM)                                                    & yes & 0.805 & 0.9978 \\
TRON (S-LSTM), dropout 0.2, mean of 50 samples                   & yes & 0.864 & 0.9976 \\ \bottomrule
\end{tabular}
\par\vspace{2pt}
\raggedright\footnotesize $^{\mathrm{a}}$Uses the reference field, which is not available at inference.
\label{tab:forecast_comparison}
\end{table}

\begin{figure}[!tbp]
\centering
\makebox[\textwidth][c]{\includegraphics[width=180mm]{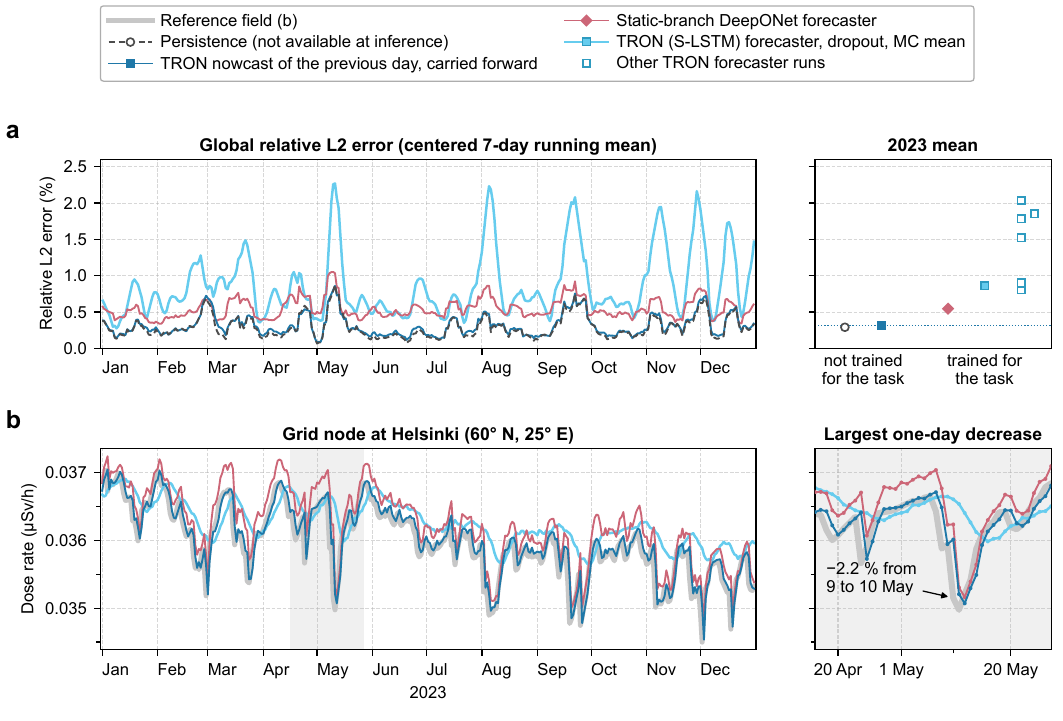}}
\caption{One-day-ahead prediction of the 2023 reference fields (365 target days); each trained model is a single training run. \textbf{a}, Global relative $L_2$ error (centered 7-day running mean). Right: 2023 means, including the TRON forecaster without dropout and five further seeds of the dropout configuration with dropout disabled (open squares); dotted line, carried-forward nowcast. Persistence, the previous day's reference field, is not available at inference; MC mean, mean of 50 Monte Carlo dropout samples; S-LSTM, single-branch TRON with an LSTM (long short-term memory) encoder; DeepONet, deep operator network. \textbf{b}, Reference dose rate and predictions at the Helsinki grid node (60$^{\circ}$\,N, 25$^{\circ}$\,E). Right: the weeks around the largest one-day decrease of 2023 (9 to 10 May), daily values marked; shaded gray at left.}
\label{fig:forecast}
\end{figure}

None of the trained models improved on carrying the nowcast forward in the 2023 mean (Table~\ref{tab:forecast_comparison}, Fig.~\ref{fig:forecast}a). Their mean relative $L_2$ errors were 1.7 (static-branch DeepONet) and 2.6--2.8 (TRON) times that of the carried-forward nowcast, and their SSIM was also lower; on individual days, the static-branch DeepONet and the TRON forecaster (dropout, Monte Carlo mean) had the lower error on 41 and 28 of the 365 days. The model with no access to the history within the window (static-branch DeepONet) had the lowest error of the trained models, but the spread across training seeds of the dropout configuration (0.90--2.04\%) is larger than the differences between the trained models, so single runs do not establish their ranking. All eight trained models, however, had 2023 mean errors above the 0.314\% of the carried-forward nowcast. Because the field is a one-parameter family in time (Section~\ref{sec:intrinsic}), one node is representative: the 2023 series at the twelve city nodes of main-text Fig.~3 are correlated with $r > 0.9998$. At the Helsinki grid node (Fig.~\ref{fig:forecast}b), none of the predictors anticipated the largest one-day decrease of 2023 (9 to 10 May); the carried-forward nowcast and the static-branch DeepONet followed it one day later, the TRON forecaster more slowly.

The day-to-day change of the reference field averaged 0.291\% over 2023, three times the nowcast error of the same S-LSTM (0.096\%, Table~\ref{tab:performance}). Because the reference field is a one-parameter family in time (Section~\ref{sec:intrinsic}), forecasting the field one day ahead amounts to forecasting that single scalar one day ahead. These experiments give no evidence that a 30-day window of counts predicts its next-day change better than assuming no change, and TRON is accordingly presented in the main text as a nowcasting operator.

\section{Evaluation Under New-Station Integration Scenarios}
\label{sec:newstation}
We tested whether a newly installed neutron monitor can be added to the trained model without retraining. The MXCO monitor (Mexico City, $19.33^{\circ}$\,N, $99.18^{\circ}$\,W) is treated as becoming available in 2023 and is appended as a thirteenth input channel of the S-LSTM ($T = 30$ days). The first-layer input weights of the LSTM branch are expanded from 12 to 13 columns: the existing columns, all recurrent weights and biases, the branch projection layer, the output bias and the trunk are copied from the published model, and the new column is initialized at zero. The expanded model therefore reproduces the published model until it is adapted (maximum difference $6\times10^{-7}$\,\textmu Sv\,h$^{-1}$ over the test year). It is then adapted on the first 7, 14 or 30 days of 2023, updating only the first-layer input weights and input biases of the LSTM (the recurrent biases stay fixed), the branch projection layer and the output bias (AdamW, learning rate $10^{-4}$, weight decay $10^{-4}$, 12 full-batch epochs, gradient-norm clipping at 1.0, mean squared error on the per-point normalized targets), and evaluated on the remaining days of 2023. Each window starts again from the published model. Because the new column starts at zero and each window is a single full batch, the procedure is deterministic, and repeating it with three random seeds gave identical results.

The MXCO daily series contains isolated values far from its running level (for example 10.7 on 2009-05-07, against a median of about 225). Days deviating by more than $20\%$ from a centered 31-day running median (39 of 8,400 days) were treated as missing and linearly interpolated before min--max scaling; on the remaining days the deviations from the running median are much smaller (99th percentile $4.3\%$). The min--max scaling of the MXCO channel and the training-period mean used in (B) below are computed from its 2001--2021 record, which a newly installed monitor would not have.

We compare (A) the adapted 13-channel model, (B) the same adapted model with MXCO replaced by its training-period mean, and (O) the published 12-station model evaluated on the same days. Results in the North American sector ($140^{\circ}$--$40^{\circ}$\,W, $0^{\circ}$--$60^{\circ}$\,N) are given in Table~\ref{tab:adapt_local} and global results in Table~\ref{tab:adapt_world}.

\begin{table}[h!]
\centering
\caption{Relative $L_2$ error in the North American sector after adding MXCO, mean over the evaluation days (one adapted model per window).}
\label{tab:adapt_local}
\begin{tabular}{@{}ccccc@{}}
\toprule
\begin{tabular}[c]{@{}c@{}}Adaptation\\window (days)\end{tabular} & \begin{tabular}[c]{@{}c@{}}Evaluation\\days\end{tabular} & O (12 stations) & A (13-ch adapted) & B (MXCO masked) \\
\midrule
7  & 358 & 0.126\% & 0.134\% & 0.134\% \\
14 & 351 & 0.127\% & 0.136\% & 0.136\% \\
30 & 335 & 0.128\% & 0.141\% & 0.141\% \\
\bottomrule
\end{tabular}
\end{table}

\begin{table}[h!]
\centering
\caption{Global relative $L_2$ error after adding MXCO, mean over the evaluation days (one adapted model per window).}
\label{tab:adapt_world}
\begin{tabular}{@{}ccccc@{}}
\toprule
\begin{tabular}[c]{@{}c@{}}Adaptation\\window (days)\end{tabular} & \begin{tabular}[c]{@{}c@{}}Evaluation\\days\end{tabular} & O (12 stations) & A (13-ch adapted) & B (MXCO masked) \\
\midrule
7  & 358 & 0.097\% & 0.104\% & 0.104\% \\
14 & 351 & 0.097\% & 0.107\% & 0.107\% \\
30 & 335 & 0.098\% & 0.110\% & 0.110\% \\
\bottomrule
\end{tabular}
\end{table}

Adapting to MXCO does not improve the reconstruction. In every window the adapted model is slightly less accurate than the published model, by $8$--$12\%$ in relative terms globally, and the difference grows with the length of the window; it is better than the published model on 97 of 358, 96 of 351 and 74 of 335 evaluation days. SSIM is $0.9994$ for all three models. Masking MXCO leaves the error unchanged to the stated precision: after adaptation MXCO has no measurable effect on the prediction, so the small loss of accuracy arises from the adaptation of the other parameters on a few days of data. Within this protocol, appending the channel reproduces the published model, but adaptation on 7--30 days raises the error by 8--12\%, and a short commissioning period does not extract information from the new station; whether retraining with its history would make use of it, we did not test.
\section{Empirical Probing of Latent Representations}
\label{sec:latent}
To empirically assess how TRON encodes coupled sensor dynamics, the latent representations learned by both the single- and multi-branch architectures were analyzed. 
In this analysis, single-branch and multi-branch TRON models trained with a sequential input length of 30~days were utilized. 
Latent activations were computed over the 365 days of the 2023 test year, with 12 sensors and a latent dimensionality of 128 for each LSTM encoder, giving arrays of shape $(365,128)$ for the single-branch encoder, $(12,365,128)$ for the multi-branch pre-fusion outputs, and $(365,128)$ for the post-fusion combined latent representation.

From each trained model, latent activations were extracted from the final hidden state of the LSTM-based temporal encoder ($h_n[-1]$) following layer normalization and linear projection. 
In the single-branch configuration, the temporal sequences from all sensors were concatenated and processed jointly by a unified encoder $\mathcal{B}$ to produce a latent feature vector at the final time step,
\begin{equation}
    \mathbf{b} = \mathcal{B}(\mathbf{Y}) \in \mathbb{R}^{d},
\end{equation}
where $\mathbf{Y} \in \mathbb{R}^{T \times S}$ denotes the concatenated temporal input, $T$ is the sequence length, and $d=128$ represents the latent dimensionality. 

In the multi-branch configuration, each sensor sequence $\mathbf{Y}_s$ was processed independently by a dedicated encoder $\mathcal{B}_s$, producing sensor-specific latent outputs
\begin{equation}
    \mathbf{b}_s = \mathcal{B}_s(\mathbf{Y}_s) \in \mathbb{R}^{d}, \quad s = 1, \dots, S.
\end{equation}
The fused latent representation analyzed in this study corresponds to the multiplicative fusion defined in the main text, expressed as
\begin{equation}
    \mathbf{b} = \prod_{s=1}^{S} \mathbf{b}_s \in \mathbb{R}^{d},
\end{equation}
where the element-wise product is taken across the sensor dimension ($S=12$). 
This fusion combines the stations only after each has been encoded separately.

Feature--feature correlation matrices were computed using the Pearson correlation coefficient to quantify the degree of coupling among latent dimensions within each configuration. 
For the multi-branch configuration, pairwise inter-sensor correlations were also evaluated by flattening each sensor-specific latent and computing the mean absolute correlation between all sensor pairs, i.e.,
\begin{equation}
    \rho_{ss'} = \operatorname{corr}\!\big(\mathrm{vec}(\mathbf{b}_s), \mathrm{vec}(\mathbf{b}_{s'})\big),
    \quad 1 \le s < s' \le S.
\end{equation}

The single-branch architecture exhibited strong internal feature coupling (mean $|\rho| = 0.81$), consistent with the stations being encoded jointly: the twelve stations carry essentially one common modulation signal (Section~\ref{sec:intrinsic}), and a jointly trained encoder represents it in many correlated features. 
In contrast, the multi-branch architecture exhibited weak feature correlations prior to fusion ($|\rho| = 0.25$) and low inter-sensor coupling ($|\rho| = 0.10$), so the separately encoded branches produced largely uncorrelated, sensor-specific features. After fusion the mean $|\rho|$ rises to 0.52; 16 of the 128 fused features are constant over the test year and are counted with $\rho = 0$, and over the other 112 the mean is 0.68. The three correlation matrices are shown in main-text Fig.~4b--d.

The observed differences in latent coupling are consistent with distinct optimization behaviors between the single- and multi-branch TRON architectures. 
In the single-branch configuration, gradients from all sensors propagate through a shared encoder. In the multi-branch configuration the branches are trained jointly, but each receives gradients scaled by the product of the other branch outputs, and global coupling emerges only after the multiplicative fusion rather than during temporal sequence encoding. We did not analyze the loss landscape, and these observations do not isolate a mechanism. 

The difference in latent coupling is accompanied by the larger mean errors and wider day-to-day spread of the multi-branch variants in main-text Fig.~2a and Table~\ref{tab:performance}.

\newpage
\section{Demonstration of TRON on Additional Physics Problems}
\label{sec:physics}
The cosmic radiation dose field reconstruction presented in the main text involves relatively smooth spatial variations at global scales, which raises questions about TRON's performance on problems with sharper gradients and stronger local features. To address this concern and evaluate the architecture's broader applicability, we apply TRON to two additional physics problems that exhibit fundamentally different spatial structures: internal state estimation in heat transfer systems and fluid flow field reconstruction from sparse vorticity sensors. The heat exchanger problem introduces tightly confined geometry with interior velocity fields that contain recirculation patterns and steep gradients near the walls. The Navier-Stokes flow problem further increases complexity by adding temporal dynamics, featuring sharp velocity gradients, boundary layers, and evolving vortical structures that require the model to resolve fine-scale spatiotemporal features from sparse measurements. These applications demonstrate that TRON successfully handles field reconstruction beyond the smooth global-scale variations of atmospheric radiation fields, maintaining accurate inference even in the presence of sharp spatial gradients, localized flow structures, and transient dynamics. Neither benchmark includes a comparison model, so they show that the formulation applies to these problems, not that it outperforms alternatives.

\subsection{Internal State Estimation in Heat Transfer Systems (Spatial Problem)}

Heat exchangers in nuclear microreactor applications present a challenging virtual sensing problem: internal flow structures evolve within tightly confined channels where direct instrumentation is physically impossible \cite{ahmed2024enhancing}. The confined geometry prevents sensor placement, forcing diagnostic systems to infer interior velocity fields entirely from accessible boundary measurements. This creates a sparse-to-dense reconstruction challenge where limited boundary information must be transformed into detailed internal flow field estimates.

\begin{figure}[!tbp] 
\centering 
\makebox[\textwidth][c]{\includegraphics[width=180mm]{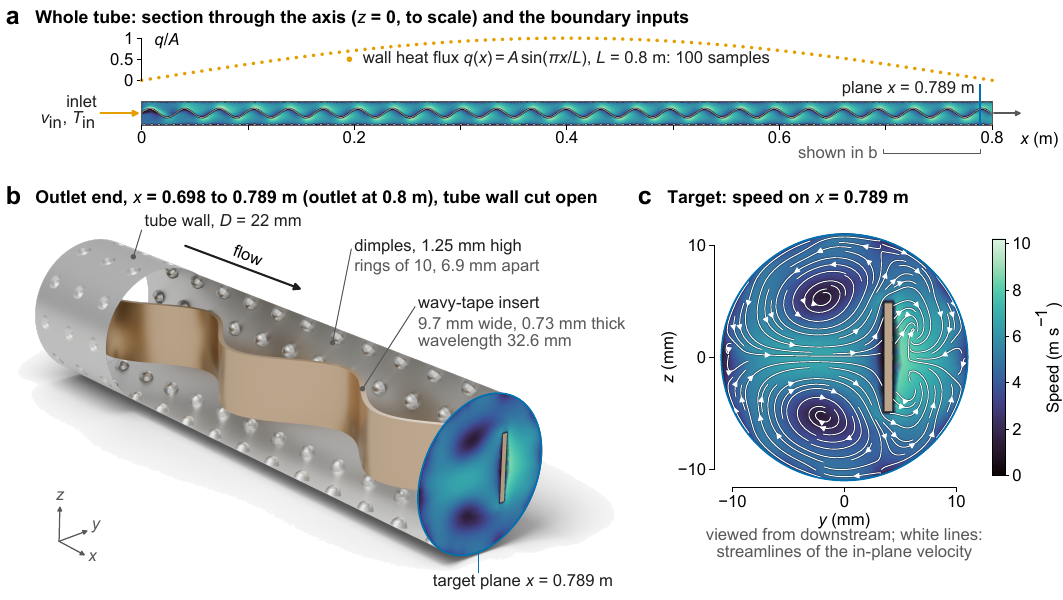}}
\caption{Heat-exchanger configuration: computational fluid dynamics mesh and solution of simulation 0 of the 1{,}546-simulation dataset of ref.~\cite{kobayashi2026virtual}. \textbf{a} Speed on the axial section $z = 0$ of the whole tube (to scale) with the boundary inputs of TRON: wall heat flux $q(x) = A \sin(\pi x/L)$ ($L$, tube length; 100 samples, normalized by $A$) and uniform inlet velocity $v_{\text{in}}$ and temperature $T_{\text{in}}$. Orange, boundary inputs; blue lines, target plane. \textbf{b} The last 91~mm upstream of the target plane, tube wall cut open: dimpled tube ($D$, diameter) with wavy-tape insert, dimensions as labeled. \textbf{c} Reconstruction target: speed on the plane $x = 0.789$~m (3{,}977 nodes), viewed from downstream; white lines, streamlines of the in-plane velocity. One color bar for all panels.}
\label{fig:hx}
\end{figure}

We consider a compact heat exchanger: a dimpled tube of length $L = 0.8$~m and diameter 22~mm with a wavy-tape insert, a configuration of the type studied in ref.~\cite{ahmed2024enhancing} (Fig.~\ref{fig:hx}). The wall heat flux follows a prescribed distribution along the tube axis,
\begin{equation}
q(x) = A \sin\left( \frac{\pi x}{L} \right), \quad x \in [0, L].
\end{equation}
The system is further driven by scalar inlet velocity and temperature boundary conditions. As shown in Fig.~\ref{fig:hx}, the reconstruction target is the velocity magnitude of the converged flow on the cross-sectional plane $x = 0.789$~m, 11~mm upstream of the outlet.

The dataset is the heat-exchanger dataset of ref.~\cite{kobayashi2026virtual}, computed with ANSYS Fluent. It comprises 1{,}546 simulations in which the heat-flux amplitude $A$, the inlet temperature $T_{\text{in}}$ and the inlet velocity $v_{\text{in}}$ were sampled independently from uniform distributions spanning $\pm 10\%$ of their nominal values ($v_{\text{in}}$ from 4.05 to 4.95~m\,s$^{-1}$ and $T_{\text{in}}$ from 263.7 to 322.3~K). For each simulation the converged solution is sampled on the plane $x = 0.789$~m. The profile varies along the tube axis, as the wall-normal temperature gradient of the three-dimensional solution confirms (it follows $\sin(\pi x/L)$ with $R^2 = 0.9994$).

The wall heat flux was sampled at 100 equally spaced locations along $x$ and concatenated with the two scalar inputs into one vector $\mathbf{u} \in \mathbb{R}^{102}$, which the recurrent branch receives as a sequence of length one; because every profile has the shape $\sin(\pi x/L)$, the 100 samples differ between simulations only through the amplitude $A$. The dataset was divided at random, with a fixed seed and the same permutation for inputs and targets, into 1{,}236 training, 155 validation and 155 test samples, with each input mapped to a velocity magnitude field on the cross-sectional plane with 3{,}977 spatial nodes.

TRON uses a 4-layer GRU branch network (256 hidden units, output dimension 256) on the 102-dimensional input and a 4-layer fully connected trunk network (256 units per layer, ReLU activation) on the 2D spatial coordinates $(y, z)$, scaled to $[-1, 1]$; the output at each point is the inner product of the branch and trunk outputs plus a learnable scalar bias. The two scalar inputs are standardized per input and the heat-flux samples with one common mean and standard deviation, all on the training set. Training used the Adam optimizer (initial learning rate $10^{-3}$, weight decay $10^{-4}$) on the relative $L_2$ loss of mini-batches of 16 samples, with ReduceLROnPlateau scheduling (factor 0.5, patience 10 epochs), early stopping on the validation loss (patience 20 epochs) and at most 500 epochs.

Over three training runs with different random initializations on the same split, TRON achieved relative $L_2$ errors of 0.63\%, 0.71\% and 2.79\% on the test set (averaged over test samples), corresponding to RMSEs of 0.028, 0.032 and 0.127~m\,s$^{-1}$ and $R^2$ of 0.9997, 0.9997 and 0.9946 ($R^2$ over all test values). The run with the largest error stopped early after 70 epochs; the other two reached the 500-epoch limit. These results show that TRON reconstructs the interior velocity-magnitude field from the boundary conditions of this configuration.

\subsection{Fluid Flow Field Reconstruction from Sparse Sensors (Spatiotemporal Problem)} 

State estimation to recover the full flow field from sparse flow information is a challenging topic in fluid mechanics \cite{milano2002neural,kumar2022state}. Whether for fluid control applications or aerodynamics design, full-field data is often essential for resolving non-linear structures like vortices and pressure gradients. In this demonstration, we undertake the canonical example of a flow past a cylinder and aim to reconstruct the vorticity field from sparse sensor measures of vorticity. The flow is transient in nature, which makes the problem quite challenging compared to the reconstruction of steady-state flow fields.

A circular cylinder with a diameter of $D$ is considered. The center of the cylinder is located at a distance of $8D$ from the inlet, and the outlet is located at a distance of $25D$ from the center of the cylinder. The sidewalls are located 4 units away from the center of the cylinder. A uniform velocity of 1 unit along the $x$-direction is applied in the inlet, whereas a zero pressure boundary condition is considered at the outlet. The walls and the cylinder surface assume a no-slip boundary condition.
The training and testing datasets are generated using the Unsteady Reynolds-averaged Navier-Stokes (URANS) simulation in OpenFOAM. The problem domain is discretized into 63420 elements; however, only the rear half-cylinder with a downstream wake extension is cut out during data generation. The cutout snapshot has a resolution of $252 \times 502$. The time step is taken as $ \Delta t = 0.02$ units. The training and testing datasets are generated by saving snapshots at the time step of $10 \Delta t$. In total, the training dataset comprises 1{,}197 snapshots, 399 sequential snapshots at each of the Reynolds numbers 180, 190 and 200, and the test set comprises 798 snapshots, 399 sequential snapshots at each of Re = 185 and 195, which are not seen in training.

\begin{figure}[!ht]
    \centering
    \makebox[\textwidth][c]{\includegraphics[width=180mm]{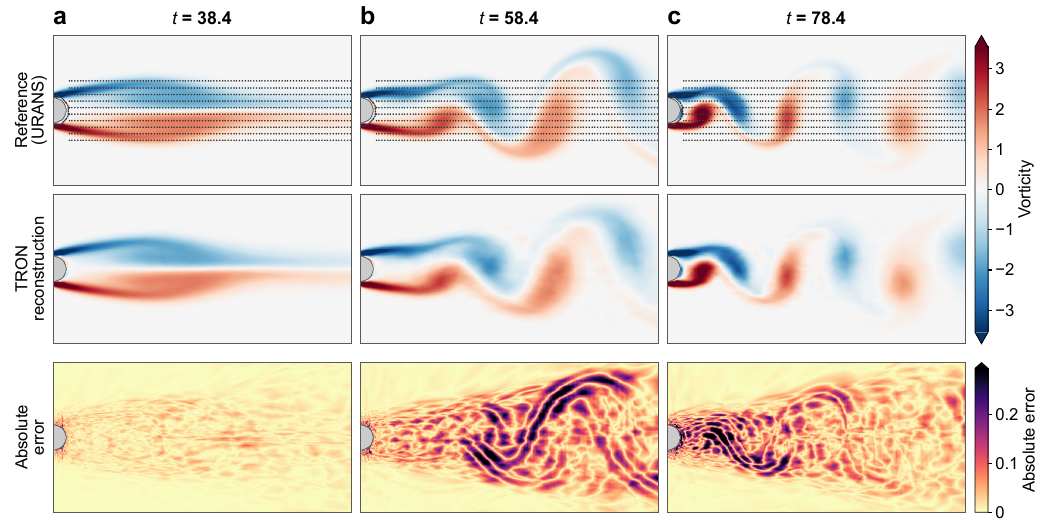}}
    \caption{Vorticity ($\omega$) reconstruction for the Reynolds number Re = 195 test sequence (training used Re = 180, 190 and 200). \textbf{a}--\textbf{c} Three instants; $t$, dimensionless time since the first stored snapshot (snapshots every 0.2). Rows: reference from unsteady Reynolds-averaged Navier--Stokes (URANS) simulation; TRON reconstruction from the current and two preceding readings of the 1{,}030 vorticity sensors (dark dots, top row; cylinder in gray); absolute error. Vorticity scale symmetric, clipped at $\pm 3.55$ (99.5th percentile of $|\omega|$ over the three reference fields; maximum 11.3); error scale clipped at 0.295 (99.5th percentile of the three error fields; maximum 1.26); both maxima at the cylinder surface. Single training run; the test sequences were also used for early stopping.}
    \label{fig:ns_reconstruct}
\end{figure}

The spatiotemporal reconstruction is performed using the TRON architecture featuring an 8-layer GRU branch network (512 hidden units, output dimension 256) and a fully connected trunk network with five layers of 128, 256, 512, 512 and 256 units (ReLU activation). The branch receives windows of $T = 3$ consecutive snapshots of the vorticity measured at 1{,}030 sensor locations (a regular array of $100 \times 10$ points in the wake and 30 points around the rear half of the cylinder), and the trunk encodes the grid coordinates of the point at which the field is evaluated, mapped to $[0, 2\pi] \times [0, \pi]$. The windows are cut from the concatenated sequences, so two test windows span the change from Re = 185 to 195 (and two training windows each change of the training Reynolds number); their errors are $17.3\%$ and $10.0\%$.
The Adam optimizer is used for updating the network parameters (initial learning rate $10^{-3}$, weight decay $10^{-6}$) with the relative $L_2$ loss of each mini-batch of 25 windows. The learning rate is adapted using the ReduceLROnPlateau scheduler (factor 0.5, patience 5 epochs), and training stops after 10 epochs without improvement, with at most 1{,}000 epochs and random seed 12345. No separate validation set was held out: the loss on the test sequences was used for the learning-rate schedule and for early stopping, so the test errors below are not independent of model selection.

The predicted full flow fields for three representative non-dimensional time steps are illustrated in Fig. \ref{fig:ns_reconstruct}. Over the 796 test windows, TRON obtains a relative $L_2$ error of 5.66\% computed over all test snapshots jointly (4.72\% averaged over snapshots), an RMSE of 0.0425 against a root-mean-square vorticity of 0.750, and $R^2 = 0.9968$, with $R^2$ computed about the mean of all test values (single training run). Because the test sequences were used for early stopping, these values are not an independent test. These metrics demonstrate that the TRON architecture can reconstruct full flow fields from sparse flow measurements with sufficient accuracy.

\section{Intrinsic Dimensionality of the Reference Data}
\label{sec:intrinsic}
The reference effective dose fields used throughout this work are generated with EXPACS
under a fixed atmospheric configuration: sea level, an atmospheric depth of
$1033$~g\,cm$^{-2}$, a fixed water fraction in the ground (0.15), and the time-invariant cut-off rigidity
map supplied with the toolkit. Under these settings the dose at a location is a
deterministic function of the static cut-off rigidity and of the scalar solar modulation
index,
\begin{equation}
X(\mathbf{r}, t) \;=\; F\!\left(R_{\mathrm{c}}(\mathbf{r}),\, W(t)\right),
\label{eq:intrinsic}
\end{equation}
so that the temporal variability of the data set carries a single degree of freedom. This
section quantifies that structure, and the corresponding redundancy of the station inputs,
so that the reader can judge what the reported reconstruction accuracy does and does not
establish.

\subsection{Principal component structure of the dose fields}

We formed the complete reference tensor of $8{,}400$ days $\times$ $65{,}341$ grid points,
subtracted the 2001--2023 mean at each grid point, and computed the exact temporal Gram
matrix over \emph{all} grid points by chunked accumulation; no spatial subsampling was
used. The eigenvalue spectrum is given in Table~\ref{tab:intrinsic_pca} and in
Fig.~\ref{fig:intrinsic}a.

\begin{table}[!htbp]
\centering
\caption{Principal component spectrum of the reference dose fields, computed exactly over
the full grid. The participation-ratio effective rank, $(\sum_i \lambda_i)^2 / \sum_i
\lambda_i^2$, is $1.000$.}
\label{tab:intrinsic_pca}
\begin{tabular*}{0.7\textwidth}{@{\extracolsep\fill}lcc@{}}
\toprule
Component & Explained variance ratio & Cumulative \\
\midrule
PC1 & $0.999789$ & $0.999789$ \\
PC2 & $2.079\times10^{-4}$ & $0.9999972$ \\
PC3 & $2.800\times10^{-6}$ & $0.99999999$ \\
PC4 & $7\times10^{-9}$ & $1.000000$ \\
PC5, PC6 & $<10^{-9}$ & $1.000000$ \\
\bottomrule
\end{tabular*}
\end{table}

The day-to-day variation of the global field is therefore a single fixed spatial pattern
scaled by a single time-varying scalar, to within $2\times10^{-4}$ of the total variance.
We note that principal component analysis is a linear decomposition, so this establishes a
low \emph{linear} rank; the further claim that the underlying manifold is one-dimensional
rests on the analysis in the following subsection rather than on Table~\ref{tab:intrinsic_pca}
alone.

\subsection{Recovery of the modulation index from the station inputs}

Sato \cite{sato2015analytical} gives the affine relation $W_s = u_{1,s} C_s + u_{2,s}$
between the count rate $C_s$ of an individual neutron monitor, in count\,min$^{-1}$, and
the modulation index, with the index taken as the mean of the station-wise values.
Applying the published coefficients for Oulu, $u_1 = -0.0931$ and $u_2 = 639$, to our own
input series, after converting its count rates from counts per second to counts per minute and with no fitted parameters, reconstructs a $W$ series spanning $-1.7$ to
$206.4$ with a mean of $46.0$.

Regressing the PC1 coefficient of the dose fields on this reconstructed index gives
$R^2 = 0.99511$ under a quadratic fit; the fit saturates at second order
($R^2 = 0.9537$, $0.99511$ and $0.99513$ at first, second and third order). The additional
components in Table~\ref{tab:intrinsic_pca} are therefore consistent with the curvature of a
one-dimensional manifold embedded in a linear basis rather than additional degrees of freedom.
Fig.~\ref{fig:intrinsic}b shows that days spanning two solar cycles fall close to a single curve; the residuals of the quadratic fit drift slightly from year to year.

\subsection{Redundancy of the station network}

\begin{figure}[!htbp]
\centering
\makebox[\textwidth][c]{\includegraphics[width=180mm]{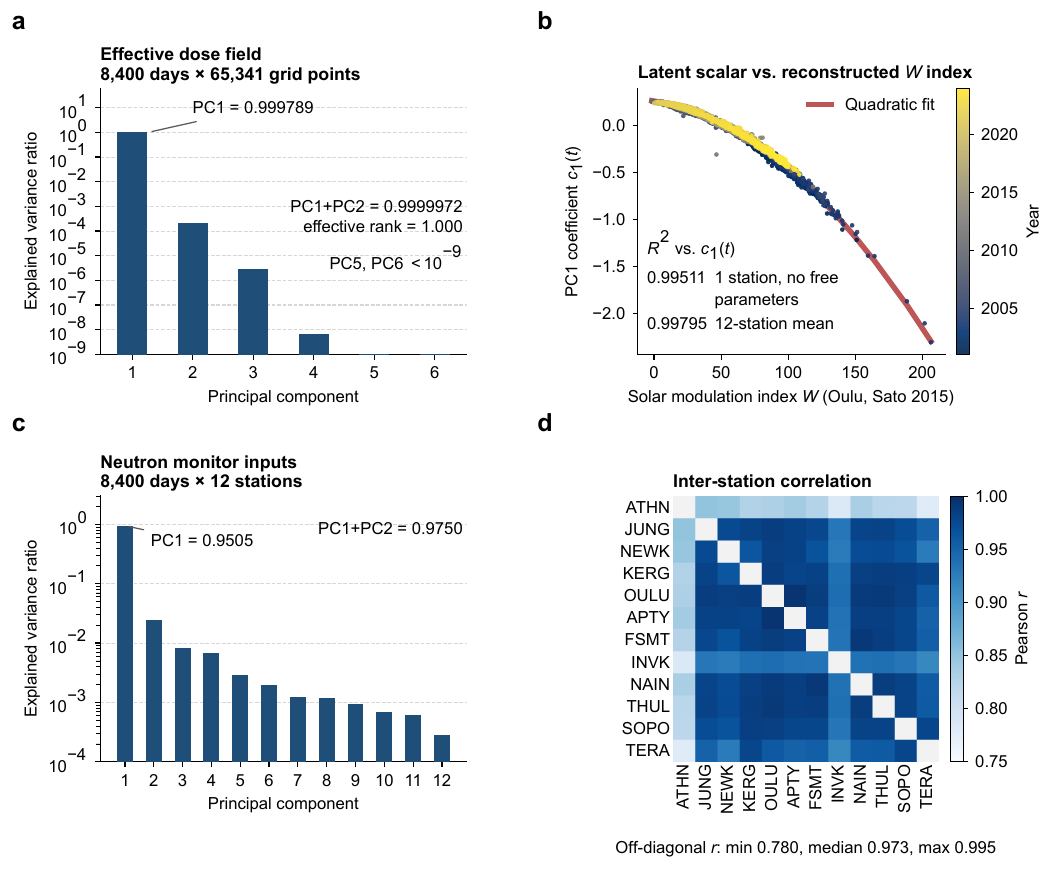}}
\caption{Intrinsic dimensionality of the EXPACS reference data and of the station inputs.
PC$k$, $k$th principal component; effective rank, participation ratio (Table~\ref{tab:intrinsic_pca}).
\textbf{(a)} Principal component spectrum of the reference dose field (centered at each grid point; full grid).
\textbf{(b)} PC1 coefficient $c_1(t)$ against the solar modulation index $W$, reconstructed
from the Oulu count rate (Sato 2015, Table~2; no fitted parameters); points colored by year; red line, quadratic fit. Inset: $R^2$ of a quadratic fit of $c_1$ on $W$ and of a fifth-order fit of $c_1$ on the mean of the twelve raw count rates.
\textbf{(c)} Principal component spectrum of the twelve daily neutron count
series, standardized per station. \textbf{(d)} Inter-station Pearson correlation $r$ of the
raw daily counts; diagonal masked. Station labels, NMDB (Neutron Monitor Database) identifiers (Table~\ref{tab:neutron_stations}).}
\label{fig:intrinsic}
\end{figure}

A principal component analysis of the twelve daily count series, standardized per station
so that sites with higher absolute count rates do not dominate, gives PC1 $= 0.9505$ and
PC1$+$PC2 $= 0.9750$ (Fig.~\ref{fig:intrinsic}c). Off-diagonal inter-station Pearson
correlations of the raw counts range from $0.780$ to $0.995$, with a median of $0.973$
(Fig.~\ref{fig:intrinsic}d). The stations therefore carry substantially the same
information about the modulation state.

What the multiplicity of stations contributes is consistent with variance reduction rather than additional
information about the modulation state. We fitted the same per-location quadratic function of the count rate
on the training period only, once to the mean of the twelve raw count rates and once to the raw count rate of each single station.
The single station increases the relative $L_2$ error by a median factor of
$6.7$: $0.063\%$ from the twelve-station mean against $0.22$--$1.08\%$ from an individual
station. The factor depends on the decoder and on how the stations are combined. For linear, quadratic and cubic decoders applied to raw or per-station standardized count rates, the median factor ranges from $1.2$ to $6.7$, and $6.7$ is the largest of these readings. Single-station performance is markedly heterogeneous, ranging over a factor of
$4.9$ across the network, with Oulu closest to the full network at a factor of $3.5$ and
Newark furthest at $17.3$, so the median rather than the mean is the appropriate summary.
Pooling the stations averages out persistent station-specific deviations that are only weakly correlated between stations. These figures are
reproduced by \texttt{code/analysis/single\_station\_ratio.py}. We therefore do not read the improvement obtained by
pooling as evidence that the stations carry distinct information, and we have not attempted
to separate counting statistics from rigidity-dependent detector response within the
residual. The full network is retained for that variance reduction and for continuity of
operation when individual stations are unavailable, which is consistent with the
construction of the index itself: Sato \cite{sato2015analytical} obtains the daily $W$
series as the mean of the station-wise values rather than from a single monitor.

The same twelve-station quadratic decoder is also more accurate than every TRON variant on the 2023 test year: its $0.063\%$ is below the lowest TRON error in Table~\ref{tab:performance} ($8.52\times10^{-2}\%$, S-GRU at $T = 90$ days) and below S-LSTM at $T = 30$ days ($9.59\times10^{-2}\%$). It uses the one-parameter structure of the reference directly, through a pooled index and a per-point function chosen with that structure in mind. TRON uses no index and no functional form chosen for this structure and learns the mapping from the station window in a single step. The per-point linear regressions of Table~\ref{tab:performance} cannot represent the curvature of $c_1$ in $W$ (Fig.~\ref{fig:intrinsic}b).

\subsection{Scope of the reported accuracy}
As a check that the reported accuracy reflects a response to the inputs rather than
reproduction of the time-averaged field, we compare against climatology. Predicting the
2001--2021 mean field for every test day gives a relative $L_2$ error of $2.55\%$; the
2023 mean field, a constant field that minimizes the squared error over the test year and is unavailable in
operation because it requires the test period in advance, gives $1.01\%$. TRON (S-GRU, $T = 90$ days, the most accurate configuration of Table~\ref{tab:performance}; a single training run) gives
$0.085\%$, a skill score $1 - (e/e_{\mathrm{ref}})^2$ of $0.999$ against the operational climatology and $0.993$
against the constant-field reference. The model therefore tracks the temporal modulation
rather than reproducing the time-averaged field. These values are computed by \texttt{code/analysis/climatology\_skill.py}.

The reconstruction accuracies reported in this work should be read against this structure.
The reference fields form a one-parameter family, and the modulation index that
parameterizes them is recoverable from the station inputs by construction. The comparisons
between architectures presented in Sections~\ref{sec:baselines} and~\ref{sec:performance} are therefore comparisons
under identical data, and the ratio between the number of output coordinates and the
number of input sensors does not by itself characterize the inference problem. The low
temporal rank documented here is a property of the reference configuration, and any benchmark derived from EXPACS under fixed atmospheric settings inherits it.

\clearpage
\section{Measured-data experiments}
\label{sec:measured}
Supplementary Table~\ref{tab:meas_datasets} summarizes the three datasets used in the main text. Supplementary Tables~\ref{tab:meas_ghcnd} and~\ref{tab:meas_ghcnd_paired} give every branch and baseline for GHCN daily maximum temperature and the paired comparisons per station split; Supplementary Tables~\ref{tab:meas_pressure} and~\ref{tab:meas_pressure_tests} give those for PeakWeather pressure to 2\,m temperature and its pre-registered tests; Supplementary Table~\ref{tab:meas_dh} gives those for PeakWeather degree-hours. Supplementary Table~\ref{tab:meas_negative} lists the measured datasets that were attempted but are not used as evidence, with the reason, and Supplementary Table~\ref{tab:meas_runs} gives the training recipe, the run inventory and the compute. All four branches are reported for every dataset and clock setting; S-LSTM was declared the primary variant before the pressure runs and before GHCN splits 1 and 2 were trained. Every value in these tables is generated by \texttt{code/analysis/realdata/make\_si\_tables\_measured.py} from the run files and baseline outputs.

\subsection{Data, station selection and windows}
GHCN-Daily TMAX~\cite{menne2012ghcnd,menne2012ghcnddata} is used for contiguous-US stations with near-complete records over 2003--2023, without quality-flagged values. In each of 3 station splits, 12 input stations are chosen by farthest-point sampling in longitude and latitude, 100 held-out stations are drawn at random, and the rest are training targets; short input gaps are interpolated, and missing targets are masked in the loss and scores. Windows are assigned by their last day (training to 2018, validation 2019--2020, test 2021--2023). PeakWeather~\cite{zambon2025peakweather} MeteoSwiss data are averaged to hourly values, with 40 held-out stations drawn at random. The 30 pressure inputs are station pressures (not reduced to sea level) from barometers of stable height, chosen by farthest-point traversal in easting, northing and 100 times the barometer height from a pool that excludes the held-out stations, their sites and barometers within 1\,km of them; 3 draws differ in the first pick. Heating degree-hours are $\sum_h \max(18\,^\circ\mathrm{C} - \theta_h, 0)$ over the last 24, 48 or 72 hourly temperatures $\theta_h$. The degree-hours runs and the thermometer reference use 30 thermometer inputs. PeakWeather windows are assigned by their last hour (training to 2022, validation 2023, test from 2024 (degree-hours from 31 December 2023; Table~\ref{tab:meas_datasets})); pressure windows lie inside one split.

\subsection{Networks and training}
The branches (S-LSTM, S-GRU, the static-branch DeepONet reading the last step, and a flattened-window DeepONet whose feed-forward branch reads the whole window) have 4 layers of 128 units; the trunk maps (longitude, latitude, elevation) through three layers of 128 units, and the output is $\mathbf{b}\cdot\mathbf{t}+\beta$. Inputs are min--max scaled per channel on training windows. No per-point range exists at a held-out station, so the target uses one global min--max range from the training stations over the training period. Trunk coordinates are min--max scaled over training and held-out coordinates (GHCN, degree-hours; no target values) or training stations only (pressure). Calendar channels are the sine and cosine of the day of year and, for PeakWeather, of the hour. Training uses Adam (learning rate $10^{-3}$, weight decay $10^{-6}$, halving the learning rate when the validation loss had not improved for more than five epochs), batches of 32, at most 300 epochs and early stopping with patience 20 on the training stations' validation loss, keeping the best checkpoint; each configuration uses seeds 1234567, 11 and 22.

\subsection{Baselines}
The two-stage linear model fits a ridge regression from the input window (with calendar harmonics when the network has calendar channels) to the training stations, with the penalty chosen on the validation period, and decodes to held-out coordinates with a Nadaraya--Watson kernel in longitude, latitude and scaled elevation (lapse-rate-detrended for pressure). Kernel family, bandwidth and elevation scale are chosen by fivefold cross-validation grouped by training station on the training period, so held-out values are read only for scoring. The non-learned GHCN baselines use the last day without calendar: nearest input, inverse-distance weighting, and a daily plane in longitude, latitude and elevation. For pressure, the calendar-only model is the two-stage pipeline with calendar harmonics as its only input, and the hypsometric baseline derives layer-mean temperatures from barometer pairs at different heights and weights them to each station with a height-difference term. Rows that use held-out data are references only. In GHCN split 0, held-out station \GhcnPairSLstmZeroTopStation{} contributes \GhcnPairSLstmZeroTopSharePct\% of the paired difference between S-LSTM (with calendar channels) and the two-stage model (Supplementary Table~\ref{tab:meas_ghcnd_paired}).

\subsection{Displays in main-text Fig.~8 and test-window reach-back}
The map in main-text Fig.~8a (\FigFiveGhcnDay, the median-RMSE day of \FigFiveGhcnNTestDays{} test days; \FigFiveGridOutsidePct\% of cells outside the trunk's training range) comes from the saved network evaluated on the GPU that trained it, which reproduces the saved held-out predictions exactly (maximum difference \FigFiveGridReproMax\,$^\circ$C); a CPU evaluation does not. Main-text Fig.~8b spans $\pm 6\,^\circ$C. Main-text Fig.~8f shows the lower-median-RMSE test hour (the lower of the two middle values; rule fixed before mapping); \PressGridExtrapPct\% of its cells extrapolate beyond the fit stations' coordinate or height range, and its color-bar arrow marks extrapolated cells below 0\,$^\circ$C. The station in main-text Fig.~8h is the lower-median held-out station by S-LSTM RMSE, shown over the lower-median of \FigFiveDhNBlocks{} 28-day blocks with at least 90\% of hours (default seeds). A test window is one whose end falls after the validation cut-off, so its inputs can predate it. For GHCN this holds for \GhcnReachBackZero, \GhcnReachBackOne{} and \GhcnReachBackTwo{} of the \GhcnTestDaysZero, \GhcnTestDaysOne{} and \GhcnTestDaysTwo{} test windows (splits 0--2), of which \GhcnShareValZero, \GhcnShareValOne{} and \GhcnShareValTwo{} share input days with a validation window. For degree-hours it holds for \DhTestNBeforeCut{} of \DhTestN{} test windows (\DhTestPctBeforeCut\%), none of which shares an hour with a validation window, because of a \DhGapHoursMissing-hour data gap before \DhValCutoff. Degree-hours test windows end \PeakTestFirst{} to \PeakTestLast; the pressure test set has \PressNTestWindows{} windows from \PressTestFirst.

\subsection{Tables}
\begin{table}[!ht]
\centering
\footnotesize
\caption{\textbf{Measured datasets used in main-text Fig.~8.} Input stations feed the branch; training stations are the targets of the loss; held-out stations enter neither the loss, any scaler nor early stopping. Distances are great-circle, except the pre-registered pressure-input distances (Swiss LV95 plane). PC1 variance share: variance of the training-station field over the training period (886 GHCN stations, range over the 3 splits; 105 PeakWeather stations) explained by the first principal component (PC) after removing each station's mean (raw) or harmonic climatology (anomaly; three annual harmonics, plus for PeakWeather three diurnal harmonics and their products). Unobserved GHCN targets are interpolated in time for this analysis only. The pressure time axis omits hours failing pressure quality control. QFE, station-level pressure; val., validation.}
\label{tab:meas_datasets}
\begin{tabularx}{\textwidth}{@{}>{\raggedright\arraybackslash}p{30mm}>{\raggedright\arraybackslash}X>{\raggedright\arraybackslash}X>{\raggedright\arraybackslash}X@{}}
\toprule
& GHCN TMAX & PeakWeather pressure $\to$ 2\,m temperature & PeakWeather degree-hours \\
\midrule
Target (unit) & daily maximum temperature, TMAX ($^\circ$C) & hourly 2\,m temperature ($^\circ$C) & heating degree-hours below 18\,$^\circ$C summed over the past 24, 48 or 72\,h ($^\circ$C\,h); instantaneous 2\,m temperature as control \\
Branch input & TMAX at 12 input stations, 14-day window & station pressure (QFE) at 30 barometers, 24\,h window; no thermometer & 2\,m temperature at 30 thermometers, 72\,h window \\
Indirect input & no (same quantity) & yes (pressure for temperature) & no (same quantity; target is a window functional of it) \\
Source; license & GHCN-Daily\cite{menne2012ghcnd,menne2012ghcnddata}; CC0-1.0 & PeakWeather\cite{zambon2025peakweather}; CC-BY-4.0, attribution to MeteoSwiss & PeakWeather\cite{zambon2025peakweather}; CC-BY-4.0, attribution to MeteoSwiss \\
Stations: input / held-out / training & 12 / 100 / 886 & 30 / 40 / 105 & 30 / 40 / 105 \\
Station splits & 3 random splits (0, 1, 2) & 1 held-out split; 3 overlapping barometer draws (0, 1, 2) & 1 held-out split \\
Training seeds & 3 per cell (default, 11, 22) & 3 per cell (default, 11, 22) & 3 per cell (default, 11, 22) \\
Training period & 2003-01-01 / 2003-05-09 / 2003-01-01 (splits 0 / 1 / 2) to 2018-12-31 & 2017-01-19 to 2022-12-31 & 2017-01-19 to 2022-12-31 \\
Validation period & to 2020-12-31 & to 2023-12-31 & to 2023-12-31 \\
Test period & 2021-01-01 to 2023-12-31 & 2024-01-01 to 2025-10-13 & 2023-12-31 to 2025-10-13 \\
Windows: train / val.\ / test & 5{,}009 / 658 / 847 ; 5{,}100 / 646 / 996 ; 5{,}563 / 731 / 885 (splits 0; 1; 2) & 36{,}701 / 5{,}479 / 11{,}086 & 32{,}576 / 4{,}630 / 9{,}974 \\
Median distance, held-out station to nearest input (km) & 402 / 409 / 387 (splits 0 / 1 / 2) & 21.2 / 21.0 / 18.8 (draws 0 / 1 / 2) & 19.9 \\
Median distance, held-out station to nearest training station (km) & 36 / 39 / 39 & 10.9 & 10.9 \\
PC1 variance share, raw / anomaly & 0.787--0.792 / 0.230--0.235 & 0.922 / 0.739 & 0.921 / 0.739 \\
PCs for 90\% of variance, raw / anomaly & 5 / 80--82 & 1 / 6 & 1 / 6 \\
\bottomrule
\end{tabularx}
\end{table}

\begin{table}[!htbp]
\centering
\small
\caption{\textbf{GHCN TMAX: every branch and baseline at the 100 held-out stations.} RMSE, root-mean-square error over all observed held-out station-days of the test period; pooled $R^2 = 1 - \mathrm{SSE}/\mathrm{SST}$ about the single mean of all held-out test values. Network rows: $n = 9$ runs (3 station splits $\times$ 3 training seeds); split columns, mean over seeds; last two columns, mean $\pm$ s.d.\ over the $n$ runs (seed and split variation) or fits. Baseline rows: one deterministic fit per split ($n = 3$). S-LSTM (pre-declared primary) and S-GRU, TRON with a recurrent branch; flattened-window and static-branch DeepONets, feed-forward branch reading the whole window or the last step. Clock: sine and cosine of the day of year as extra inputs. IDW, inverse-distance weighting with power $p$. Baselines are described in Section~\ref{sec:measured}. ORACLE rows use the held-out stations' own data; they are references, not attainable baselines.}
\label{tab:meas_ghcnd}
\begin{tabular}{@{}llrcccccc@{}}
\toprule
& & & \multicolumn{3}{c}{RMSE by split ($^\circ$C)} & & \\
\cmidrule(lr){4-6}
method & clock & $n$ & 0 & 1 & 2 & RMSE ($^\circ$C) & pooled $R^2$ \\
\midrule
S-LSTM & off & 9 & 3.430 & 3.459 & 3.571 & 3.487 $\pm$ 0.066 & 0.912 $\pm$ 0.003 \\
S-LSTM & on & 9 & 3.423 & 3.434 & 3.489 & 3.449 $\pm$ 0.036 & 0.914 $\pm$ 0.004 \\
S-GRU & off & 9 & 3.383 & 3.437 & 3.562 & 3.461 $\pm$ 0.080 & 0.914 $\pm$ 0.002 \\
S-GRU & on & 9 & 3.394 & 3.386 & 3.518 & 3.433 $\pm$ 0.068 & 0.915 $\pm$ 0.003 \\
flattened-window DeepONet & off & 9 & 3.448 & 3.481 & 3.562 & 3.497 $\pm$ 0.058 & 0.912 $\pm$ 0.003 \\
flattened-window DeepONet & on & 9 & 3.443 & 3.445 & 3.560 & 3.483 $\pm$ 0.065 & 0.912 $\pm$ 0.003 \\
static-branch DeepONet & off & 9 & 3.506 & 3.552 & 3.631 & 3.563 $\pm$ 0.057 & 0.908 $\pm$ 0.003 \\
static-branch DeepONet & on & 9 & 3.501 & 3.568 & 3.568 & 3.546 $\pm$ 0.040 & 0.909 $\pm$ 0.005 \\
\midrule
two-stage linear & off & 3 & 3.549 & 3.512 & 3.578 & 3.546 $\pm$ 0.033 & 0.909 $\pm$ 0.005 \\
two-stage linear & on & 3 & 3.523 & 3.482 & 3.571 & 3.525 $\pm$ 0.045 & 0.910 $\pm$ 0.005 \\
plane fit, last day & -- & 3 & 5.078 & 5.130 & 5.426 & 5.212 $\pm$ 0.188 & 0.804 $\pm$ 0.001 \\
IDW ($p = 2$), last day & -- & 3 & 5.110 & 5.756 & 5.881 & 5.582 $\pm$ 0.413 & 0.775 $\pm$ 0.026 \\
nearest input, last day & -- & 3 & 6.522 & 6.386 & 6.896 & 6.601 $\pm$ 0.264 & 0.685 $\pm$ 0.008 \\
\midrule
ORACLE two-stage & off & 3 & 3.549 & 3.512 & 3.551 & 3.537 $\pm$ 0.022 & 0.909 $\pm$ 0.006 \\
ORACLE two-stage & on & 3 & 3.523 & 3.482 & 3.546 & 3.517 $\pm$ 0.033 & 0.911 $\pm$ 0.005 \\
ORACLE climatology & -- & 3 & 5.103 & 5.235 & 5.165 & 5.168 $\pm$ 0.066 & 0.807 $\pm$ 0.015 \\
ORACLE station mean & -- & 3 & 10.395 & 9.902 & 10.556 & 10.284 $\pm$ 0.340 & 0.236 $\pm$ 0.046 \\
\bottomrule
\end{tabular}
\end{table}

\begin{table}[!htbp]
\centering
\footnotesize
\caption{\textbf{GHCN TMAX: paired differences per station split.} $\Delta$RMSE, first minus second model's held-out RMSE on the same station-days, mean over 3 training seeds ($^\circ$C; negative favors the first); two-stage, the CV-tuned two-stage linear model with the same clock (Supplementary Table~\ref{tab:meas_ghcnd}; CV, cross-validation). 95\% CI, percentile interval of a paired bootstrap over the 100 held-out stations and calendar-week blocks (2000 replicates, seed 0). Stations better: held-out stations where the first model's seed-mean RMSE is lower. Outcome, band $\delta = 0.10\,^\circ$C: Better, CI upper bound $< 0$ and $\Delta < -\delta$; Worse, CI lower bound $> 0$ and $\Delta > \delta$; Tie, CI within $[-\delta, \delta]$; Unresolved, otherwise. The rule was pre-registered for the pressure dataset; for GHCN it is applied by analogy, not pre-registered. Each split is one station partition, tested separately without correction for the 24 two-stage contrasts. Branches as in Supplementary Table~\ref{tab:meas_ghcnd}.}
\label{tab:meas_ghcnd_paired}
\begin{tabular}{@{}lllrcrl@{}}
\toprule
contrast & clock & split & $\Delta$RMSE ($^\circ$C) & 95\% CI & stations better & outcome \\
\midrule
S-LSTM $-$ two-stage & off & 0 & $-$0.119 & [$-$0.270, $-$0.008] & 50/100 & Better \\
S-LSTM $-$ two-stage & off & 1 & $-$0.052 & [$-$0.113, $+$0.008] & 53/100 & Unresolved \\
S-LSTM $-$ two-stage & off & 2 & $-$0.008 & [$-$0.101, $+$0.081] & 43/100 & Unresolved \\
S-LSTM $-$ two-stage & on & 0 & $-$0.101 & [$-$0.241, $+$0.001] & 50/100 & Unresolved \\
S-LSTM $-$ two-stage & on & 1 & $-$0.048 & [$-$0.108, $+$0.013] & 59/100 & Unresolved \\
S-LSTM $-$ two-stage & on & 2 & $-$0.082 & [$-$0.185, $+$0.012] & 45/100 & Unresolved \\
\addlinespace
S-GRU $-$ two-stage & off & 0 & $-$0.166 & [$-$0.321, $-$0.058] & 58/100 & Better \\
S-GRU $-$ two-stage & off & 1 & $-$0.075 & [$-$0.131, $-$0.018] & 61/100 & Unresolved \\
S-GRU $-$ two-stage & off & 2 & $-$0.016 & [$-$0.111, $+$0.068] & 37/100 & Unresolved \\
S-GRU $-$ two-stage & on & 0 & $-$0.129 & [$-$0.270, $-$0.029] & 54/100 & Better \\
S-GRU $-$ two-stage & on & 1 & $-$0.095 & [$-$0.149, $-$0.039] & 63/100 & Unresolved \\
S-GRU $-$ two-stage & on & 2 & $-$0.053 & [$-$0.148, $+$0.035] & 45/100 & Unresolved \\
\addlinespace
flattened-window $-$ two-stage & off & 0 & $-$0.101 & [$-$0.262, $+$0.017] & 45/100 & Unresolved \\
flattened-window $-$ two-stage & off & 1 & $-$0.031 & [$-$0.096, $+$0.033] & 53/100 & Tie \\
flattened-window $-$ two-stage & off & 2 & $-$0.016 & [$-$0.108, $+$0.070] & 39/100 & Unresolved \\
flattened-window $-$ two-stage & on & 0 & $-$0.080 & [$-$0.232, $+$0.032] & 46/100 & Unresolved \\
flattened-window $-$ two-stage & on & 1 & $-$0.036 & [$-$0.097, $+$0.027] & 53/100 & Tie \\
flattened-window $-$ two-stage & on & 2 & $-$0.011 & [$-$0.108, $+$0.074] & 39/100 & Unresolved \\
\addlinespace
static-branch $-$ two-stage & off & 0 & $-$0.043 & [$-$0.206, $+$0.076] & 43/100 & Unresolved \\
static-branch $-$ two-stage & off & 1 & $+$0.041 & [$-$0.037, $+$0.123] & 41/100 & Unresolved \\
static-branch $-$ two-stage & off & 2 & $+$0.053 & [$-$0.041, $+$0.148] & 40/100 & Unresolved \\
static-branch $-$ two-stage & on & 0 & $-$0.022 & [$-$0.172, $+$0.089] & 42/100 & Unresolved \\
static-branch $-$ two-stage & on & 1 & $+$0.086 & [$+$0.012, $+$0.165] & 36/100 & Unresolved \\
static-branch $-$ two-stage & on & 2 & $-$0.003 & [$-$0.107, $+$0.097] & 46/100 & Unresolved \\
\addlinespace
S-LSTM $-$ S-GRU & off & 0 & $+$0.047 & [$+$0.024, $+$0.071] & 29/100 & Tie \\
S-LSTM $-$ S-GRU & off & 1 & $+$0.022 & [$+$0.006, $+$0.039] & 32/100 & Tie \\
S-LSTM $-$ S-GRU & off & 2 & $+$0.009 & [$-$0.024, $+$0.047] & 56/100 & Tie \\
S-LSTM $-$ S-GRU & on & 0 & $+$0.028 & [$+$0.009, $+$0.047] & 32/100 & Tie \\
S-LSTM $-$ S-GRU & on & 1 & $+$0.047 & [$+$0.021, $+$0.079] & 34/100 & Tie \\
S-LSTM $-$ S-GRU & on & 2 & $-$0.029 & [$-$0.067, $+$0.005] & 53/100 & Tie \\
S-LSTM $-$ flattened-window & off & 0 & $-$0.017 & [$-$0.039, $+$0.005] & 62/100 & Tie \\
S-LSTM $-$ flattened-window & off & 1 & $-$0.021 & [$-$0.045, $+$0.002] & 63/100 & Tie \\
S-LSTM $-$ flattened-window & off & 2 & $+$0.008 & [$-$0.027, $+$0.048] & 57/100 & Tie \\
S-LSTM $-$ flattened-window & on & 0 & $-$0.021 & [$-$0.046, $+$0.005] & 69/100 & Tie \\
S-LSTM $-$ flattened-window & on & 1 & $-$0.012 & [$-$0.039, $+$0.021] & 71/100 & Tie \\
S-LSTM $-$ flattened-window & on & 2 & $-$0.071 & [$-$0.105, $-$0.039] & 69/100 & Unresolved \\
S-LSTM $-$ static-branch & off & 0 & $-$0.076 & [$-$0.125, $-$0.024] & 58/100 & Unresolved \\
S-LSTM $-$ static-branch & off & 1 & $-$0.093 & [$-$0.142, $-$0.046] & 66/100 & Unresolved \\
S-LSTM $-$ static-branch & off & 2 & $-$0.060 & [$-$0.116, $-$0.004] & 57/100 & Unresolved \\
S-LSTM $-$ static-branch & on & 0 & $-$0.079 & [$-$0.121, $-$0.034] & 63/100 & Unresolved \\
S-LSTM $-$ static-branch & on & 1 & $-$0.134 & [$-$0.185, $-$0.079] & 75/100 & Better \\
S-LSTM $-$ static-branch & on & 2 & $-$0.079 & [$-$0.127, $-$0.030] & 59/100 & Unresolved \\
\bottomrule
\end{tabular}
\end{table}

\begin{table}[!htbp]
\centering
\footnotesize
\caption{\textbf{PeakWeather pressure $\to$ 2\,m temperature: every branch, control and baseline at the 40 held-out stations.} RMSE, pooled $R^2$ and branches as in Supplementary Table~\ref{tab:meas_ghcnd}; 11{,}086 hourly test windows. Pressure-input network rows: $n = 9$ runs (3 barometer draws $\times$ 3 training seeds; one held-out station set); thermometer-input rows: $n = 3$ seeds. Baseline rows: one deterministic fit per draw ($n = 3$) or in total ($n = 1$, no s.d.). Mean $\pm$ s.d.\ over the $n$ runs or fits; range, minimum--maximum. Clock: sine and cosine of hour of day and day of year (networks); annual and diurnal harmonics (linear models). Time-shuffled: station channels permuted in time, calendar aligned (pre-registered control). H2: Supplementary Table~\ref{tab:meas_pressure_tests}. NW, Nadaraya--Watson decoder without lapse-rate detrending; last hour, last hour of the window only. Italic rows use held-out data or true temperatures and are references, not attainable baselines: ORACLE ridge, fitted to held-out targets; ORACLE two-stage, decoder tuned on them; the last row decodes the true training-station temperatures.}
\label{tab:meas_pressure}
\setlength{\tabcolsep}{4pt}
\begin{tabular}{@{}lllrccc@{}}
\toprule
method & input & clock & $n$ & RMSE ($^\circ$C) & range ($^\circ$C) & pooled $R^2$ \\
\midrule
S-LSTM & pressure & off & 9 & 1.897 $\pm$ 0.034 & 1.853--1.964 & 0.949 $\pm$ 0.002 \\
S-LSTM & pressure & on & 9 & 1.866 $\pm$ 0.045 & 1.802--1.952 & 0.950 $\pm$ 0.002 \\
S-GRU & pressure & off & 9 & 1.891 $\pm$ 0.036 & 1.844--1.959 & 0.949 $\pm$ 0.002 \\
S-GRU & pressure & on & 9 & 1.832 $\pm$ 0.026 & 1.787--1.870 & 0.952 $\pm$ 0.001 \\
flattened-window DeepONet & pressure & off & 9 & 2.008 $\pm$ 0.070 & 1.924--2.132 & 0.942 $\pm$ 0.004 \\
flattened-window DeepONet & pressure & on & 9 & 2.104 $\pm$ 0.075 & 1.988--2.227 & 0.937 $\pm$ 0.004 \\
static-branch DeepONet & pressure & off & 9 & 2.036 $\pm$ 0.051 & 1.963--2.106 & 0.941 $\pm$ 0.003 \\
static-branch DeepONet & pressure & on & 9 & 1.912 $\pm$ 0.029 & 1.856--1.944 & 0.948 $\pm$ 0.002 \\
\addlinespace
S-LSTM, time-shuffled & pressure & off & 9 & 7.810 $\pm$ 0.010 & 7.800--7.832 & 0.127 $\pm$ 0.002 \\
S-LSTM, time-shuffled & pressure & on & 9 & 3.965 $\pm$ 0.055 & 3.902--4.036 & 0.775 $\pm$ 0.006 \\
\addlinespace
S-LSTM & thermometer & off & 3 & 1.837 $\pm$ 0.119 & 1.728--1.964 & 0.952 $\pm$ 0.006 \\
S-LSTM & thermometer & on & 3 & 1.789 $\pm$ 0.150 & 1.694--1.962 & 0.954 $\pm$ 0.008 \\
S-GRU & thermometer & off & 3 & 1.923 $\pm$ 0.147 & 1.754--2.016 & 0.947 $\pm$ 0.008 \\
S-GRU & thermometer & on & 3 & 1.856 $\pm$ 0.182 & 1.687--2.049 & 0.950 $\pm$ 0.010 \\
flattened-window DeepONet & thermometer & off & 3 & 1.697 $\pm$ 0.037 & 1.664--1.737 & 0.959 $\pm$ 0.002 \\
flattened-window DeepONet & thermometer & on & 3 & 1.751 $\pm$ 0.054 & 1.695--1.803 & 0.956 $\pm$ 0.003 \\
static-branch DeepONet & thermometer & off & 3 & 1.629 $\pm$ 0.008 & 1.620--1.636 & 0.962 $\pm$ 0.000 \\
static-branch DeepONet & thermometer & on & 3 & 1.689 $\pm$ 0.043 & 1.639--1.720 & 0.959 $\pm$ 0.002 \\
\midrule
two-stage linear (H2 comparator) & pressure & on & 3 & 1.906 $\pm$ 0.016 & 1.893--1.925 & 0.948 $\pm$ 0.001 \\
two-stage linear & pressure & off & 3 & 1.944 $\pm$ 0.029 & 1.913--1.969 & 0.946 $\pm$ 0.002 \\
two-stage linear, NW decoder & pressure & on & 3 & 1.986 $\pm$ 0.024 & 1.964--2.012 & 0.944 $\pm$ 0.001 \\
two-stage linear, NW decoder & pressure & off & 3 & 2.028 $\pm$ 0.037 & 1.986--2.058 & 0.941 $\pm$ 0.002 \\
two-stage linear, last hour & pressure & on & 3 & 1.972 $\pm$ 0.027 & 1.951--2.002 & 0.944 $\pm$ 0.002 \\
hypsometric & pressure & on & 3 & 2.710 $\pm$ 0.033 & 2.680--2.745 & 0.895 $\pm$ 0.003 \\
hypsometric & pressure & off & 3 & 3.469 $\pm$ 0.094 & 3.412--3.578 & 0.828 $\pm$ 0.009 \\
calendar-only linear & none & on & 1 & 3.940 & -- & 0.778 \\
calendar-only, NW decoder & none & on & 1 & 3.966 & -- & 0.775 \\
training-station mean, decoded & none & off & 1 & 7.810 & -- & 0.128 \\
two-stage linear & thermometer & on & 1 & 1.643 & -- & 0.961 \\
two-stage linear & thermometer & off & 1 & 1.642 & -- & 0.961 \\
\addlinespace
\textit{ORACLE ridge} & pressure & on & 3 & 1.493 $\pm$ 0.023 & 1.476--1.519 & 0.968 $\pm$ 0.001 \\
\textit{ORACLE two-stage} & pressure & on & 3 & 1.934 $\pm$ 0.019 & 1.921--1.957 & 0.946 $\pm$ 0.001 \\
\textit{ORACLE ridge} & thermometer & on & 1 & 1.038 & -- & 0.985 \\
\textit{ORACLE calendar climatology} & none & on & 1 & 3.814 & -- & 0.792 \\
\textit{ORACLE own test-period mean} & none & off & 1 & 7.708 & -- & 0.150 \\
\textit{true training-station T, decoded} & -- & -- & 1 & 1.551 & -- & 0.966 \\
\bottomrule
\end{tabular}
\end{table}

\begin{table}[!htbp]
\centering
\footnotesize
\caption{\textbf{PeakWeather pressure $\to$ 2\,m temperature: pre-registered tests.} $R_T$, RMSE of the primary cell (S-LSTM, pressure, clock on; mean of 9 runs). S1 (sanity): $R_T$ against $0.9\times$ the calendar-only linear RMSE. H1: $R_T$ minus clock-on time-shuffled S-LSTM, paired by draw and seed. H2: $R_T$ minus the CV-tuned two-stage linear RMSE (same inputs and clock); outcome classes as in Supplementary Table~\ref{tab:meas_ghcnd_paired}, $\delta$ in column thresh.\ (threshold). The pooled H2 row decides; per-draw rows are descriptive; H2-type rows (secondary) cover other branches and clock off. H2a: the same contrast at the 105 training stations against the ridge stage; H2 credits pressure only if H2a favors TRON. H2b: pressure minus thermometer margin over the two-stage model; H1 and H2b are supported if the CI is below 0. H3: minimum time-shuffled RMSE against a pre-set floor (pass if above). H4: thermometer- minus pressure-input mean RMSE, untested; REVERSAL, thermometer less accurate. H5: last-hour minus 24\,h-window two-stage RMSE (single fits); no window claim. 95\% CI as in Supplementary Table~\ref{tab:meas_ghcnd_paired} (105 stations for H2a). Runs, network runs used; stations, held-out stations (of 40) where the averaged network RMSE is below the comparator's (--, not computed). Primary-cell seed s.d.\ (pooled within draw): 0.048\,$^\circ$C. Flat.-window, flattened-window DeepONet; static-branch, static-branch DeepONet; st., stations.}
\label{tab:meas_pressure_tests}
\setlength{\tabcolsep}{4pt}
\begin{tabularx}{\textwidth}{@{}l>{\raggedright\arraybackslash}Xrrcrr>{\raggedright\arraybackslash}p{19mm}@{}}
\toprule
test & cell & runs & value ($^\circ$C) & 95\% CI & thresh. & stations & outcome \\
\midrule
S1 & S-LSTM pressure, clock on & 9 & 1.866 & -- & 3.546 & -- & pass \\
H1 & S-LSTM pressure, clock on & 18 & $-$2.098 & [$-$2.442, $-$1.767] & 0.000 & -- & supported \\
H2 & S-LSTM pressure, clock on & 9 & $-$0.040 & [$-$0.151, $+$0.073] & 0.100 & 27/40 & Unresolved \\
H2, per draw & S-LSTM pressure, clock on, draw 0 & 3 & $-$0.006 & [$-$0.117, $+$0.104] & 0.100 & -- & Unresolved \\
H2, per draw & S-LSTM pressure, clock on, draw 1 & 3 & $-$0.041 & [$-$0.157, $+$0.078] & 0.100 & -- & Unresolved \\
H2, per draw & S-LSTM pressure, clock on, draw 2 & 3 & $-$0.073 & [$-$0.212, $+$0.064] & 0.100 & -- & Unresolved \\
H2-type & S-LSTM pressure, clock off & 9 & $-$0.048 & [$-$0.163, $+$0.068] & 0.100 & 25/40 & Unresolved \\
H2-type & S-GRU pressure, clock on & 9 & $-$0.074 & [$-$0.177, $+$0.035] & 0.100 & 29/40 & Unresolved \\
H2-type & S-GRU pressure, clock off & 9 & $-$0.054 & [$-$0.168, $+$0.060] & 0.100 & 26/40 & Unresolved \\
H2-type & static-branch pressure, clock on & 9 & $+$0.005 & [$-$0.107, $+$0.129] & 0.100 & 25/40 & Unresolved \\
H2-type & static-branch pressure, clock off & 9 & $+$0.091 & [$-$0.015, $+$0.195] & 0.100 & 15/40 & Unresolved \\
H2-type & flat.-window pressure, clock on & 9 & $+$0.198 & [$+$0.091, $+$0.302] & 0.100 & 12/40 & Worse \\
H2-type & flat.-window pressure, clock off & 9 & $+$0.063 & [$-$0.038, $+$0.160] & 0.100 & 18/40 & Unresolved \\
H2a & S-LSTM pressure, clock on, 105 training st. & 9 & $-$0.081 & [$-$0.120, $-$0.043] & 0.000 & -- & favors TRON \\
H2a-type & S-LSTM pressure, clock off, 105 training st. & 9 & $-$0.060 & [$-$0.100, $-$0.021] & 0.000 & -- & favors TRON \\
H2b & S-LSTM, clock on, pressure vs thermometer inputs & 12 & $-$0.186 & [$-$0.281, $-$0.101] & 0.000 & -- & supported \\
H3 & S-LSTM time-shuffled, clock off & 9 & 7.800 & -- & 7.000 & -- & pass \\
H3 & S-LSTM time-shuffled, clock on & 9 & 3.902 & -- & 3.569 & -- & pass \\
H4 & S-LSTM, clock off & 12 & $-$0.060 & -- & 0.000 & -- & as expected (th $\le$ p) \\
H4 & S-LSTM, clock on & 12 & $-$0.078 & -- & 0.000 & -- & as expected (th $\le$ p) \\
H4 & S-GRU, clock off & 12 & $+$0.033 & -- & 0.000 & -- & REVERSAL \\
H4 & S-GRU, clock on & 12 & $+$0.024 & -- & 0.000 & -- & REVERSAL \\
H4 & static-branch, clock off & 12 & $-$0.407 & -- & 0.000 & -- & as expected (th $\le$ p) \\
H4 & static-branch, clock on & 12 & $-$0.223 & -- & 0.000 & -- & as expected (th $\le$ p) \\
H4 & flat.-window, clock off & 12 & $-$0.310 & -- & 0.000 & -- & as expected (th $\le$ p) \\
H4 & flat.-window, clock on & 12 & $-$0.353 & -- & 0.000 & -- & as expected (th $\le$ p) \\
H5 & linear models, last hour $-$ window & -- & $+$0.065 & -- & -- & -- & no window claim \\
\bottomrule
\end{tabularx}
\end{table}

\begin{table}[!htbp]
\centering
\small
\caption{\textbf{PeakWeather degree-hours: every branch, horizon and clock setting at the 40 held-out stations.} Heating degree-hours: hourly deficit of 2\,m temperature below 18\,$^\circ$C summed over the past 24, 48 or 72\,h; instantaneous temperature is the control. One station split; $n = 3$ training seeds per cell; mean $\pm$ s.d.\ over seeds. RMSE as in Supplementary Table~\ref{tab:meas_ghcnd}, over 9{,}974 hourly test windows; $1 - R^2$, unexplained fraction of the pooled held-out target variance (target s.d.: 8.46\,$^\circ$C, 165.0, 325.5, 483.5\,$^\circ$C\,h for instantaneous, 24, 48, 72\,h). The 48\,h horizon was run with clock only. Branches as in Supplementary Table~\ref{tab:meas_pressure_tests}, clock as in Supplementary Table~\ref{tab:meas_pressure}. Held-out stations where static-branch RMSE exceeds S-LSTM's (24\,h, clock on): 37, 35 and 35 of 40 for seeds default, 11 and 22 (default-seed median ratio 1.61). A planned 120\,h horizon is excluded: it exceeds the 72\,h window, so its 6 runs duplicated the 72\,h target (Supplementary Table~\ref{tab:meas_runs}). No linear baseline was fitted.}
\label{tab:meas_dh}
\begin{tabular}{@{}llcccc@{}}
\toprule
target & clock & S-LSTM & S-GRU & flat.-window & static-branch \\
\midrule
\multicolumn{6}{@{}l}{\textit{RMSE, mean $\pm$ s.d.}} \\
instantaneous ($^\circ$C) & off & 1.770 $\pm$ 0.070 & 1.949 $\pm$ 0.064 & 1.807 $\pm$ 0.043 & 1.689 $\pm$ 0.070 \\
instantaneous ($^\circ$C) & on & 1.854 $\pm$ 0.173 & 1.980 $\pm$ 0.378 & 1.775 $\pm$ 0.085 & 1.715 $\pm$ 0.038 \\
24\,h ($^\circ$C\,h) & off & 27.2 $\pm$ 1.0 & 29.9 $\pm$ 3.0 & 29.7 $\pm$ 1.5 & 39.5 $\pm$ 0.3 \\
24\,h ($^\circ$C\,h) & on & 28.8 $\pm$ 1.7 & 30.5 $\pm$ 5.9 & 27.9 $\pm$ 0.3 & 37.7 $\pm$ 0.5 \\
48\,h ($^\circ$C\,h) & on & 52.4 $\pm$ 3.1 & 60.4 $\pm$ 2.6 & 51.8 $\pm$ 2.4 & 87.1 $\pm$ 0.7 \\
72\,h ($^\circ$C\,h) & off & 74.3 $\pm$ 3.3 & 78.9 $\pm$ 11.4 & 78.0 $\pm$ 2.6 & 144.1 $\pm$ 2.8 \\
72\,h ($^\circ$C\,h) & on & 81.8 $\pm$ 5.9 & 87.4 $\pm$ 12.1 & 77.0 $\pm$ 7.2 & 138.5 $\pm$ 0.8 \\
\midrule
\multicolumn{6}{@{}l}{\textit{$1 - R^2$ (\%), mean $\pm$ s.d.}} \\
instantaneous & off & 4.4 $\pm$ 0.3 & 5.3 $\pm$ 0.4 & 4.6 $\pm$ 0.2 & 4.0 $\pm$ 0.3 \\
instantaneous & on & 4.8 $\pm$ 0.9 & 5.6 $\pm$ 2.2 & 4.4 $\pm$ 0.4 & 4.1 $\pm$ 0.2 \\
24\,h & off & 2.7 $\pm$ 0.2 & 3.3 $\pm$ 0.7 & 3.2 $\pm$ 0.3 & 5.7 $\pm$ 0.1 \\
24\,h & on & 3.1 $\pm$ 0.4 & 3.5 $\pm$ 1.4 & 2.9 $\pm$ 0.1 & 5.2 $\pm$ 0.1 \\
48\,h & on & 2.6 $\pm$ 0.3 & 3.5 $\pm$ 0.3 & 2.5 $\pm$ 0.2 & 7.2 $\pm$ 0.1 \\
72\,h & off & 2.4 $\pm$ 0.2 & 2.7 $\pm$ 0.8 & 2.6 $\pm$ 0.2 & 8.9 $\pm$ 0.3 \\
72\,h & on & 2.9 $\pm$ 0.4 & 3.3 $\pm$ 0.9 & 2.6 $\pm$ 0.5 & 8.2 $\pm$ 0.1 \\
\bottomrule
\end{tabular}
\end{table}

\begin{table}[!htbp]
\centering
\footnotesize
\caption{\textbf{Measured datasets attempted but not used as main-text evidence.} NOAA, SNOTEL and PeakWeather instantaneous runs come from an earlier audit (Supplementary Table~\ref{tab:meas_runs}); TAO/TRITON stopped at pre-set gates before any network was trained. Mean $\pm$ s.d.\ over 3 training seeds (one station or cell partition) unless marked single fit. NOAA OI SST, optimum-interpolation sea-surface temperature, 20\% of grid cells withheld; relative $L_2$ error, weekly $\|\hat{x} - x\|_2 / \|x\|_2$, averaged; IDW climatology, training-period week-of-year climatology interpolated to withheld cells by inverse-distance weighting (power 4), no sensors; skill, $1 - \mathrm{MSE}/\mathrm{MSE}_\mathrm{clim}$ (MSE, mean squared error). SNOTEL: snow water equivalent (SWE) at manual snow courses from snow-pillow stations; $R^2$ pooled over held-out courses. PeakWeather: lapse-rate fit, per-hour least-squares fit of temperature on elevation through the 12 inputs. TAO/TRITON: Tropical Atmosphere Ocean mooring winds as input for Copernicus Marine (CMEMS) sea-level anomaly (SLA); gates G-T1 (input availability) and G-T2 (two-stage linear skill); ORACLE, ridge fitted at held-out cells. Branches as in Supplementary Table~\ref{tab:meas_pressure_tests}.}
\label{tab:meas_negative}
\begin{tabularx}{\textwidth}{@{}>{\raggedright\arraybackslash}p{22mm}>{\raggedright\arraybackslash}p{27mm}>{\raggedright\arraybackslash}X>{\raggedright\arraybackslash}p{19mm}>{\raggedright\arraybackslash}p{21mm}@{}}
\toprule
dataset & input $\to$ target & key numbers & runs & reason \\
\midrule
NOAA OI SST, weekly\cite{reynolds2002oisst} & SST at 12 grid cells $\to$ SST at withheld grid cells & random 20\% of cells: held-out relative $L_2$ error (\%) S-LSTM 5.94 $\pm$ 0.44; S-GRU 5.99 $\pm$ 0.05; flat.-window 5.55 $\pm$ 0.24; static-branch 5.74 $\pm$ 0.04; sensor-free IDW climatology 4.38 (skill against it $-$0.88 to $-$0.61). 20\% in 10$^\circ$ blocks: held-out relative $L_2$ error (\%) S-LSTM 6.92 $\pm$ 0.33; S-GRU 6.95 $\pm$ 0.07; flat.-window 6.54 $\pm$ 0.20; static-branch 6.79 $\pm$ 0.24; sensor-free IDW climatology 6.36 (skill against it $-$0.20 to $-$0.06) & 24 runs (4 branches $\times$ 2 designs $\times$ 3 seeds) & no network beats a climatology that uses no sensors \\
\addlinespace
SNOTEL\cite{nrcs_snotel} & snow-pillow SWE and fluxes $\to$ SWE at 40 manual snow courses & regional, flux inputs, 26-day window, 3 seeds: $R^2$ S-LSTM 0.237 $\pm$ 0.078; S-GRU 0.233 $\pm$ 0.055; flat.-window 0.181 $\pm$ 0.025; static-branch 0.079 $\pm$ 0.083; spatial mean of training courses 0.163 (single fit). Seed-mean $R^2$ over all 24 branch $\times$ configuration cells: $-$0.06 to 0.38 & 32 runs & weak reconstruction by every method \\
\addlinespace
PeakWeather, instantaneous\cite{zambon2025peakweather} & 2\,m temperature at 12 thermometers $\to$ 40 held-out stations (clock off, 24\,h window) & held-out $R^2$: S-LSTM 0.929 $\pm$ 0.009; S-GRU 0.936 $\pm$ 0.007; flat.-window 0.942 $\pm$ 0.005; static-branch 0.941 $\pm$ 0.003; per-hour lapse-rate fit 0.929 (single fit). RMSE: S-LSTM 2.16 $\pm$ 0.14; S-GRU 2.05 $\pm$ 0.11; flat.-window 1.96 $\pm$ 0.08; static-branch 1.98 $\pm$ 0.06; lapse-rate fit 2.17\,$^\circ$C & 12 runs (4 branches $\times$ 3 seeds) & little gain over a per-hour lapse-rate fit \\
\addlinespace
TAO/TRITON $\to$ CMEMS SLA & mooring winds $\to$ sea-level anomaly at block-held-out 1$^\circ$ cells & G-T1 (planned: $\ge 10$ moorings at $\ge 70$\% valid daily wind in every split and $\ge 40$\% of days jointly usable): 14 moorings qualify individually, but the best 10-mooring set is jointly usable on 0.237 of the days of its worst split (best 12-mooring set 0.159); the largest set reaching 0.40 has 7 moorings. Fail; replaced post hoc by an input-availability rule. G-T2 (planned: two-stage lagged-ridge skill $\ge 0.3$ against a harmonic climatology, 60-day window, test years, block-held-out cells): 0.053 / 0.049 / 0.264 in partitions 0 / 1 / 2 (ORACLE ridge 0.273 / 0.212 / 0.301). Fail & no network run & dropped at gates G-T1 and G-T2 \\
\bottomrule
\end{tabularx}
\end{table}

\begin{table}[!htbp]
\centering
\footnotesize
\caption{\textbf{Training recipe, run inventory and compute.} Upper block: experiments of main-text Fig.~8 (Section~\ref{sec:measured}); lower block: earlier audit runs behind Supplementary Table~\ref{tab:meas_negative} and a GHCN input-count sweep. Window, input length; batch, mini-batch size; cap, maximum epochs; pat., early-stopping patience (audit: values across its run directories); LR, initial learning rate; seeds, training seeds (def., default 1234567); runs, runs with a result file (duplicate audit files counted once; cap30ep, one audit run capped at 30 epochs); h, wall-clock training time summed over runs (inference excluded; jobs could share a GPU). Degree-hours runs: 84 analyzed, 19 in cells dropped before any result was examined, 6 invalid 120\,h runs (Supplementary Table~\ref{tab:meas_dh}). Hardware: as recorded in the run files; none records the DeltaAI GH200 system. The 295 measured-data runs took 17.9\,h in total.}
\label{tab:meas_runs}
\setlength{\tabcolsep}{3.5pt}
\begin{tabularx}{\textwidth}{@{}>{\raggedright\arraybackslash}p{27mm}lllllllr>{\raggedright\arraybackslash}X@{}}
\toprule
dataset & window & batch & cap & pat. & LR & seeds & runs & h & hardware \\
\midrule
GHCN TMAX & 14\,d & 32 & 300 & 20 & 0.001 & def., 11, 22 & 72 & 1.4 & RTX PRO 4000 Blackwell; host not recorded \\
PeakWeather pressure $\to$ 2\,m temperature & 24\,h & 32 & 300 & 20 & 0.001 & def., 11, 22 & 114 & 8.8 & RTX PRO 4000 Blackwell (workstation) \\
PeakWeather degree-hours & 72\,h & 32 & 300 & 20 & 0.001 & def., 11, 22 & 109 & 7.7 & workstation GPUs 0 and 2 (queue scripts); model not recorded \\
\midrule
NOAA OI SST (audit) & various & -- & -- & 10/15/20/40 & -- & -- & 290 & 38.4 & not recorded \\
PeakWeather (audit) & various & -- & -- & 10/20/30/cap30ep & -- & -- & 312 & 12.5 & not recorded \\
GHCN (audit) & various & -- & -- & 20 & -- & -- & 116 & 2.6 & not recorded \\
SNOTEL (audit) & various & -- & -- & 25 & -- & -- & 34 & 0.05 & not recorded \\
\bottomrule
\end{tabularx}
\end{table}

\end{document}